\pdfoutput=1

\documentclass[11pt]{article}

\usepackage[preprint]{acl}

\usepackage{times}
\usepackage{latexsym}
\usepackage{amsmath}
\usepackage[T1]{fontenc}

\usepackage[utf8]{inputenc}

\usepackage{makecell}
\usepackage{arydshln}
\usepackage{fix-cm}
\usepackage{fancyvrb,fvextra}
\usepackage[framemethod=TikZ]{mdframed}
\definecolor{querycolor}{HTML}{3742fa}
\newenvironment{query}[2][]{
    \ifstrempty{#1}
    {
        \mdfsetup{
            frametitle={
                \tikz[baseline=(current bounding box.east),outer sep=0pt]
                \node[anchor=east,rectangle,fill=querycolor]
                {};
            }
        }
    }{
        \mdfsetup{
            frametitle={
                \tikz[baseline=(current bounding box.east),outer sep=0pt]
                \node[anchor=east,rectangle,fill=querycolor]
                {\color{white}{#1}};
            }
        }
    }
    \mdfsetup{
        linecolor=querycolor,
        linewidth=2pt,topline=true,
        frametitleaboveskip=\dimexpr-\ht\strutbox\relax
    }
    \begin{mdframed}[]\relax
}{
    \end{mdframed}
}
\definecolor{groundingcolor}{HTML}{57606f}
\newenvironment{grounding}[2][]{
    \ifstrempty{#1}
    {
        \mdfsetup{
            frametitle={
                \tikz[baseline=(current bounding box.east),outer sep=0pt]
                \node[anchor=east,rectangle,fill=groundingcolor]
                {};
            }
        }
    }{
        \mdfsetup{
            frametitle={
                \tikz[baseline=(current bounding box.east),outer sep=0pt]
                \node[anchor=east,rectangle,fill=groundingcolor]
                {\color{white}{#1}};
            }
        }
    }
    \mdfsetup{
        linecolor=groundingcolor,
        linewidth=2pt,topline=true,
        frametitleaboveskip=\dimexpr-\ht\strutbox\relax
    }
    \begin{mdframed}[]\relax
}{
    \end{mdframed}
}
\definecolor{chosencolor}{HTML}{2ecc71}
\newenvironment{chosen}[2][]{
    \ifstrempty{#1}
    {
        \mdfsetup{
            frametitle={
                \tikz[baseline=(current bounding box.east),outer sep=0pt]
                \node[anchor=east,rectangle,fill=chosencolor]
                {};
            }
        }
    }{
        \mdfsetup{
            frametitle={
                \tikz[baseline=(current bounding box.east),outer sep=0pt]
                \node[anchor=east,rectangle,fill=chosencolor]
                {\color{white}{#1}};
            }
        }
    }
    \mdfsetup{
        linecolor=chosencolor,
        linewidth=2pt,topline=true,
        frametitleaboveskip=\dimexpr-\ht\strutbox\relax
    }
    \begin{mdframed}[]\relax
}{
    \end{mdframed}
}
\definecolor{rejectedcolor}{HTML}{e74c3c}
\newenvironment{rejected}[2][]{
    \ifstrempty{#1}
    {
        \mdfsetup{
            frametitle={
                \tikz[baseline=(current bounding box.east),outer sep=0pt]
                \node[anchor=east,rectangle,fill=rejectedcolor]
                {};
            }
        }
    }{
        \mdfsetup{
            frametitle={
                \tikz[baseline=(current bounding box.east),outer sep=0pt]
                \node[anchor=east,rectangle,fill=rejectedcolor]
                {\color{white}{#1}};
            }
        }
    }
    \mdfsetup{
        linecolor=rejectedcolor,
        linewidth=2pt,topline=true,
        frametitleaboveskip=\dimexpr-\ht\strutbox\relax
    }
    \begin{mdframed}[]\relax
}{
    \end{mdframed}
}
\definecolor{predictioncolor}{HTML}{57606f}
\newenvironment{prediction}[2][]{
    \ifstrempty{#1}
    {
        \mdfsetup{
            frametitle={
                \tikz[baseline=(current bounding box.east),outer sep=0pt]
                \node[anchor=east,rectangle,fill=predictioncolor]
                {};
            }
        }
    }{
        \mdfsetup{
            frametitle={
                \tikz[baseline=(current bounding box.east),outer sep=0pt]
                \node[anchor=east,rectangle,fill=predictioncolor]
                {\color{white}{#1}};
            }
        }
    }
    \mdfsetup{
        linecolor=predictioncolor,
        linewidth=2pt,topline=true,
        frametitleaboveskip=\dimexpr-\ht\strutbox\relax
    }
    \begin{mdframed}[]\relax
}{
    \end{mdframed}
}
\definecolor{systemcolor}{HTML}{a55eea}
\newenvironment{system}[2][]{
    \ifstrempty{#1}
    {
        \mdfsetup{
            frametitle={
                \tikz[baseline=(current bounding box.east),outer sep=0pt]
                \node[anchor=east,rectangle,fill=systemcolor]
                {};
            }
        }
    }{
        \mdfsetup{
            frametitle={
                \tikz[baseline=(current bounding box.east),outer sep=0pt]
                \node[anchor=east,rectangle,fill=systemcolor]
                {\color{white}{#1}};
            }
        }
    }
    \mdfsetup{
        linecolor=systemcolor,
        linewidth=2pt,topline=true,
        frametitleaboveskip=\dimexpr-\ht\strutbox\relax
    }
    \begin{mdframed}[]\relax
}{
    \end{mdframed}
}
\definecolor{usercolor}{HTML}{fdcb6e}
\newenvironment{user}[2][]{
    \ifstrempty{#1}
    {
        \mdfsetup{
            frametitle={
                \tikz[baseline=(current bounding box.east),outer sep=0pt]
                \node[anchor=east,rectangle,fill=usercolor]
                {};
            }
        }
    }{
        \mdfsetup{
            frametitle={
                \tikz[baseline=(current bounding box.east),outer sep=0pt]
                \node[anchor=east,rectangle,fill=usercolor]
                {#1};
            }
        }
    }
    \mdfsetup{
        linecolor=usercolor,
        linewidth=2pt,topline=true,
        frametitleaboveskip=\dimexpr-\ht\strutbox\relax
    }
    \begin{mdframed}[]\relax
}{
    \end{mdframed}
}

\usepackage{microtype}

\usepackage{inconsolata}

\usepackage{graphicx}

\title{\textit{RAGferee}: Building Contextual Reward Models \\for Retrieval-Augmented Generation}

\author{
Andrei C. Coman\thanks{~~Work was done during an internship at Amazon AGI.} \\
Idiap Research Institute, EPFL \\
\href{mailto:andrei.coman@idiap.ch}{andrei.coman@idiap.ch} \\
\And
Ionut-Teodor Sorodoc \\
Amazon AGI \\
\href{mailto:csorionu@amazon.es}{csorionu@amazon.es}
\And
Leonardo F. R. Ribeiro \\
Amazon AGI \\
\href{mailto:leonribe@amazon.com}{leonribe@amazon.com}
\AND
Bill Byrne \\
Amazon AGI \\
\href{mailto:willbyrn@amazon.co.uk}{willbyrn@amazon.co.uk}
\And
James Henderson \\
Idiap Research Institute \\
\href{mailto:james.henderson@idiap.ch}{james.henderson@idiap.ch}
\And
Adrià de Gispert \\
Amazon AGI \\
\href{mailto:agispert@amazon.es}{agispert@amazon.es}
}

\begin{document}
\maketitle

\begin{abstract}

Existing Reward Models (RMs), typically trained on general preference data, struggle in Retrieval Augmented Generation (RAG) settings, which require judging responses for faithfulness to retrieved context, relevance to the user query, appropriate refusals when context is insufficient, completeness and conciseness of information. To address the lack of publicly available RAG-centric preference datasets and specialised RMs, we introduce \textit{RAGferee}\footnote{The \textit{RAGferee} dataset and models can be found at \href{https://github.com/amazon-science/RAGferee}{https://github.com/amazon-science/RAGferee}}, a methodology that repurposes question-answering (QA) datasets into preference pairs that prioritise groundedness over stylistic features, enabling the training of contextual RMs better suited to judging RAG responses. Using \textit{RAGferee}, we curate a small preference dataset of 4K samples and fine-tune RMs ranging from 7B to 24B parameters. Our RAG-centric RMs achieve state-of-the-art performance on \textsc{ContextualJudgeBench}, surpassing existing 70B+ RMs trained on much larger (up to 2.4M samples) general corpora, with an absolute improvement of +15.5\%.

\end{abstract}

\section{Introduction}

Reinforcement Learning from Human Feedback (RLHF) has become a key method for aligning Large Language Models (LLMs) with human preferences \citep{NEURIPS2022_b1efde53}. Building on this foundation, policy optimisation techniques \citep{DBLP:journals/corr/SchulmanWDRK17, NEURIPS2023_a85b405e} incorporated Reinforcement Learning from AI Feedback (RLAIF) \citep{bai2022constitutional} which uses Reward Models (RMs), LLMs trained to judge the quality of generated responses \citep{Bai2022TrainingAH}, as scalable proxies for human evaluation. These RMs are typically trained on general-purpose preference datasets \citep{NEURIPS2024_02fd91a3, NEURIPS2024_0f69b4b9, xu2024magpie, liu2024skywork} and
are expected to act as domain-agnostic evaluators capable of assessing model outputs across a broad range of tasks and domains \citep{vu-etal-2024-foundational, alexandru2025atlaseleneminigeneral}. However, their effectiveness remains under-explored in  Retrieval-Augmented Generation (RAG) \citep{NEURIPS2020_6b493230, pmlr-v119-guu20a, JMLR:v24:23-0037} scenarios, where responses must be judged based on externally retrieved information rather than solely parametric knowledge \citep{ye2024scalarrewardmodellearning, saha2025learningplanreason}.

\begin{figure}
    \centering
    \includegraphics[width=\linewidth]{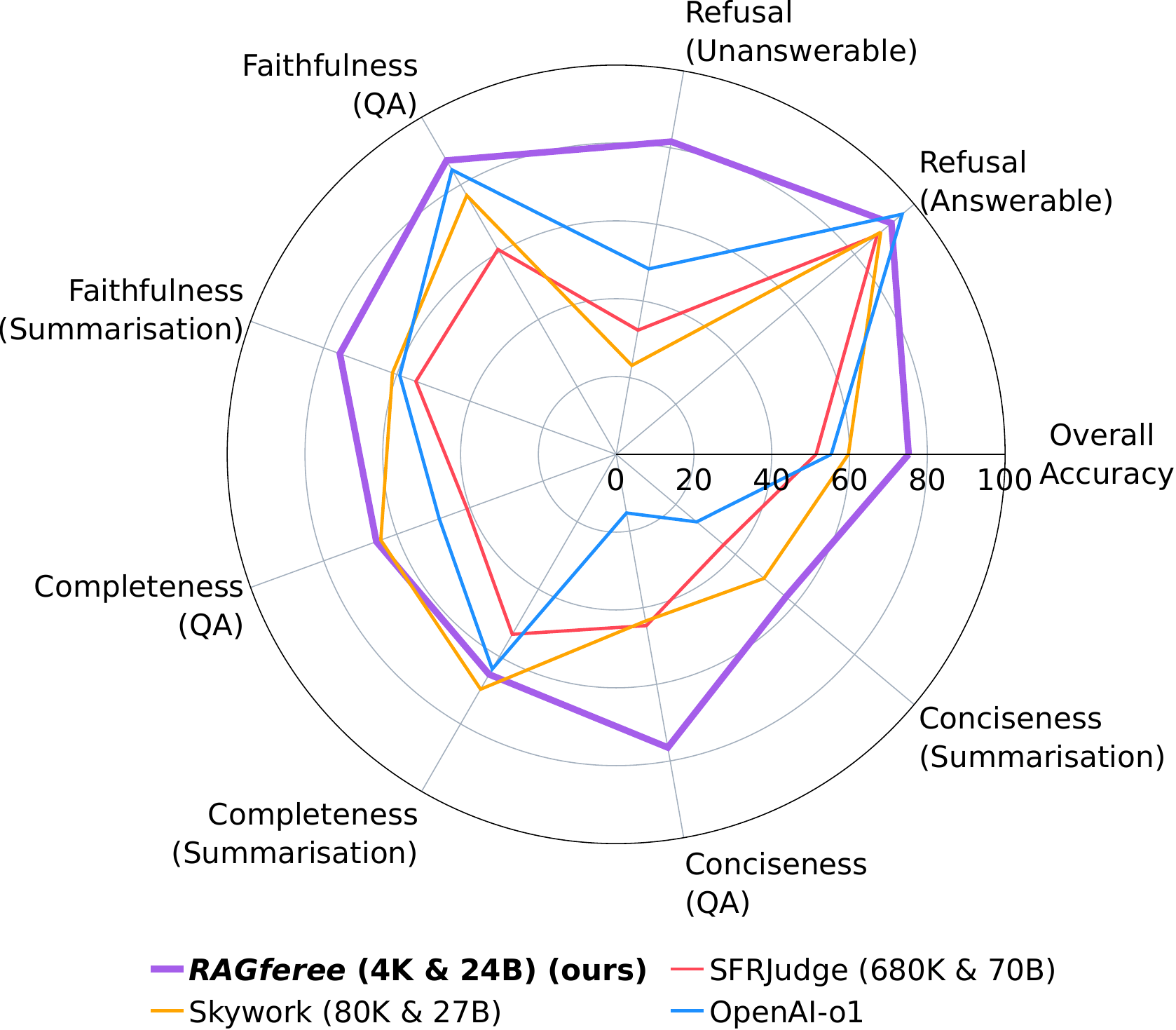}
    \caption{Top RMs (preference pairs \& model size) in \textsc{ContextualJudgeBench}. \textit{RAGferee} is a well-rounded model, showing significant improvements on deflection (Refusal), faithfulness, and conciseness cases.}
    \vspace{-1.5em}
    \label{fig:radar}
\end{figure}

\begin{figure*}[htbp]
    \centering
    \includegraphics[width=0.96\linewidth]{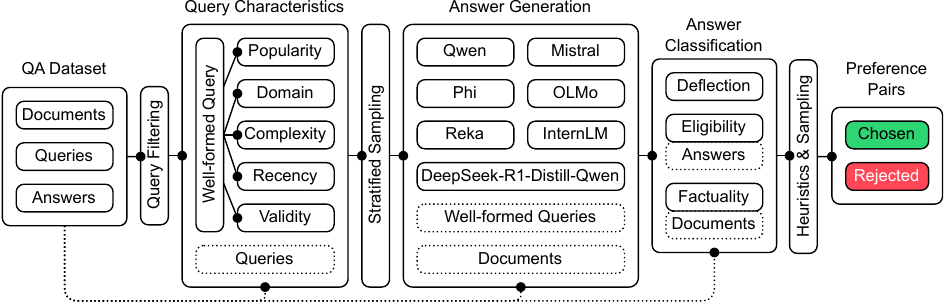}
    \caption{
    \textit{RAGferee} creates RAG-specific preference pairs by repurposing QA datasets. Queries are first filtered, transformed, and categorised into multiple types (Subsection \ref{subsec:query-characteristics}). A stratified subset is then selected to ensure balanced coverage across query types (Subsection \ref{subsec:stratified-sampling}). Candidate answers are generated using a set of LLMs, and labelled based on key qualitative aspects (Subsection \ref{subsec:answer-generation-and-classification}). Finally, heuristics are used to select preference pairs aimed at training contextual RMs (Subsection \ref{subsec:constructing-preference-pairs}).
    }
    \label{fig:ragferee}
\end{figure*}

Context-aware reward modelling introduces unique challenges: RMs must assess not only the quality of responses, but also their faithfulness to the retrieved context, relevance to the user query, and appropriateness of refusals when no valid answer can be provided \citep{jin2024ragrewardbenchbenchmarkingrewardmodels}. Additionally, effective evaluation of RAG responses requires assessing the completeness of the information, ensuring that responses fully incorporate relevant content, as well as their conciseness, making sure the responses are informative without being overly verbose \citep{xu2025doescontextmattercontextualjudgebench}.

\noindent Benchmarks like \textsc{ContextualJudgeBench} \citep{xu2025doescontextmattercontextualjudgebench} question the effectiveness of domain-agnostic evaluators, showing that even large RMs trained on extensive general preference data, struggle in RAG-specific settings (Figure \ref{fig:radar}).

A key barrier to developing contextual RMs is the lack of publicly available preference datasets and openly licensed RMs specifically designed for judging RAG responses \citep{xu2025doescontextmattercontextualjudgebench}. Addressing this critical gap and challenging the reliance on extensive general-purpose preference datasets and large RMs, we make the following \textbf{contributions}:

\vspace{-0.5em}

\begin{itemize}
    \setlength\itemsep{0em}
    \item We introduce \textit{RAGferee}, a methodology for constructing RAG-specific preference datasets aimed at training contextual RMs by repurposing QA datasets into preference pairs.
    \item We curate a small preference dataset of 4K samples via stratified sampling and heuristics that select pairs across queries, models, and answers, prioritising diversity over quantity.
    \item We fine-tune RAG-centric RMs ranging from 7B to 24B parameters that significantly outperform existing 70B+ RMs trained on much larger (up to 2.4M samples) general corpora, achieving state-of-the-art performance on \textsc{ContextualJudgeBench}, with an absolute improvement of +15.5\%.
\end{itemize}

\section{\textit{RAGferee}}

The \textit{RAGferee} methodology (Figure \ref{fig:ragferee}) takes as input a QA dataset $D_{QA} = (q_i, a_i, c_i)_{i=1}^{N}$ where $q_i$ represents the user query, $a_i$ is the reference answer, and $c_i$ denotes the retrieved context. It outputs a set of preference pairs $\hat{D}_{QA} = (\hat{q}_j, \hat{a}_{j_{chosen}}, \hat{a}_{j_{rejected}}, c_j)_{j=1}^{\hat{N} \ll N}$ where $\hat{a}_{j_{chosen}}$ and $\hat{a}_{j_{rejected}}$ are the preferred and non-preferred answers, respectively. To achieve this, \textit{RAGferee} applies the stages described below.

\subsection{Query Characteristics}
\label{subsec:query-characteristics}

Query $q$ is first mapped to a well-formed version $\hat{q} = LLM(q, p_{wf})$, where the $LLM$ function receives specific guidelines from prompt $p_{wf}$ to ensure grammatical correctness, appropriate punctuation, and consistent capitalisation, all while strictly preserving the original semantic meaning. This well-formed version is crucial for subsequent categorisations, ensuring that all queries are processed from a clear and well-structured format.

Query $\hat{q}$ is then mapped to a feature vector $\mathbf{\hat{q}} = LLM(\hat{q}, p_{d})$, where the $LLM$ function, guided by prompt $p_{d}$, extracts discrete features that describe the query along several key dimensions. The resulting feature vector $\mathbf{\hat{q}}$ includes components such as $\hat{q}_{validity}$, which evaluates whether the query is clear, non-harmful, and genuinely seeks factual information; $\hat{q}_{recency}$ which captures how frequently the information in the query changes, from timeless facts to fast-changing, event-based content; $\hat{q}_{popularity}$ which reflects how widely a topic is known, from common subjects to more niche queries; $\hat{q}_{complexity}$ considers the level of reasoning required to provide an answer, from simple answers to those needing synthesis or deeper analysis; and $\hat{q}_{domain}$ categorises each query by its main topic, such as science, entertainment, etc.

The specific guidelines provided to the $LLM$, along with the definitions and category sets for each dimension, are outlined in Appendix \ref{subsubsec:query-charactersitcs-prompts}.

\subsection{Stratified Sampling}
\label{subsec:stratified-sampling}

Feature vector $\mathbf{\hat{q}}$ from Subsection \ref{subsec:query-characteristics} is used to select a representative subset of queries. Initially, these are filtered using the $\hat{q}_{validity}$ feature, where only the valid ones ($\hat{q}_{validity}^{+}$) are retained. Stratified sampling \citep{Neyman1934OnTT} is then applied to the remaining features, ensuring that the final subset spans the full range of query types.

The stratification process also takes into account if a query has a reference answer, which depends on whether relevant information is found in the retrieved context. If a reference answer is available, its length (in words) is used to classify it as short, medium, or long. These categories are based on the 25th percentile (for short) and the 75th percentile (for long) of the overall answer length distribution. Queries without a reference answer are assigned to a separate zero-length category and are treated as deflection queries. This extra stratification ensures a balanced representation of query complexity.

Further details on the implementation of the stratification process are provided in Appendix \ref{subsec:stratified-sampling-details}.

\subsection{Answer Generation and Classification}
\label{subsec:answer-generation-and-classification}

Each query $\hat{q}$ retained after the stratification process, along with its corresponding retrieved context $c$ and the prompt $p_{g}$, is used to generate candidate answers $\hat{a}_k = M_k(\hat{q}, c, p_{g})$, where $M_k$ denotes the $k$-th $LLM$ in a selected set $M = \{M_{1}, M_2, \dots, M_{k}\}_{k=1}^K$ of models (see Section \ref{sec:ragferee-use-case} for the list). Prompt $p_{g}$ is formatted to align with typical RAG scenarios (see Appendix \ref{subsec:answer-generation-prompt}).

Each candidate answer $\hat{a}$ is first mapped to a feature vector $\mathbf{\hat{a}} = LLM(\hat{a}, p_{\hat{a}})$, where the $LLM$ function is guided by a dedicated prompt $p_{\hat{a}}$ (available in Appendix \ref{subsec:answer-classification-prompt}) that contains detailed evaluation guidelines for assessing each answer on specific criteria. The resulting feature vector $\mathbf{\hat{a}}$ includes components such as $\hat{a}_{deflection}$, which evaluates how well the answer handles situations where the query is unanswerable, ensuring the model recognises when no relevant information is available and responds appropriately; $\hat{a}_{eligibility}$, which assesses the relevance of the answer to the query, focusing on how well the response aligns with the user's intent and the reference answer; and $\hat{a}_{factuality}$, which examines the factual accuracy of the answer, ensuring that it contains verifiable, correct information based on the retrieved context.

\subsection{Constructing Preference Pairs}
\label{subsec:constructing-preference-pairs}

This process involves selecting appropriate pairs of candidate answers $\hat{a}$, where one answer is \textit{chosen} (preferred) and the other is \textit{rejected} (non-preferred). We leverage the labels from the feature vector $\mathbf{\hat{a}}$ and follow a set of heuristics described below.

For queries with answers, the \textit{chosen} answer must be both eligible ($\hat{a}_{eligibility}^{+}$), meaning it appropriately addresses the query, and factual ($\hat{a}_{factuality}^{+}$), meaning it is accurate based on the provided grounding. The \textit{rejected} answer is either not eligible ($\hat{a}_{eligibility}^{-}$), meaning it fails to address the query, or eligible ($\hat{a}_{eligibility}^{+}$) but not factual ($\hat{a}_{factuality}^{-}$), where the model may have relied on parametric knowledge rather than the grounding.

For queries without answer, the \textit{chosen} response is one where the model correctly deflects ($\hat{a}_{deflection}^{+}$), acknowledging no valid answer is available from the retrieved context. The \textit{rejected} response is one where the model attempts to answer ($\hat{a}_{deflection}^{-}$) despite lacking relevant or complete information. This tests the model's ability to recognise when no valid response can be given.

To account for variability in models $M_k$ performance and avoid favouring any model based on surface-level clues during RMs training, stratified sampling across models is applied to ensure a balanced distribution of \textit{chosen} and \textit{rejected} pairs. Additionally, since the heuristics may have disrupted the earlier balance of query types, a second round of stratified sampling is conducted to restore it.

\section{\textit{RAGferee}: Use Case}
\label{sec:ragferee-use-case}

Our use case focuses on the MS-MARCO v2.1 training set \citep{bajaj2018msmarcohumangenerated}, which contains approximately 800K queries, 38\% of which are labelled \textit{"No Answer Present."}. While such queries could in principle be used to build deflection cases, we consider these as easy deflections and exclude them, choosing instead to focus only on queries that have explicit answers. Each such query is associated with 10 short passages linked to URLs, and each passage is labelled as either contributive (i.e., can be used to construct the answer) or non-contributive. We attempt to resolve each URL to a corresponding full document in the TREC RAG 2024 corpus \citep{TRECRAG} consisting of roughly 10M documents. If any of the contributive passage cannot be linked to a document in the TREC corpus, the associated query is marked as unanswerable and relabelled as \textit{"No Answer Present."}. These deflection queries are more challenging than the initially excluded ones as they often include relevant (non-contributive) passages but lack key contributive ones, requiring the model to recognise the absence of a grounded answer and respond appropriately. At this stage, the dataset includes approximately 500K queries.

To ensure the final dataset remains free from licensing constraints and supports open research, we exclusively leverage models released under permissive licences, such as Apache 2.0 \citep{ApacheLicense} and MIT \citep{MITLicense}. We use DeepSeek-V3 \citep{deepseekai2024deepseekllmscalingopensource} for the query characteristics (Subsection \ref{subsec:query-characteristics}) and answer classification (Subsection \ref{subsec:answer-generation-and-classification}). We use stratified sampling (Subsection \ref{subsec:stratified-sampling}) to select a balanced 50K subset from the 500K queries, with 45K queries with answers and 5K queries without answers. Then, for answer generation (Subsection \ref{subsec:answer-generation-and-classification}), we leverage models such as Qwen \citep{Bai2023QwenTR, qwen2025qwen25technicalreport}, Mistral \citep{Jiang2023Mistral7}, Phi \citep{gunasekar2023textbooks, abdin2024phi4technicalreport}, OLMo \citep{groeneveld2024olmoacceleratingsciencelanguage}, Reka \citep{rekateam2024rekacoreflashedge}, InternLM \citep{cai2024internlm2}, and DeepSeek \citep{deepseekai2025deepseekr1incentivizingreasoningcapability}. Finally, we construct 5K preference pairs (Subsection \ref{subsec:constructing-preference-pairs}), with 4.5K containing answers (90\%) and 500 without answers (10\%). Of these, 4K (80\%) are used for training, 500 (10\%) for development, and 500 (10\%) for testing.

The full list of models, together with the distributions before and after the sampling stages are reported in Appendix \ref{subsec:experimental-configurations} and Appendix \ref{subsec:distributions}.

\section{Experimental Setup}
\label{sec:experimental-setup}

\paragraph{\textsc{ContextualJudgeBench}} \citep{xu2025doescontextmattercontextualjudgebench} is a benchmark that combines both human annotations and model-based perturbations to provide a diverse and robust evaluation setting of RMs in RAG scenarios. It consists of 2,000 samples, with breakdowns for both QA and summarisation tasks, and evaluates models across four subsets: \textbf{Refusal} evaluates how models handle questions when the context might not contain sufficient information. This includes assessing if a model correctly identifies that a substantive response is better than a refusal for a question answerable from the context ("Refusal (Answerable)"), and conversely, if a model correctly chooses to refuse to answer when the question cannot be answered from the provided context ("Refusal (Unanswerable)"). \textbf{Faithfulness} measures the consistency of the response with the context, ensuring all factual statements in the response are attributable to the context and there are no hallucinations. \textbf{Completeness} assesses how comprehensive the response is, ensuring it covers all essential information needed for a thorough and useful answer. \textbf{Conciseness} determines if the response avoids including more information than what was asked. This includes preventing trivial copy-pasting without meaningful synthesis.

\vspace{0.5em}

\noindent We use \textit{consistent accuracy}, as defined in \textsc{ContextualJudgeBench}, to assess model performance. This metric is tailored to different types of RMs, namely generative RMs and discriminative RMs.

\paragraph{Generative RMs (GRMs)} \citep{Mahan2024GenerativeRM} generate text for a $(\hat{q}, c, \hat{a}_{chosen} \land \hat{a}_{rejected})$ tuple, expressing a preference or comparative judgment. Each test instance is evaluated by jointly comparing the $\hat{a}_{chosen}$ and $\hat{a}_{rejected}$ responses in two orders: first with $\hat{a}_{chosen}$ preceding the $\hat{a}_{rejected}$, and then with the order reversed. A prediction is considered correct only if the model consistently selects the $\hat{a}_{chosen}$ response in both evaluations. This mitigates positional bias and ensures evaluation robustness. Under this setup, random choice corresponds to a \textit{consistent accuracy} of 25\%.

\paragraph{Discriminative RMs (DRMs)} \citep{Yang2024RegularizingHS} assign a scalar score to a $(\hat{q}, c, \hat{a}_{chosen} \lor \hat{a}_{rejected})$ tuple, typically trained with a pairwise loss like Bradley-Terry \citep{Bradley1952RankAO} to estimate relative response quality. Each test instance is evaluated by independently assigning a score to $\hat{a}_{chosen}$ and $\hat{a}_{rejected}$. A prediction is considered correct only if $score(\hat{a}_{chosen})>score(\hat{a}_{rejected})$. Since this setting is not affected by the responses order, random choice corresponds to a \textit{consistent accuracy} of 50\%.

\vspace{0.5em}

\noindent Given our goal of fine-tuning relatively small models (from 7B to 24B) that can potentially serve as value functions in online preference optimisation, we focus on discriminative RMs trained with the Bradley-Terry loss for their ability to produce scalar feedback without decoding.

\begin{table*}[t]
    \centering
    \scalebox{0.645}{
        \begin{tabular}{lccccccccccc}
            \hline
            \textbf{Model} & \textbf{Param.} & \textbf{Pairs} & \textbf{\makecell{Refusal\\(Ans.)}} & \textbf{\makecell{Refusal\\(Unans.)}} & \textbf{\makecell{Faithful.\\(QA)}} & \textbf{\makecell{Faithful.\\(Summ.)}} & \textbf{\makecell{Complete.\\(QA)}} & \textbf{\makecell{Complete.\\(Summ.)}} & \textbf{\makecell{Concise.\\(QA)}} & \textbf{\makecell{Concise.\\(Summ.)}} & \textbf{\makecell{Overall\\Accuracy}} \\
            \hline
            \noalign{\vskip 0.25em}
            \textcolor{gray}{\textit{Random generative}} & \textcolor{gray}{\textit{-}} & \textcolor{gray}{\textit{-}} & \textcolor{gray}{\textit{25.0}} & \textcolor{gray}{\textit{25.0}} & \textcolor{gray}{\textit{25.0}} & \textcolor{gray}{\textit{25.0}} & \textcolor{gray}{\textit{25.0}} & \textcolor{gray}{\textit{25.0}} & \textcolor{gray}{\textit{25.0}} & \textcolor{gray}{\textit{25.0}} & \textcolor{gray}{\textit{25.0}} \\
            \noalign{\vskip 0.25em}
            \hline
            \multicolumn{12}{c}{\textit{Generative (non-reward) Models \citep{xu2025doescontextmattercontextualjudgebench}}}\\
            \hline
            LLaMA-3.1 & 8B & - & 28.0 & 43.2 & 34.8 & 34.8 & 23.2 & 41.0 & 11.4 & 21.3 & 29.7 \\
            LLaMA-3.1 & 70B & - & 59.6 & 48.0 & 58.0 & 48.4 & 38.0 & 51.8 & 15.7 & 27.5 & 43.4 \\
            DeepSeek-R1 & 685B & - & 92.0 & 52.0 & 72.0 & 50.4 & 41.2 & 60.6 & 20.4 & 26.2 & 51.9 \\
            OpenAI-o1 & - & - & 96.0 & 48.4 & 84.4 & 59.2 & 48.4 & 63.7 & 15.3 & 27.0 & \underline{55.3} \\
            \hline
            \multicolumn{12}{c}{\textit{Generative Reward Models \citep{xu2025doescontextmattercontextualjudgebench}}} \\
            \hline
            LLaMA-3.1-Skywork & 8B & 80K & 60.8 & 12.0 & 38.8 & 31.6 & 38.4 & 26.7 & 29.4 & 21.3 & 32.4 \\
            LLaMA-3.1-SFRJudge & 8B & 680K & 70.8 & 22.0 & 40.4 & 38.8 & 40.4 & 43.4 & 27.5 & 31.1 & 39.3 \\
            LLaMA-3.1-SFRJudge & 70B & 680K & 87.6 & 32.4 & 60.8 & 54.8 & 40.8 & 53.4 & 44.7 & 36.1 & \underline{51.4} \\
            LLaMA-3.1-STEval & 70B & 20K & 50.0 & 42.0 & 51.2 & 45.6 & 40.8 & 39.4 & 36.1 & 29.9 & 41.9 \\
            \hline\hline
            \noalign{\vskip 0.25em}
            \textcolor{gray}{\textit{Random discriminative}} & \textcolor{gray}{\textit{-}} & \textcolor{gray}{\textit{-}} & \textcolor{gray}{\textit{50.0}} & \textcolor{gray}{\textit{50.0}} & \textcolor{gray}{\textit{50.0}} & \textcolor{gray}{\textit{50.0}} & \textcolor{gray}{\textit{50.0}} & \textcolor{gray}{\textit{50.0}} & \textcolor{gray}{\textit{50.0}} & \textcolor{gray}{\textit{50.0}} & \textcolor{gray}{\textit{50.0}} \\
            \noalign{\vskip 0.25em}
            \hline
            \multicolumn{12}{c}{\textit{Discriminative Reward Models (baselines)}} \\
            \hline
            InternLM-2 & 7B & 2400K & 78.8 & 12.4 & 71.2 & 67.6 & 46.8 & 70.1 & 16.1 & 38.1 & 50.1 \\
            InternLM-2 & 20B & 2400K & 84.4 & 31.2 & 75.2 & 67.2 & 53.2 & 70.5 & 28.6 & 45.5 & 57.0 \\
            LLaMA-3.1-Skywork-v0.2 & 8B & 80K & 92.8 & 8.0 & 72.8 & 62.8 & 64.4 & 72.9 & 52.5 & 48.4 & 59.4 \\
            Gemma-2-Skywork-v0.2 & 27B & 80K & 88.8 & 23.2 & 76.8 & 61.2 & 64.4 & 69.7 & 43.5 & 49.6 & \underline{59.7} \\
            \hline
            \multicolumn{12}{c}{\textit{RAGferee Discriminative RMs (ours)}} \\
            \hline
            Qwen-2.5-RAGferee & 7B & 4K & 92.8 & 66.4 & 85.2 & 66.8 & 57.6 & 64.1 & 73.3 & 53.7 & 70.0 \\
            Qwen-2.5-RAGferee & 14B & 4K & 92.8 & 71.2 & 86.8 & 70.8 & 65.2 & 66.9 & 71.4 & 52.0 & 72.2 \\
            Mistral-Nemo-RAGferee & 12B & 4K & 92.0 & 82.8 & 82.8 & 68.8 & 62.4 & 62.9 & 86.3 & 57.0 & 74.5 \\
            Mistral-Small-RAGferee & 24B & 4K & 92.4 & 81.6 & 87.2 & 75.6 & 65.6 & 65.3 & 76.5 & 57.0 & \underline{\textbf{75.2}} \\
            \hline
        \end{tabular}
    }
    \caption{\textsc{ContextualJudgeBench} results (best \textbf{overall}/\underline{within group} \textit{consistent accuracy}). For generative models, the metric evaluates whether the \textit{chosen} response is consistently selected over the \textit{rejected} response, regardless of their ordering, with a random chance baseline of 25\%. For discriminative models, the metric evaluates whether the \textit{chosen} response has a higher score than the \textit{rejected} response, with a random chance baseline of 50\%.}
    \label{tab:contextual-judge-bench-results}
\end{table*}

\section{Results and Discussion}
\label{sec:results-and-discussion}

Table \ref{tab:contextual-judge-bench-results} shows the performance of various models on \textsc{ContextualJudgeBench}. The prompts used for inference are provided in Appendix \ref{subsubsec:grm-drm-prompts}.

\paragraph{Generative (non-reward) Models} perform the weakest overall. This is somewhat expected, as they are designed to be general-purpose models and are not specifically optimised to function as RMs. While they can be prompted to perform evaluations, this relies on their general language understanding rather than fine-grained training to judge RAG-specific dimensions. The smaller model, LLaMA-3.1-8B \citep{grattafiori2024llama3herdmodels}, struggles with conciseness (in both QA and summarisation) and completeness (in QA). Larger models, such as LLaMA-3.1-70B, DeepSeek-R1 \citep{deepseekai2025deepseekr1incentivizingreasoningcapability}, and OpenAI-o1 \citep{openai2024openaio1card}, show some improvement but still have noticeable weaknesses in conciseness and completeness, despite stronger performance in refusal and faithfulness.

\vspace{-0.5em}

\paragraph{Generative RMs} are generative models fine-tuned with preference data to enhance their performance as evaluators. Models like LLaMA-3.1-STEval \citep{wang2024selftaughtevaluators}, LLaMA-3.1-Skywork-v0.2 \citep{liu2024skywork}, and LLaMA-3.1-SFRJudge \citep{wang2024directjudgementpreferenceoptimization}, incorporate between 20K and 680K preference pairs. Although their effectiveness in RAG settings remains limited, they generally outperform their non-reward counterparts. While not explicitly optimised for RAG settings, the fine-tuning process enables these models to better capture general answer preferences, such as favouring more informative or relevant responses, which can transfer to contextual evaluations, despite differences in task structure. A notable improvement is seen in the "Refusal (Answerable)" subset, where these models reliably choose an actual answer over a deflection. However, they struggle with "Refusal (Unanswerable)" because their training data lacks deflection signals, making them more likely to select direct answers even when a refusal would be more appropriate.

\vspace{-0.5em}

\paragraph{Discriminative RMs} (baselines) generally perform better than their generative counterparts. While models like InternLM \citep{cai2024internlm2} use an extensive number of preference pairs (up to 2.4M), their performance is not superior to models trained on smaller, more carefully curated datasets, such as the 80K pairs used for the Skywork \citep{liu2024skywork} models. This suggests that the quality and relevance of training data are critical factors. However, challenges remain, particularly in handling appropriate deflections. For example, the LLaMA-3.1-Skywork-v0.2's performance on "Refusal (Unanswerable)" significantly drops to 8.0\% from its base model's initial value of 43.2\%.

\vspace{-0.5em}

\paragraph{\textit{RAGferee} Discriminative RMs} (ours) models \underline{significantly outperform all existing models}, despite being relatively small (from 7B to 24B) and fine-tuned on only 4K RAG-specific preference pairs. They perform well across all subsets, surpassing 70B+ RMs trained on much larger general corpora (from 20K to 2.4M). Notably, they handle deflection ("Refusal (Unanswerable)") cases more effectively, where many others decline sharply. Faithfulness and conciseness are also consistently high. The QA task typically achieves higher scores than summarisation, likely due to the \textit{RAGferee} methodology being specifically tailored for QA in the creation of preference pairs.

\vspace{-0.5em}

\begin{table}[ht]
    \centering
    \scalebox{0.8}{
        \begin{tabular}{lcc|c}
        \hline
        \textbf{Model} & \textbf{Param.} & \textbf{Pairs} & \textbf{\makecell{Overall\\Accuracy}} \\
        \hline
        Qwen-2.5-RAGferee & 7B & 4K & 70.0 \\
        Qwen-2.5-RAGferee & 14B & 4K & 72.2 \\
        Mistral-Nemo-RAGferee & 12B & 4K & 74.5 \\
        Mistral-Small-RAGferee & 24B & 4K & 75.5 \\
        \hdashline
        \multicolumn{4}{c}{\textit{trained w/o grounding (ablation)}} \\
        \hdashline
        Qwen-2.5-RAGferee & 7B & 4K & 52.3 \\
        Qwen-2.5-RAGferee & 14B & 4K & 56.6 \\
        Mistral-Nemo-RAGferee & 12B & 4K & 54.8 \\
        Mistral-Small-RAGferee & 24B & 4K & 56.9 \\
        \hline
        \end{tabular}
    }
    \caption{\textsc{ContextualJudgeBench} results (\textit{contextual accuracy}) for \textit{RAGferee} discriminative RMs from Table \ref{tab:contextual-judge-bench-results} trained without grounding. Preference data alone is not sufficient. Incorporating retrieved context is crucial for accurately judging RAG responses.}
    \label{tab:contextual-judge-bench-results-ragferee-wo-grounding}
\end{table}

\vspace{-0.5em}

\noindent In Table \ref{tab:contextual-judge-bench-results-ragferee-wo-grounding}, we present the results of a \textbf{contrastive study} in which the \textit{RAGferee} models from Table \ref{tab:contextual-judge-bench-results} were trained without grounding information. This setup mimics baseline discriminative RMs by relying solely on preference data related to the answers. We followed the procedure outlined in Subsection \ref{subsec:constructing-preference-pairs} to construct preference pairs, with the only modification being the exclusion of the $\hat{a}_{factuality}$ feature in the candidate answers. The results show a notable drop in performance compared to models trained with grounding, with the performance now aligning with that of baseline discriminative RMs. This emphasises that preference data alone is insufficient and that incorporating retrieved context is crucial for accurately judging RAG responses.

\vspace{-0.5em}

\begin{figure}[hb]
    \centering
    \includegraphics[width=0.94\linewidth]{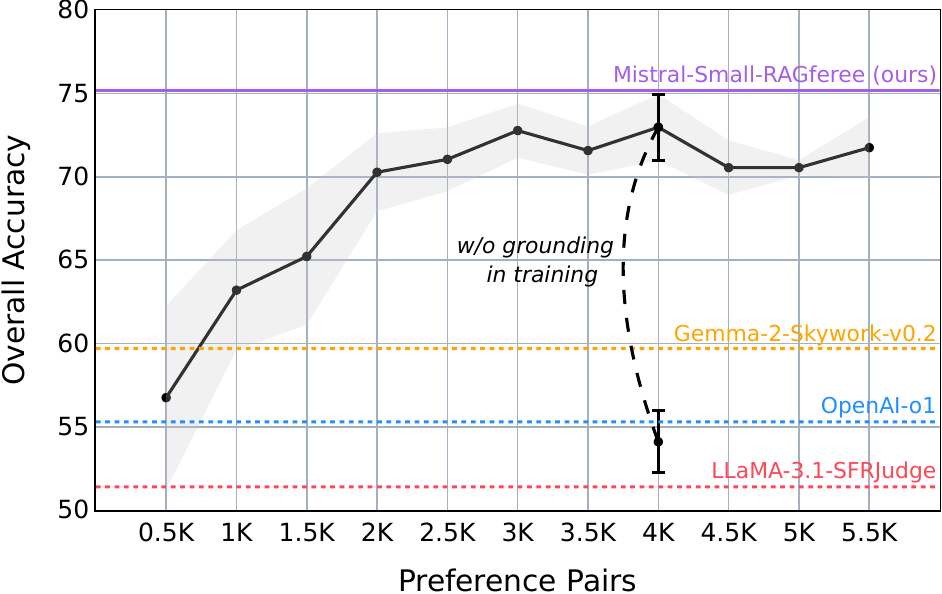}
    \vspace{-0.5em}
    \caption{Performance of \textit{RAGferee} RMs from Table \ref{tab:contextual-judge-bench-results} on \textsc{ContextualJudgeBench} as a function of training preference pairs. The solid black line represents the mean, while the shaded grey area indicates the standard deviation. Dotted lines represent best existing models.}
    \label{fig:data-ablation}
\end{figure}

\noindent Figure \ref{fig:data-ablation} presents the results of a \textbf{data ablation} study, showing the performance of the \textit{RAGferee} models from Table \ref{tab:contextual-judge-bench-results} as a function of the number of training preference pairs. The solid black line represents the mean \textit{"Overall (consistent) Accuracy"} of the models, while the shaded grey area highlights the standard deviation across the models. Even with just 500 preference pairs, the \textit{RAGferee} models already outperform most baseline discriminative RMs (indicated by the dotted lines), and with 1K pairs, they surpass all previous models. The performance peaks at 4K preference pairs. At the same data point, we also plot the results from Table \ref{tab:contextual-judge-bench-results-ragferee-wo-grounding}, revealing a significant performance drop when the models are trained without grounding information. The ungrounded models fall within the performance range of the baseline discriminative RMs, highlighting the importance of grounding in training for achieving good results.

\begin{table}[ht]
    \centering
    \scalebox{0.8}{
        \begin{tabular}{lcc|c}
        \hline
        \textbf{Model} & \textbf{Param.} & \textbf{Pairs} & \textbf{\makecell{Overall\\Accuracy}} \\
        \hline
        \multicolumn{4}{c}{\textit{Discriminative RMs (baselines)}} \\
        \hline
        InternLM-2 & 7B & 2400K & 67.3 \\
        InternLM-2 & 20B & 2400K & 68.7 \\
        LLaMA-3.1-Skywork-v0.2 & 8B & 80K & 71.6 \\
        Gemma-2-Skywork-v0.2 & 27B & 80K & 74.1 \\
        \hdashline
        \multicolumn{4}{c}{\textit{inferenced w/o grounding (ablation)}} \\
        \hdashline
        InternLM-2 & 7B & 2400K & 64.1 \\
        InternLM-2 & 20B & 2400K & 65.1 \\
        LLaMA-3.1-Skywork-v0.2 & 8B & 80K & 67.1 \\
        Gemma-2-Skywork-v0.2 & 27B & 80K & 70.0 \\
        \hline
        \multicolumn{4}{c}{\textit{RAGferee Discriminative RMs (ours)}} \\
        \hline
        Qwen-2.5-RAGferee & 7B & 4K & 59.2 \\
        Qwen-2.5-RAGferee & 14B & 4K & 63.1 \\
        Mistral-Nemo-RAGferee & 12B & 4K & 60.5 \\
        Mistral-Small-RAGferee & 24B & 4K & 61.5 \\
        \hdashline
        \multicolumn{4}{c}{\textit{inferenced w/o grounding (ablation)}} \\
        \hdashline
        Qwen-2.5-RAGferee & 7B & 4K & 67.1 \\
        Qwen-2.5-RAGferee & 14B & 4K & 69.2 \\
        Mistral-Nemo-RAGferee & 12B & 4K & 63.3 \\
        Mistral-Small-RAGferee & 24B & 4K & 67.9 \\
        \hline
        \end{tabular}
    }
    \caption{\textsc{RAG-RewardBench} results (\textit{consistent accuracy}) for discriminative RMs inferenced with or without grounding. Grounding has minimal impact on baseline discriminative RMs (non-RAG), but it significantly influences our \textit{RAGferee} discriminative RMs, which are sensitive to grounding by design.}
    \label{tab:rag-reward-bench}
\end{table}

\paragraph{\textsc{RAG-RewardBench}} \cite{jin2024ragrewardbenchbenchmarkingrewardmodels} is a fully synthetic benchmark of 1,485 samples, designed to evaluate RMs in RAG-specific scenarios such as multi-hop reasoning, fine-grained citation, appropriate abstention, and conflict robustness.

We adopt the same experimental setup as outlined in Section \ref{sec:experimental-setup} and present the results in Table \ref{tab:rag-reward-bench}. Interestingly, in contrast to \textsc{Contextual-JudgeBench} (Table \ref{tab:contextual-judge-bench-results} and Figure \ref{fig:data-ablation}), where the baseline discriminative RMs were significantly outperformed by our grounding-aware \textit{RAGferee} discriminative RMs, the situation here is reversed. In this case, the baseline discriminative RMs outperform our \textit{RAGferee} discriminative RMs, despite the benchmark being designed to assess RAG-specific dimensions. This result is counter-intuitive, as one would expect grounding-aware models to have a clear advantage in a RAG-focused setting.

To further investigate this discrepancy, we conduct an \textbf{ablation study} by entirely removing the grounding from the benchmark and analysing model performance under these conditions. The underlying hypothesis is that if grounding is truly essential to \textsc{RAG-RewardBench}, then the performance of all discriminative RMs will deteriorate towards the random-chance baseline of 50\%. However, this is not the case. The baseline discriminative RMs exhibit only a modest decline of around 4\% in overall accuracy. In contrast, our \textit{RAGferee} discriminative RMs show a notable improvement in performance when grounding is removed, once again contrary to the expected behaviour. This strongly suggests that our models are highly sensitive to grounding and actively use it during inference, as shown on \textsc{Contextual-JudgeBench} (Table \ref{tab:contextual-judge-bench-results} and Figure \ref{fig:data-ablation}), where their accuracy drops much closer to chance in the absence of grounding.

These findings suggest that, despite its stated focus on RAG, \textsc{RAG-RewardBench} may place greater emphasis on general response preferences rather than evaluating grounded behaviour. We believe this is due to the benchmark's fully synthetic data generation and pairs selection process, which may fail to accurately capture signals related to the importance of grounding. Consequently, the grounding provided in \textsc{RAG-RewardBench} may often be irrelevant or unhelpful for reliable response evaluation, potentially accounting for the unexpected performance disparity. Appendix \ref{subsubsec:ragferee-samples} includes examples of issues with the benchmark.

\paragraph{\textit{RAGferee} test set} consists of 500 samples and is used for in-domain evaluation. While it is not a benchmark per se, unlike \textsc{ContextualJudgeBench} and \textsc{RAG-RewardBench}, we report comparative results in Table \ref{tab:ragferee-results}. The overall pattern mirrors the findings from \textsc{ContextualJudgeBench} (Section \ref{sec:results-and-discussion}). Baseline discriminative RMs perform the worst, consistent with earlier observations that they fail to leverage grounding information. This limitation is further highlighted by the substantially better performance of our \textit{RAGferee} models, and their marked drop in performance (close to chance) when trained without grounding, where they again fall within the range of the baseline discriminative RMs.

\begin{table}[ht]
    \centering
    \scalebox{0.8}{
        \begin{tabular}{lcc|c}
        \hline
        \textbf{Model} & \textbf{Param.} & \textbf{Pairs} & \textbf{\makecell{Overall\\Accuracy}} \\
        \hline
        \multicolumn{4}{c}{\textit{Discriminative RMs (baselines)}} \\
        \hline
        InternLM-2 & 7B & 2400K & 46.6 \\
        InternLM-2 & 20B & 2400K & 52.0 \\
        LLaMA-3.1-Skywork-v0.2 & 8B & 80K & 52.8 \\
        Gemma-2-Skywork-v0.2 & 27B & 80K & 57.0 \\
        \hline
        \multicolumn{4}{c}{\textit{RAGferee Discriminative RMs (ours)}} \\
        \hline
        Qwen-2.5-RAGferee & 7B & 4K & 77.4 \\
        Qwen-2.5-RAGferee & 14B & 4K & 83.4 \\
        Mistral-Nemo-RAGferee & 12B & 4K & 81.6 \\
        Mistral-Small-RAGferee & 24B & 4K & 81.8 \\
        \hdashline
        \multicolumn{4}{c}{\textit{trained w/o grounding (ablation)}} \\
        \hdashline
        Qwen-2.5-RAGferee & 7B & 4K & 53.6 \\
        Qwen-2.5-RAGferee & 14B & 4K & 55.2 \\
        Mistral-Nemo-RAGferee & 12B & 4K & 53.2 \\
        Mistral-Small-RAGferee & 24B & 4K & 51.4 \\
        \hline
        \end{tabular}
    }
    \caption{\textsc{RAGferee} test set results (\textit{consistent accuracy}). \textit{RAGferee} models outperform baseline discriminative RMs, but drop to baseline levels when trained without grounding. This shows that grounding information is essential for effective contextual RMs.}
    \label{tab:ragferee-results}
\end{table}

\begin{table}[h]
    \centering
    \scalebox{0.8}{
        \begin{tabular}{lcc|c}
        \hline
        \textbf{Model} & \textbf{Param.} & \textbf{Pairs} & \textbf{\makecell{Overall\\Accuracy}} \\
        \hline
        \multicolumn{4}{c}{\textit{Generative (non-reward) Models (baselines)}} \\
        \hline
        Qwen-2.5 & 7B & - & 28.0 \\
        Qwen-2.5 & 14B & - & 37.3 \\
        Mistral-Nemo & 12B & - & 22.3 \\
        Mistral-Small & 24B & - & 42.4 \\
        \hline
        \multicolumn{4}{c}{\textit{RAGferee Generative RMs (SFT) (ours)}} \\
        \hline
        Qwen-2.5-RAGferee & 7B & 4K & 46.8 \\
        Qwen-2.5-RAGferee & 14B & 4K & 53.9 \\
        Mistral-Nemo-RAGferee & 12B & 4K & 50.0 \\
        Mistral-Small-RAGferee & 24B & 4K & 50.4 \\
        \hline
        \end{tabular}
    }
    \caption{\textsc{ContextualJudgeBench} results (\textit{consistent accuracy}) of \textit{RAGferee} generative RMs trained to output the indicator of the preferred response.}
    \label{tab:generative-ragferee}
\end{table}

\paragraph{\textit{RAGferee} Generative RMs} are an extension of our discriminative RMs, where the models generate textual outputs instead of assigning numerical scores. Specifically, we use supervised fine-tuning (SFT) for training generative RMs which produce completions that include an indicator of the preferred response, such as "\texttt{<answer>A</answer>}" or "\texttt{<answer>B</answer>}". The results on \textsc{ContextualJudgeBench} are presented in Table \ref{tab:generative-ragferee}. The \textit{RAGferee} generative RMs significantly outperform their non-reward counterparts and also surpass all generative models (both reward and non-reward) from Table \ref{tab:contextual-judge-bench-results}, with the exception of the OpenAI-o1 model. However, they still fall behind the baseline discriminative RMs and are considerably less effective than our \textit{RAGferee} discriminative RMs. This outcome aligns with previous studies, such as the one of \citet{Mahan2024GenerativeRM}, which found similar results with non-RAG RMs.

\section{Related Work}

Automatic evaluation of responses remains a persistent challenge \citep{10.1145/3485766, 10.1145/3641289}. Traditional methods, such as matching-based metrics like BLEU \citep{10.3115/1073083.1073135} and ROUGE \citep{lin-2004-rouge}, as well as embedding-based metrics like BERTScore \citep{zhang2020bertscoreevaluatingtextgeneration} and BARTScore \citep{NEURIPS2021_e4d2b6e6}, often struggle to capture subtle semantic nuances and deliver limited performance \citep{li2024llmasajudge}.

Prior works such as MT-Bench \cite{NEURIPS2023_91f18a12}, G-Eval \citep{liu-etal-2023-g}, FLASK \citep{ye2024flaskfinegrainedlanguagemodel}, Prometheus \citep{kim2024prometheusinducingfinegrainedevaluation, kim-etal-2024-prometheus}, FLAME \citep{vu-etal-2024-foundational}, STEval \citep{wang2024selftaughtevaluators}, SFRJudge \citep{wang2024directjudgementpreferenceoptimization}, Skywork-RM \citep{liu2024skywork}, InternLM-RM \cite{cai2024internlm2}, and Critic-RM \citep{yu-etal-2025-self} have explored the use and fine-tuning of strong LLMs as RMs for judging open-ended responses. They show that their judgments often align with human preferences and suggest that LLMs offer a promising alternative for scoring, ranking, and selecting responses across a wide range of tasks \citep{li-etal-2024-leveraging-large, gu2024surveyllmasajudge, gao2024llmbasednlgevaluationcurrent}. 

Several recent works, including FairEval \citep{wang-etal-2024-large-language-models-fair}, PandaLM \citep{wang2024pandalmautomaticevaluationbenchmark}, OffsetBias \citep{park-etal-2024-offsetbias}, CALM \citep{ye2024justiceprejudicequantifyingbiases}, RewardBench \citep{lambert-etal-2025-rewardbench}, VeriScore \citep{song-etal-2024-veriscore}, Minicheck \citep{tang-etal-2024-minicheck}, HalluMeasure \citep{akbar-etal-2024-hallumeasure}, CrossEval \citep{zhong2024lawweakestlinkcross}, and GroUSE \citep{muller-etal-2025-grouse} reveal important limitations of RMs. These include persistent issues such as hallucinations \citep{tonmoy2024comprehensivesurveyhallucinationmitigation}, fairness and biases \citep{gallegos-etal-2024-bias} concerns, and lack of robustness \citep{10.1145/3689217.3690621}. 

Our work connects to the previous ones by challenging a prevailing assumption: that RMs trained on extensive general-purpose preference datasets such as Helpsteer \citep{NEURIPS2024_02fd91a3}, Wildguard \citep{NEURIPS2024_0f69b4b9}, Magpie \citep{xu2024magpie}, as well as those used in STEval \citep{wang2024selftaughtevaluators}, InternLM \cite{cai2024internlm2}, SFRJudge \citep{wang2024directjudgementpreferenceoptimization}, and Skywork \citep{liu2024skywork}, can serve as domain-agnostic evaluators, capable of assessing responses across a wide range of tasks and domains \citep{vu-etal-2024-foundational, alexandru2025atlaseleneminigeneral}. In line with recent efforts like \textsc{ContextualJudgeBench} \citep{xu2025doescontextmattercontextualjudgebench} and \textsc{RAG-RewardBench} \citep{jin2024ragrewardbenchbenchmarkingrewardmodels}, we show that large RMs trained on general preference datasets perform poorly in RAG-specific settings. To address this limitation, we introduce a methodology that curates a small dataset by repurposing QA datasets into preference pairs, which is then used to fine-tune relatively small RAG-centric RMs, leading to significantly improved judges for contextual evaluation scenarios.

\section{Conclusion}

We show that existing Reward Models (RMs), typically trained on general preference data, exhibit limited effectiveness in RAG settings. Such tasks require evaluating model responses based on retrieved context, across multiple dimensions: faithfulness to the source material, relevance to the user query, appropriate refusals when the context is insufficient, and the completeness and conciseness of the information provided.

We address the lack of publicly available RAG-specific preference datasets and specialised RMs by introducing \textit{RAGferee}, a methodology that repurposes existing QA datasets into preference pairs by filtering, transforming, categorising queries, selecting a stratified subset for balanced coverage, generating candidate answers from various LLMs, labelling them based on key qualitative criteria, and applying heuristics to select the best pairs for training contextual RMs.

We construct a small preference dataset of 4K samples and fine-tune RMs ranging from 7B to 24B parameters. We evaluate generative (non-reward) models, generative RMs, and discriminative RMs, and show that RAG-specific RMs trained with \textit{RAGferee} are better suited for assessing context-grounded responses. Our RAG-centric RMs achieve state-of-the-art performance (+15.5\% absolute improvement) on \textsc{ContextualJudgeBench}, surpassing 70B+ general-purpose RMs trained on up to 2.4M preference samples, despite using a much smaller dataset and models.

\section*{Limitations}

Our work focuses on a single use case, which inherently limits the generalisability of the findings. The size of the curated dataset, while carefully balanced, remains relatively small compared to large-scale preference datasets, which may constrain model robustness and performance in more diverse or complex scenarios.

While using a single strong LLM for query classification (Subsection \ref{subsec:query-characteristics}) is a straightforward approach, relying on the same model for answer classification (Subsection \ref{subsec:answer-generation-and-classification}) introduces potential bias, as the labels generated are directly used to create preference pairs. An ensemble of labelling models could help mitigate this bias by providing more robust and diverse annotations. Furthermore, we did not study the correlation of these labels with human judgments, which could be valuable for uncovering discrepancies and biases between model-generated annotations and human preferences.

In our stratified sampling procedure (Subsection \ref{subsec:stratified-sampling}), we prioritised maximising diversity across queries, answers, and models. Although this strategy yielded promising results, it may not represent the optimal approach. Similarly, the use of heuristics for selecting preference pairs, by their nature, introduces biases and may oversimplify the selection process, potentially missing finer-grained distinctions in answer quality or contextual relevance that more sophisticated methods could identify.

Although our results show promising scalability properties with models up to 24B parameters, further investigation is needed to understand behaviour at larger scales. Additionally, while our RMs perform well on standard benchmarks, their effectiveness as value functions in policy optimisation remains an open area for exploration.

Finally, our generative RMs (Section \ref{sec:results-and-discussion}) are limited to a rather straightforward fine-tuning SFT approach, where models are required to output the indicator of the preferred response. Extending this framework to incorporate Chain-of-Thought (CoT) \citep{wei2023chainofthoughtpromptingelicitsreasoning} justifications prior to the preference indicator or applying more recent strategies like GRPO \citep{shao2024deepseekmathpushinglimitsmathematical}, offers promising avenues for further research.  

\section*{Ethics Statement}

Automatic evaluation of responses remains a persistent challenge. Although using and fine-tuning strong LLMs as RMs for judging responses shows promise, these models share the same inherent risks and ethical considerations highlighted in prior research on pretrained and instruction-tuned models. Since RMs are designed to align with human preferences, they may inherit and amplify existing human biases present in training data, potentially leading to unfair or discriminatory outcomes related to race, gender, or other sensitive attributes. Furthermore, over-reliance on these models risks automating decisions that require human judgement.

Our work promotes open and responsible research by committing to transparency in model development and by exclusively using publicly available datasets and models with permissive licences.

\section*{Acknowledgements}
Andrei C. Coman gratefully acknowledges Amazon for hosting the internship during which this research was carried out. He extends his heartfelt thanks to Ionut-Teodor Sorodoc, Adrià de Gispert Ramis, Leonardo F. R. Ribeiro, Bill Byrne, and all of his other colleagues for their valuable guidance and stimulating discussions. He further expresses sincere gratitude to his PhD supervisor, James Henderson of the Idiap Research Institute, for his constant support and collaboration.

Bill Byrne holds concurrent appointments as an
Amazon Scholar and as Professor of Information
Engineering at the University of Cambridge. This
paper describes work performed at Amazon.

\bibliography{anthology,custom}

\begin{thebibliography}{81}
\providecommand{\natexlab}[1]{#1}

\bibitem[{Abdin et~al.(2024)Abdin, Aneja, Behl, Bubeck, Eldan, Gunasekar, Harrison, Hewett, Javaheripi, Kauffmann, Lee, Lee, Li, Liu, Mendes, Nguyen, Price, de~Rosa, Saarikivi, Salim, Shah, Wang, Ward, Wu, Yu, Zhang, and Zhang}]{abdin2024phi4technicalreport}
Marah Abdin, Jyoti Aneja, Harkirat Behl, Sébastien Bubeck, Ronen Eldan, Suriya Gunasekar, Michael Harrison, Russell~J. Hewett, Mojan Javaheripi, Piero Kauffmann, James~R. Lee, Yin~Tat Lee, Yuanzhi Li, Weishung Liu, Caio C.~T. Mendes, Anh Nguyen, Eric Price, Gustavo de~Rosa, Olli Saarikivi, and 8 others. 2024.
\newblock \href {https://arxiv.org/abs/2412.08905} {Phi-4 technical report}.
\newblock \emph{Preprint}, arXiv:2412.08905.

\bibitem[{Akbar et~al.(2024)Akbar, Hossain, Wood, Chin, Salinas, Alvarez, and Cornejo}]{akbar-etal-2024-hallumeasure}
Shayan~Ali Akbar, Md~Mosharaf Hossain, Tess Wood, Si-Chi Chin, Erica~M Salinas, Victor Alvarez, and Erwin Cornejo. 2024.
\newblock \href {https://doi.org/10.18653/v1/2024.emnlp-main.837} {{H}allu{M}easure: Fine-grained hallucination measurement using chain-of-thought reasoning}.
\newblock In \emph{Proceedings of the 2024 Conference on Empirical Methods in Natural Language Processing}, pages 15020--15037, Miami, Florida, USA. Association for Computational Linguistics.

\bibitem[{Alexandru et~al.(2025)Alexandru, Calvi, Broomfield, Golden, Dai, Leys, Burger, Bartolo, Engeler, Pisupati, Drane, and Park}]{alexandru2025atlaseleneminigeneral}
Andrei Alexandru, Antonia Calvi, Henry Broomfield, Jackson Golden, Kyle Dai, Mathias Leys, Maurice Burger, Max Bartolo, Roman Engeler, Sashank Pisupati, Toby Drane, and Young~Sun Park. 2025.
\newblock \href {https://arxiv.org/abs/2501.17195} {Atla selene mini: A general purpose evaluation model}.
\newblock \emph{Preprint}, arXiv:2501.17195.

\bibitem[{ASF(2004)}]{ApacheLicense}
ASF. 2004.
\newblock \href {https://www.apache.org/licenses/LICENSE-2.0} {Apache license, version 2.0}.

\bibitem[{Bai et~al.(2023)Bai, Bai, Chu, Cui, Dang, Deng, Fan, Ge, Han, Huang, Hui, Ji, Li, Lin, Lin, Liu, Liu, Lu, Lu, Ma, Men, Ren, Ren, Tan, Tan, Tu, Wang, Wang, Wang, Wu, Xu, Xu, Yang, Yang, Yang, Yang, Yang, Yao, Yu, Bowen, Yuan, Yuan, Zhang, Zhang, Zhang, Zhang, Zhou, Zhou, Zhou, and Zhu}]{Bai2023QwenTR}
Jinze Bai, Shuai Bai, Yunfei Chu, Zeyu Cui, Kai Dang, Xiaodong Deng, Yang Fan, Wenhang Ge, Yu~Han, Fei Huang, Binyuan Hui, Luo Ji, Mei Li, Junyang Lin, Runji Lin, Dayiheng Liu, Gao Liu, Chengqiang Lu, K.~Lu, and 31 others. 2023.
\newblock \href {https://api.semanticscholar.org/CorpusID:263134555} {Qwen technical report}.
\newblock \emph{ArXiv}, abs/2309.16609.

\bibitem[{Bai et~al.(2022{\natexlab{a}})Bai, Jones, Ndousse, Askell, Chen, Dassarma, Drain, Fort, Ganguli, Henighan, Joseph, Kadavath, Kernion, Conerly, El-Showk, Elhage, Hatfield-Dodds, Hernandez, Hume, Johnston, Kravec, Lovitt, Nanda, Olsson, Amodei, Brown, Clark, McCandlish, Olah, Mann, and Kaplan}]{Bai2022TrainingAH}
Yuntao Bai, Andy Jones, Kamal Ndousse, Amanda Askell, Anna Chen, Nova Dassarma, Dawn Drain, Stanislav Fort, Deep Ganguli, Tom Henighan, Nicholas Joseph, Saurav Kadavath, John Kernion, Tom Conerly, Sheer El-Showk, Nelson Elhage, Zac Hatfield-Dodds, Danny Hernandez, Tristan Hume, and 12 others. 2022{\natexlab{a}}.
\newblock \href {https://api.semanticscholar.org/CorpusID:248118878} {Training a helpful and harmless assistant with reinforcement learning from human feedback}.
\newblock \emph{ArXiv}, abs/2204.05862.

\bibitem[{Bai et~al.(2022{\natexlab{b}})Bai, Kadavath, Kundu, Askell, Kernion, Jones, Chen, Goldie, Mirhoseini, McKinnon et~al.}]{bai2022constitutional}
Yuntao Bai, Saurav Kadavath, Sandipan Kundu, Amanda Askell, Jackson Kernion, Andy Jones, Anna Chen, Anna Goldie, Azalia Mirhoseini, Cameron McKinnon, and 1 others. 2022{\natexlab{b}}.
\newblock Constitutional ai: Harmlessness from ai feedback.
\newblock \emph{arXiv preprint arXiv:2212.08073}.

\bibitem[{Bajaj et~al.(2018)Bajaj, Campos, Craswell, Deng, Gao, Liu, Majumder, McNamara, Mitra, Nguyen, Rosenberg, Song, Stoica, Tiwary, and Wang}]{bajaj2018msmarcohumangenerated}
Payal Bajaj, Daniel Campos, Nick Craswell, Li~Deng, Jianfeng Gao, Xiaodong Liu, Rangan Majumder, Andrew McNamara, Bhaskar Mitra, Tri Nguyen, Mir Rosenberg, Xia Song, Alina Stoica, Saurabh Tiwary, and Tong Wang. 2018.
\newblock \href {https://arxiv.org/abs/1611.09268} {Ms marco: A human generated machine reading comprehension dataset}.
\newblock \emph{Preprint}, arXiv:1611.09268.

\bibitem[{Bradley and Terry(1952)}]{Bradley1952RankAO}
Ralph~Allan Bradley and Milton~E. Terry. 1952.
\newblock \href {https://api.semanticscholar.org/CorpusID:125209808} {Rank analysis of incomplete block designs: I. the method of paired comparisons}.
\newblock \emph{Biometrika}, 39:324.

\bibitem[{Cai et~al.(2024)Cai, Cao, Chen, Chen, Chen, Chen, Chen, Chen, Chen, Chu, Dong, Duan, Fan, Fei, Gao, Ge, Gu, Gu, Gui, Guo, Guo, He, Hu, Huang, Jiang, Jiao, Jin, Lei, Li, Li, Li, Li, Li, Li, Liu, Liu, Hong, Liu, Liu, Liu, Lv, Lv, Lv, Ma, Ma, Ma, Ning, Ouyang, Qiu, Qu, Shang, Shao, Song, Song, Sui, Sun, Sun, Tang, Wang, Wang, Wang, Wang, Wang, Wang, Wang, Wei, Weng, Wu, Xiong, Xu, Xu, Yan, Yan, Yang, Ye, Ying, Yu, Yu, Zang, Zhang, Zhang, Zhang, Zhang, Zhang, Zhang, Zhang, Zhang, Zhang, Zhang, Zhang, Zhao, Zhao, Zhao, Zhou, Zhou, Zhuo, Zou, Qiu, Qiao, and Lin}]{cai2024internlm2}
Zheng Cai, Maosong Cao, Haojiong Chen, Kai Chen, Keyu Chen, Xin Chen, Xun Chen, Zehui Chen, Zhi Chen, Pei Chu, Xiaoyi Dong, Haodong Duan, Qi~Fan, Zhaoye Fei, Yang Gao, Jiaye Ge, Chenya Gu, Yuzhe Gu, Tao Gui, and 81 others. 2024.
\newblock \href {https://arxiv.org/abs/2403.17297} {Internlm2 technical report}.
\newblock \emph{Preprint}, arXiv:2403.17297.

\bibitem[{Chang et~al.(2024)Chang, Wang, Wang, Wu, Yang, Zhu, Chen, Yi, Wang, Wang, Ye, Zhang, Chang, Yu, Yang, and Xie}]{10.1145/3641289}
Yupeng Chang, Xu~Wang, Jindong Wang, Yuan Wu, Linyi Yang, Kaijie Zhu, Hao Chen, Xiaoyuan Yi, Cunxiang Wang, Yidong Wang, Wei Ye, Yue Zhang, Yi~Chang, Philip~S. Yu, Qiang Yang, and Xing Xie. 2024.
\newblock \href {https://doi.org/10.1145/3641289} {A survey on evaluation of large language models}.
\newblock \emph{ACM Trans. Intell. Syst. Technol.}, 15(3).

\bibitem[{Dao(2023)}]{dao2023flashattention2fasterattentionbetter}
Tri Dao. 2023.
\newblock \href {https://arxiv.org/abs/2307.08691} {Flashattention-2: Faster attention with better parallelism and work partitioning}.
\newblock \emph{Preprint}, arXiv:2307.08691.

\bibitem[{DeepSeek-AI et~al.(2024)DeepSeek-AI, :, Bi, Chen, Chen, Chen, Dai, Deng, Ding, Dong, Du, Fu, Gao, Gao, Gao, Ge, Guan, Guo, Guo, Hao, Hao, He, Hu, Huang, Li, Li, Li, Li, Li, Liang, Lin, Liu, Liu, Liu, Liu, Liu, Liu, Lu, Lu, Luo, Ma, Nie, Pei, Piao, Qiu, Qu, Ren, Ren, Ruan, Sha, Shao, Song, Su, Sun, Sun, Tang, Wang, Wang, Wang, Wang, Wang, Wu, Wu, Xie, Xie, Xie, Xiong, Xu, Xu, Xu, Yang, You, Yu, Yu, Zhang, Zhang, Zhang, Zhang, Zhang, Zhang, Zhang, Zhang, Zhao, Zhao, Zhou, Zhou, Zhu, and Zou}]{deepseekai2024deepseekllmscalingopensource}
DeepSeek-AI, :, Xiao Bi, Deli Chen, Guanting Chen, Shanhuang Chen, Damai Dai, Chengqi Deng, Honghui Ding, Kai Dong, Qiushi Du, Zhe Fu, Huazuo Gao, Kaige Gao, Wenjun Gao, Ruiqi Ge, Kang Guan, Daya Guo, Jianzhong Guo, and 69 others. 2024.
\newblock \href {https://arxiv.org/abs/2401.02954} {Deepseek llm: Scaling open-source language models with longtermism}.
\newblock \emph{Preprint}, arXiv:2401.02954.

\bibitem[{DeepSeek-AI et~al.(2025)DeepSeek-AI, Guo, Yang, Zhang, Song, Zhang, Xu, Zhu, Ma, Wang, Bi, Zhang, Yu, Wu, Wu, Gou, Shao, Li, Gao, Liu, Xue, Wang, Wu, Feng, Lu, Zhao, Deng, Zhang, Ruan, Dai, Chen, Ji, Li, Lin, Dai, Luo, Hao, Chen, Li, Zhang, Bao, Xu, Wang, Ding, Xin, Gao, Qu, Li, Guo, Li, Wang, Chen, Yuan, Qiu, Li, Cai, Ni, Liang, Chen, Dong, Hu, Gao, Guan, Huang, Yu, Wang, Zhang, Zhao, Wang, Zhang, Xu, Xia, Zhang, Zhang, Tang, Li, Wang, Li, Tian, Huang, Zhang, Wang, Chen, Du, Ge, Zhang, Pan, Wang, Chen, Jin, Chen, Lu, Zhou, Chen, Ye, Wang, Yu, Zhou, Pan, Li, Zhou, Wu, Ye, Yun, Pei, Sun, Wang, Zeng, Zhao, Liu, Liang, Gao, Yu, Zhang, Xiao, An, Liu, Wang, Chen, Nie, Cheng, Liu, Xie, Liu, Yang, Li, Su, Lin, Li, Jin, Shen, Chen, Sun, Wang, Song, Zhou, Wang, Shan, Li, Wang, Wei, Zhang, Xu, Li, Zhao, Sun, Wang, Yu, Zhang, Shi, Xiong, He, Piao, Wang, Tan, Ma, Liu, Guo, Ou, Wang, Gong, Zou, He, Xiong, Luo, You, Liu, Zhou, Zhu, Xu, Huang, Li, Zheng, Zhu, Ma, Tang, Zha, Yan, Ren, Ren, Sha, Fu, Xu, Xie, Zhang,
  Hao, Ma, Yan, Wu, Gu, Zhu, Liu, Li, Xie, Song, Pan, Huang, Xu, Zhang, and Zhang}]{deepseekai2025deepseekr1incentivizingreasoningcapability}
DeepSeek-AI, Daya Guo, Dejian Yang, Haowei Zhang, Junxiao Song, Ruoyu Zhang, Runxin Xu, Qihao Zhu, Shirong Ma, Peiyi Wang, Xiao Bi, Xiaokang Zhang, Xingkai Yu, Yu~Wu, Z.~F. Wu, Zhibin Gou, Zhihong Shao, Zhuoshu Li, Ziyi Gao, and 181 others. 2025.
\newblock \href {https://arxiv.org/abs/2501.12948} {Deepseek-r1: Incentivizing reasoning capability in llms via reinforcement learning}.
\newblock \emph{Preprint}, arXiv:2501.12948.

\bibitem[{Gallegos et~al.(2024)Gallegos, Rossi, Barrow, Tanjim, Kim, Dernoncourt, Yu, Zhang, and Ahmed}]{gallegos-etal-2024-bias}
Isabel~O. Gallegos, Ryan~A. Rossi, Joe Barrow, Md~Mehrab Tanjim, Sungchul Kim, Franck Dernoncourt, Tong Yu, Ruiyi Zhang, and Nesreen~K. Ahmed. 2024.
\newblock \href {https://doi.org/10.1162/coli_a_00524} {Bias and fairness in large language models: A survey}.
\newblock \emph{Computational Linguistics}, 50(3):1097--1179.

\bibitem[{Gao et~al.(2024)Gao, Hu, Ruan, Pu, and Wan}]{gao2024llmbasednlgevaluationcurrent}
Mingqi Gao, Xinyu Hu, Jie Ruan, Xiao Pu, and Xiaojun Wan. 2024.
\newblock \href {https://arxiv.org/abs/2402.01383} {Llm-based nlg evaluation: Current status and challenges}.
\newblock \emph{Preprint}, arXiv:2402.01383.

\bibitem[{Grattafiori et~al.(2024)Grattafiori, Dubey, Jauhri, Pandey, Kadian, Al-Dahle, Letman, Mathur, Schelten, Vaughan, Yang, Fan, Goyal, Hartshorn, Yang, Mitra, Sravankumar, Korenev, Hinsvark, Rao, Zhang, Rodriguez, Gregerson, Spataru, Roziere, Biron, Tang, Chern, Caucheteux, Nayak, Bi, Marra, McConnell, Keller, Touret, Wu, Wong, Ferrer, Nikolaidis, Allonsius, Song, Pintz, Livshits, Wyatt, Esiobu, Choudhary, Mahajan, Garcia-Olano, Perino, Hupkes, Lakomkin, AlBadawy, Lobanova, Dinan, Smith, Radenovic, Guzmán, Zhang, Synnaeve, Lee, Anderson, Thattai, Nail, Mialon, Pang, Cucurell, Nguyen, Korevaar, Xu, Touvron, Zarov, Ibarra, Kloumann, Misra, Evtimov, Zhang, Copet, Lee, Geffert, Vranes, Park, Mahadeokar, Shah, van~der Linde, Billock, Hong, Lee, Fu, Chi, Huang, Liu, Wang, Yu, Bitton, Spisak, Park, Rocca, Johnstun, Saxe, Jia, Alwala, Prasad, Upasani, Plawiak, Li, Heafield, Stone, El-Arini, Iyer, Malik, Chiu, Bhalla, Lakhotia, Rantala-Yeary, van~der Maaten, Chen, Tan, Jenkins, Martin, Madaan, Malo, Blecher,
  Landzaat, de~Oliveira, Muzzi, Pasupuleti, Singh, Paluri, Kardas, Tsimpoukelli, Oldham, Rita, Pavlova, Kambadur, Lewis, Si, Singh, Hassan, Goyal, Torabi, Bashlykov, Bogoychev, Chatterji, Zhang, Duchenne, Çelebi, Alrassy, Zhang, Li, Vasic, Weng, Bhargava, Dubal, Krishnan, Koura, Xu, He, Dong, Srinivasan, Ganapathy, Calderer, Cabral, Stojnic, Raileanu, Maheswari, Girdhar, Patel, Sauvestre, Polidoro, Sumbaly, Taylor, Silva, Hou, Wang, Hosseini, Chennabasappa, Singh, Bell, Kim, Edunov, Nie, Narang, Raparthy, Shen, Wan, Bhosale, Zhang, Vandenhende, Batra, Whitman, Sootla, Collot, Gururangan, Borodinsky, Herman, Fowler, Sheasha, Georgiou, Scialom, Speckbacher, Mihaylov, Xiao, Karn, Goswami, Gupta, Ramanathan, Kerkez, Gonguet, Do, Vogeti, Albiero, Petrovic, Chu, Xiong, Fu, Meers, Martinet, Wang, Wang, Tan, Xia, Xie, Jia, Wang, Goldschlag, Gaur, Babaei, Wen, Song, Zhang, Li, Mao, Coudert, Yan, Chen, Papakipos, Singh, Srivastava, Jain, Kelsey, Shajnfeld, Gangidi, Victoria, Goldstand, Menon, Sharma, Boesenberg,
  Baevski, Feinstein, Kallet, Sangani, Teo, Yunus, Lupu, Alvarado, Caples, Gu, Ho, Poulton, Ryan, Ramchandani, Dong, Franco, Goyal, Saraf, Chowdhury, Gabriel, Bharambe, Eisenman, Yazdan, James, Maurer, Leonhardi, Huang, Loyd, Paola, Paranjape, Liu, Wu, Ni, Hancock, Wasti, Spence, Stojkovic, Gamido, Montalvo, Parker, Burton, Mejia, Liu, Wang, Kim, Zhou, Hu, Chu, Cai, Tindal, Feichtenhofer, Gao, Civin, Beaty, Kreymer, Li, Adkins, Xu, Testuggine, David, Parikh, Liskovich, Foss, Wang, Le, Holland, Dowling, Jamil, Montgomery, Presani, Hahn, Wood, Le, Brinkman, Arcaute, Dunbar, Smothers, Sun, Kreuk, Tian, Kokkinos, Ozgenel, Caggioni, Kanayet, Seide, Florez, Schwarz, Badeer, Swee, Halpern, Herman, Sizov, Guangyi, Zhang, Lakshminarayanan, Inan, Shojanazeri, Zou, Wang, Zha, Habeeb, Rudolph, Suk, Aspegren, Goldman, Zhan, Damlaj, Molybog, Tufanov, Leontiadis, Veliche, Gat, Weissman, Geboski, Kohli, Lam, Asher, Gaya, Marcus, Tang, Chan, Zhen, Reizenstein, Teboul, Zhong, Jin, Yang, Cummings, Carvill, Shepard, McPhie,
  Torres, Ginsburg, Wang, Wu, U, Saxena, Khandelwal, Zand, Matosich, Veeraraghavan, Michelena, Li, Jagadeesh, Huang, Chawla, Huang, Chen, Garg, A, Silva, Bell, Zhang, Guo, Yu, Moshkovich, Wehrstedt, Khabsa, Avalani, Bhatt, Mankus, Hasson, Lennie, Reso, Groshev, Naumov, Lathi, Keneally, Liu, Seltzer, Valko, Restrepo, Patel, Vyatskov, Samvelyan, Clark, Macey, Wang, Hermoso, Metanat, Rastegari, Bansal, Santhanam, Parks, White, Bawa, Singhal, Egebo, Usunier, Mehta, Laptev, Dong, Cheng, Chernoguz, Hart, Salpekar, Kalinli, Kent, Parekh, Saab, Balaji, Rittner, Bontrager, Roux, Dollar, Zvyagina, Ratanchandani, Yuvraj, Liang, Alao, Rodriguez, Ayub, Murthy, Nayani, Mitra, Parthasarathy, Li, Hogan, Battey, Wang, Howes, Rinott, Mehta, Siby, Bondu, Datta, Chugh, Hunt, Dhillon, Sidorov, Pan, Mahajan, Verma, Yamamoto, Ramaswamy, Lindsay, Lindsay, Feng, Lin, Zha, Patil, Shankar, Zhang, Zhang, Wang, Agarwal, Sajuyigbe, Chintala, Max, Chen, Kehoe, Satterfield, Govindaprasad, Gupta, Deng, Cho, Virk, Subramanian, Choudhury,
  Goldman, Remez, Glaser, Best, Koehler, Robinson, Li, Zhang, Matthews, Chou, Shaked, Vontimitta, Ajayi, Montanez, Mohan, Kumar, Mangla, Ionescu, Poenaru, Mihailescu, Ivanov, Li, Wang, Jiang, Bouaziz, Constable, Tang, Wu, Wang, Wu, Gao, Kleinman, Chen, Hu, Jia, Qi, Li, Zhang, Zhang, Adi, Nam, Yu, Wang, Zhao, Hao, Qian, Li, He, Rait, DeVito, Rosnbrick, Wen, Yang, Zhao, and Ma}]{grattafiori2024llama3herdmodels}
Aaron Grattafiori, Abhimanyu Dubey, Abhinav Jauhri, Abhinav Pandey, Abhishek Kadian, Ahmad Al-Dahle, Aiesha Letman, Akhil Mathur, Alan Schelten, Alex Vaughan, Amy Yang, Angela Fan, Anirudh Goyal, Anthony Hartshorn, Aobo Yang, Archi Mitra, Archie Sravankumar, Artem Korenev, Arthur Hinsvark, and 542 others. 2024.
\newblock \href {https://arxiv.org/abs/2407.21783} {The llama 3 herd of models}.
\newblock \emph{Preprint}, arXiv:2407.21783.

\bibitem[{Groeneveld et~al.(2024)Groeneveld, Beltagy, Walsh, Bhagia, Kinney, Tafjord, Jha, Ivison, Magnusson, Wang, Arora, Atkinson, Authur, Chandu, Cohan, Dumas, Elazar, Gu, Hessel, Khot, Merrill, Morrison, Muennighoff, Naik, Nam, Peters, Pyatkin, Ravichander, Schwenk, Shah, Smith, Strubell, Subramani, Wortsman, Dasigi, Lambert, Richardson, Zettlemoyer, Dodge, Lo, Soldaini, Smith, and Hajishirzi}]{groeneveld2024olmoacceleratingsciencelanguage}
Dirk Groeneveld, Iz~Beltagy, Pete Walsh, Akshita Bhagia, Rodney Kinney, Oyvind Tafjord, Ananya~Harsh Jha, Hamish Ivison, Ian Magnusson, Yizhong Wang, Shane Arora, David Atkinson, Russell Authur, Khyathi~Raghavi Chandu, Arman Cohan, Jennifer Dumas, Yanai Elazar, Yuling Gu, Jack Hessel, and 24 others. 2024.
\newblock \href {https://arxiv.org/abs/2402.00838} {Olmo: Accelerating the science of language models}.
\newblock \emph{Preprint}, arXiv:2402.00838.

\bibitem[{Gu et~al.(2024)Gu, Jiang, Shi, Tan, Zhai, Xu, Li, Shen, Ma, Liu, Wang, and Guo}]{gu2024surveyllmasajudge}
Jiawei Gu, Xuhui Jiang, Zhichao Shi, Hexiang Tan, Xuehao Zhai, Chengjin Xu, Wei Li, Yinghan Shen, Shengjie Ma, Honghao Liu, Yuanzhuo Wang, and Jian Guo. 2024.
\newblock A survey on llm-as-a-judge.
\newblock \emph{arXiv preprint arXiv: 2411.15594}.

\bibitem[{Gugger et~al.(2022)Gugger, Debut, Wolf, Schmid, Mueller, Mangrulkar, Sun, and Bossan}]{accelerate}
Sylvain Gugger, Lysandre Debut, Thomas Wolf, Philipp Schmid, Zachary Mueller, Sourab Mangrulkar, Marc Sun, and Benjamin Bossan. 2022.
\newblock Accelerate: Training and inference at scale made simple, efficient and adaptable.
\newblock \url{https://github.com/huggingface/accelerate}.

\bibitem[{Gunasekar et~al.(2023)Gunasekar, Zhang, Aneja, Mendes, Del~Giorno, Gopi, Javaheripi, Kauffmann, de~Rosa, Saarikivi et~al.}]{gunasekar2023textbooks}
Suriya Gunasekar, Yi~Zhang, Jyoti Aneja, Caio C{\'e}sar~Teodoro Mendes, Allie Del~Giorno, Sivakanth Gopi, Mojan Javaheripi, Piero Kauffmann, Gustavo de~Rosa, Olli Saarikivi, and 1 others. 2023.
\newblock Textbooks are all you need.
\newblock \emph{arXiv preprint arXiv:2306.11644}.

\bibitem[{Guu et~al.(2020)Guu, Lee, Tung, Pasupat, and Chang}]{pmlr-v119-guu20a}
Kelvin Guu, Kenton Lee, Zora Tung, Panupong Pasupat, and Mingwei Chang. 2020.
\newblock \href {https://proceedings.mlr.press/v119/guu20a.html} {Retrieval augmented language model pre-training}.
\newblock In \emph{Proceedings of the 37th International Conference on Machine Learning}, volume 119 of \emph{Proceedings of Machine Learning Research}, pages 3929--3938. PMLR.

\bibitem[{Han et~al.(2024)Han, Rao, Ettinger, Jiang, Lin, Lambert, Choi, and Dziri}]{NEURIPS2024_0f69b4b9}
Seungju Han, Kavel Rao, Allyson Ettinger, Liwei Jiang, Bill~Yuchen Lin, Nathan Lambert, Yejin Choi, and Nouha Dziri. 2024.
\newblock \href {https://proceedings.neurips.cc/paper_files/paper/2024/file/0f69b4b96a46f284b726fbd70f74fb3b-Paper-Datasets_and_Benchmarks_Track.pdf} {Wildguard: Open one-stop moderation tools for safety risks, jailbreaks, and refusals of llms}.
\newblock In \emph{Advances in Neural Information Processing Systems}, volume~37, pages 8093--8131. Curran Associates, Inc.

\bibitem[{Hsu et~al.(2025)Hsu, Dai, Kothapalli, Song, Tang, Zhu, Shimizu, Sahni, Ning, and Chen}]{hsu2025ligerkernelefficienttriton}
Pin-Lun Hsu, Yun Dai, Vignesh Kothapalli, Qingquan Song, Shao Tang, Siyu Zhu, Steven Shimizu, Shivam Sahni, Haowen Ning, and Yanning Chen. 2025.
\newblock \href {https://arxiv.org/abs/2410.10989} {Liger kernel: Efficient triton kernels for llm training}.
\newblock \emph{Preprint}, arXiv:2410.10989.

\bibitem[{Izacard et~al.(2023)Izacard, Lewis, Lomeli, Hosseini, Petroni, Schick, Dwivedi-Yu, Joulin, Riedel, and Grave}]{JMLR:v24:23-0037}
Gautier Izacard, Patrick Lewis, Maria Lomeli, Lucas Hosseini, Fabio Petroni, Timo Schick, Jane Dwivedi-Yu, Armand Joulin, Sebastian Riedel, and Edouard Grave. 2023.
\newblock \href {http://jmlr.org/papers/v24/23-0037.html} {Atlas: Few-shot learning with retrieval augmented language models}.
\newblock \emph{Journal of Machine Learning Research}, 24(251):1--43.

\bibitem[{Jiang et~al.(2023)Jiang, Sablayrolles, Mensch, Bamford, Chaplot, de~Las~Casas, Bressand, Lengyel, Lample, Saulnier, Lavaud, Lachaux, Stock, Scao, Lavril, Wang, Lacroix, and Sayed}]{Jiang2023Mistral7}
Albert~Qiaochu Jiang, Alexandre Sablayrolles, Arthur Mensch, Chris Bamford, Devendra~Singh Chaplot, Diego de~Las~Casas, Florian Bressand, Gianna Lengyel, Guillaume Lample, Lucile Saulnier, L'elio~Renard Lavaud, Marie-Anne Lachaux, Pierre Stock, Teven~Le Scao, Thibaut Lavril, Thomas Wang, Timoth{\'e}e Lacroix, and William~El Sayed. 2023.
\newblock \href {https://api.semanticscholar.org/CorpusID:263830494} {Mistral 7b}.
\newblock \emph{ArXiv}, abs/2310.06825.

\bibitem[{Jin et~al.(2024)Jin, Yuan, Men, Cao, Chen, Liu, and Zhao}]{jin2024ragrewardbenchbenchmarkingrewardmodels}
Zhuoran Jin, Hongbang Yuan, Tianyi Men, Pengfei Cao, Yubo Chen, Kang Liu, and Jun Zhao. 2024.
\newblock \href {https://arxiv.org/abs/2412.13746} {Rag-rewardbench: Benchmarking reward models in retrieval augmented generation for preference alignment}.
\newblock \emph{Preprint}, arXiv:2412.13746.

\bibitem[{Kalajdzievski(2023)}]{kalajdzievski2023rankstabilizationscalingfactor}
Damjan Kalajdzievski. 2023.
\newblock \href {https://arxiv.org/abs/2312.03732} {A rank stabilization scaling factor for fine-tuning with lora}.
\newblock \emph{Preprint}, arXiv:2312.03732.

\bibitem[{Kim et~al.(2024{\natexlab{a}})Kim, Shin, Cho, Jang, Longpre, Lee, Yun, Shin, Kim, Thorne, and Seo}]{kim2024prometheusinducingfinegrainedevaluation}
Seungone Kim, Jamin Shin, Yejin Cho, Joel Jang, Shayne Longpre, Hwaran Lee, Sangdoo Yun, Seongjin Shin, Sungdong Kim, James Thorne, and Minjoon Seo. 2024{\natexlab{a}}.
\newblock \href {https://arxiv.org/abs/2310.08491} {Prometheus: Inducing fine-grained evaluation capability in language models}.
\newblock \emph{Preprint}, arXiv:2310.08491.

\bibitem[{Kim et~al.(2024{\natexlab{b}})Kim, Suk, Longpre, Lin, Shin, Welleck, Neubig, Lee, Lee, and Seo}]{kim-etal-2024-prometheus}
Seungone Kim, Juyoung Suk, Shayne Longpre, Bill~Yuchen Lin, Jamin Shin, Sean Welleck, Graham Neubig, Moontae Lee, Kyungjae Lee, and Minjoon Seo. 2024{\natexlab{b}}.
\newblock \href {https://doi.org/10.18653/v1/2024.emnlp-main.248} {Prometheus 2: An open source language model specialized in evaluating other language models}.
\newblock In \emph{Proceedings of the 2024 Conference on Empirical Methods in Natural Language Processing}, pages 4334--4353, Miami, Florida, USA. Association for Computational Linguistics.

\bibitem[{Kwon et~al.(2023)Kwon, Li, Zhuang, Sheng, Zheng, Yu, Gonzalez, Zhang, and Stoica}]{kwon2023efficient}
Woosuk Kwon, Zhuohan Li, Siyuan Zhuang, Ying Sheng, Lianmin Zheng, Cody~Hao Yu, Joseph~E. Gonzalez, Hao Zhang, and Ion Stoica. 2023.
\newblock Efficient memory management for large language model serving with pagedattention.
\newblock In \emph{Proceedings of the ACM SIGOPS 29th Symposium on Operating Systems Principles}.

\bibitem[{Lambert et~al.(2025)Lambert, Pyatkin, Morrison, Miranda, Lin, Chandu, Dziri, Kumar, Zick, Choi, Smith, and Hajishirzi}]{lambert-etal-2025-rewardbench}
Nathan Lambert, Valentina Pyatkin, Jacob Morrison, Lester James~Validad Miranda, Bill~Yuchen Lin, Khyathi Chandu, Nouha Dziri, Sachin Kumar, Tom Zick, Yejin Choi, Noah~A. Smith, and Hannaneh Hajishirzi. 2025.
\newblock \href {https://aclanthology.org/2025.findings-naacl.96/} {{R}eward{B}ench: Evaluating reward models for language modeling}.
\newblock In \emph{Findings of the Association for Computational Linguistics: NAACL 2025}, pages 1755--1797, Albuquerque, New Mexico. Association for Computational Linguistics.

\bibitem[{Lewis et~al.(2020)Lewis, Perez, Piktus, Petroni, Karpukhin, Goyal, K\"{u}ttler, Lewis, Yih, Rockt\"{a}schel, Riedel, and Kiela}]{NEURIPS2020_6b493230}
Patrick Lewis, Ethan Perez, Aleksandra Piktus, Fabio Petroni, Vladimir Karpukhin, Naman Goyal, Heinrich K\"{u}ttler, Mike Lewis, Wen-tau Yih, Tim Rockt\"{a}schel, Sebastian Riedel, and Douwe Kiela. 2020.
\newblock \href {https://proceedings.neurips.cc/paper_files/paper/2020/file/6b493230205f780e1bc26945df7481e5-Paper.pdf} {Retrieval-augmented generation for knowledge-intensive nlp tasks}.
\newblock In \emph{Advances in Neural Information Processing Systems}, volume~33, pages 9459--9474. Curran Associates, Inc.

\bibitem[{Lhoest et~al.(2021)Lhoest, del Moral, Jernite, Thakur, von Platen, Patil, Chaumond, Drame, Plu, Tunstall, Davison, Šaško, Chhablani, Malik, Brandeis, Scao, Sanh, Xu, Patry, McMillan-Major, Schmid, Gugger, Delangue, Matussière, Debut, Bekman, Cistac, Goehringer, Mustar, Lagunas, Rush, and Wolf}]{lhoest2021datasetscommunitylibrarynatural}
Quentin Lhoest, Albert~Villanova del Moral, Yacine Jernite, Abhishek Thakur, Patrick von Platen, Suraj Patil, Julien Chaumond, Mariama Drame, Julien Plu, Lewis Tunstall, Joe Davison, Mario Šaško, Gunjan Chhablani, Bhavitvya Malik, Simon Brandeis, Teven~Le Scao, Victor Sanh, Canwen Xu, Nicolas Patry, and 13 others. 2021.
\newblock \href {https://arxiv.org/abs/2109.02846} {Datasets: A community library for natural language processing}.
\newblock \emph{Preprint}, arXiv:2109.02846.

\bibitem[{Li et~al.(2024{\natexlab{a}})Li, Jiang, Huang, Beigi, Zhao, Tan, Bhattacharjee, Jiang, Chen, Wu, Shu, Cheng, and Liu}]{li2024llmasajudge}
Dawei Li, Bohan Jiang, Liangjie Huang, Alimohammad Beigi, Chengshuai Zhao, Zhen Tan, Amrita Bhattacharjee, Yuxuan Jiang, Canyu Chen, Tianhao Wu, Kai Shu, Lu~Cheng, and Huan Liu. 2024{\natexlab{a}}.
\newblock From generation to judgment: Opportunities and challenges of llm-as-a-judge.
\newblock \emph{arXiv preprint arXiv: 2411.16594}.

\bibitem[{Li et~al.(2024{\natexlab{b}})Li, Xu, Shen, Xu, Gu, Lai, Tao, and Ma}]{li-etal-2024-leveraging-large}
Zhen Li, Xiaohan Xu, Tao Shen, Can Xu, Jia-Chen Gu, Yuxuan Lai, Chongyang Tao, and Shuai Ma. 2024{\natexlab{b}}.
\newblock \href {https://doi.org/10.18653/v1/2024.emnlp-main.896} {Leveraging large language models for {NLG} evaluation: Advances and challenges}.
\newblock In \emph{Proceedings of the 2024 Conference on Empirical Methods in Natural Language Processing}, pages 16028--16045, Miami, Florida, USA. Association for Computational Linguistics.

\bibitem[{Lin(2004)}]{lin-2004-rouge}
Chin-Yew Lin. 2004.
\newblock \href {https://aclanthology.org/W04-1013/} {{ROUGE}: A package for automatic evaluation of summaries}.
\newblock In \emph{Text Summarization Branches Out}, pages 74--81, Barcelona, Spain. Association for Computational Linguistics.

\bibitem[{Liu et~al.(2024)Liu, Zeng, Liu, Yan, He, Wang, Yan, Liu, and Zhou}]{liu2024skywork}
Chris~Yuhao Liu, Liang Zeng, Jiacai Liu, Rui Yan, Jujie He, Chaojie Wang, Shuicheng Yan, Yang Liu, and Yahui Zhou. 2024.
\newblock Skywork-reward: Bag of tricks for reward modeling in llms.
\newblock \emph{arXiv preprint arXiv:2410.18451}.

\bibitem[{Liu et~al.(2023)Liu, Iter, Xu, Wang, Xu, and Zhu}]{liu-etal-2023-g}
Yang Liu, Dan Iter, Yichong Xu, Shuohang Wang, Ruochen Xu, and Chenguang Zhu. 2023.
\newblock \href {https://doi.org/10.18653/v1/2023.emnlp-main.153} {{G}-eval: {NLG} evaluation using gpt-4 with better human alignment}.
\newblock In \emph{Proceedings of the 2023 Conference on Empirical Methods in Natural Language Processing}, pages 2511--2522, Singapore. Association for Computational Linguistics.

\bibitem[{Loshchilov and Hutter(2019)}]{loshchilov2019decoupledweightdecayregularization}
Ilya Loshchilov and Frank Hutter. 2019.
\newblock \href {https://arxiv.org/abs/1711.05101} {Decoupled weight decay regularization}.
\newblock \emph{Preprint}, arXiv:1711.05101.

\bibitem[{Mahan et~al.(2024)Mahan, Phung, Rafailov, Blagden, nathan lile, Castricato, Franken, Finn, and Albalak}]{Mahan2024GenerativeRM}
Dakota Mahan, Duy Phung, Rafael Rafailov, Chase Blagden, nathan lile, Louis Castricato, Jan-Philipp Franken, Chelsea Finn, and Alon Albalak. 2024.
\newblock \href {https://api.semanticscholar.org/CorpusID:273404003} {Generative reward models}.
\newblock \emph{ArXiv}, abs/2410.12832.

\bibitem[{Mangrulkar et~al.(2022)Mangrulkar, Gugger, Debut, Belkada, Paul, and Bossan}]{peft}
Sourab Mangrulkar, Sylvain Gugger, Lysandre Debut, Younes Belkada, Sayak Paul, and Benjamin Bossan. 2022.
\newblock Peft: State-of-the-art parameter-efficient fine-tuning methods.
\newblock \url{https://github.com/huggingface/peft}.

\bibitem[{MIT(1987)}]{MITLicense}
MIT. 1987.
\newblock \href {https://opensource.org/license/mit} {The mit license}.

\bibitem[{Muller et~al.(2025)Muller, Loison, Omrani, and Viaud}]{muller-etal-2025-grouse}
Sacha Muller, Antonio Loison, Bilel Omrani, and Gautier Viaud. 2025.
\newblock \href {https://aclanthology.org/2025.coling-main.304/} {{G}ro{USE}: A benchmark to evaluate evaluators in grounded question answering}.
\newblock In \emph{Proceedings of the 31st International Conference on Computational Linguistics}, pages 4510--4534, Abu Dhabi, UAE. Association for Computational Linguistics.

\bibitem[{Neyman(1934)}]{Neyman1934OnTT}
Jerzy Neyman. 1934.
\newblock \href {https://api.semanticscholar.org/CorpusID:86859017} {On the two different aspects of the representative method: the method of stratified sampling and the method of purposive selection}.
\newblock \emph{Journal of the Royal Statistical Society}, 97:123--150.

\bibitem[{OpenAI et~al.(2024)OpenAI, :, Jaech, Kalai, Lerer, Richardson, El-Kishky, Low, Helyar, Madry, Beutel, Carney, Iftimie, Karpenko, Passos, Neitz, Prokofiev, Wei, Tam, Bennett, Kumar, Saraiva, Vallone, Duberstein, Kondrich, Mishchenko, Applebaum, Jiang, Nair, Zoph, Ghorbani, Rossen, Sokolowsky, Barak, McGrew, Minaiev, Hao, Baker, Houghton, McKinzie, Eastman, Lugaresi, Bassin, Hudson, Li, de~Bourcy, Voss, Shen, Zhang, Koch, Orsinger, Hesse, Fischer, Chan, Roberts, Kappler, Levy, Selsam, Dohan, Farhi, Mely, Robinson, Tsipras, Li, Oprica, Freeman, Zhang, Wong, Proehl, Cheung, Mitchell, Wallace, Ritter, Mays, Wang, Such, Raso, Leoni, Tsimpourlas, Song, von Lohmann, Sulit, Salmon, Parascandolo, Chabot, Zhao, Brockman, Leclerc, Salman, Bao, Sheng, Andrin, Bagherinezhad, Ren, Lightman, Chung, Kivlichan, O'Connell, Osband, Gilaberte, Akkaya, Kostrikov, Sutskever, Kofman, Pachocki, Lennon, Wei, Harb, Twore, Feng, Yu, Weng, Tang, Yu, Candela, Palermo, Parish, Heidecke, Hallman, Rizzo, Gordon, Uesato, Ward,
  Huizinga, Wang, Chen, Xiao, Singhal, Nguyen, Cobbe, Shi, Wood, Rimbach, Gu-Lemberg, Liu, Lu, Stone, Yu, Ahmad, Yang, Liu, Maksin, Ho, Fedus, Weng, Li, McCallum, Held, Kuhn, Kondraciuk, Kaiser, Metz, Boyd, Trebacz, Joglekar, Chen, Tintor, Meyer, Jones, Kaufer, Schwarzer, Shah, Yatbaz, Guan, Xu, Yan, Glaese, Chen, Lampe, Malek, Wang, Fradin, McClay, Pavlov, Wang, Wang, Murati, Bavarian, Rohaninejad, McAleese, Chowdhury, Chowdhury, Ryder, Tezak, Brown, Nachum, Boiko, Murk, Watkins, Chao, Ashbourne, Izmailov, Zhokhov, Dias, Arora, Lin, Lopes, Gaon, Miyara, Leike, Hwang, Garg, Brown, James, Shu, Cheu, Greene, Jain, Altman, Toizer, Toyer, Miserendino, Agarwal, Hernandez, Baker, McKinney, Yan, Zhao, Hu, Santurkar, Chaudhuri, Zhang, Fu, Papay, Lin, Balaji, Sanjeev, Sidor, Broda, Clark, Wang, Gordon, Sanders, Patwardhan, Sottiaux, Degry, Dimson, Zheng, Garipov, Stasi, Bansal, Creech, Peterson, Eloundou, Qi, Kosaraju, Monaco, Pong, Fomenko, Zheng, Zhou, McCabe, Zaremba, Dubois, Lu, Chen, Cha, Bai, He, Zhang, Wang,
  Shao, and Li}]{openai2024openaio1card}
OpenAI, :, Aaron Jaech, Adam Kalai, Adam Lerer, Adam Richardson, Ahmed El-Kishky, Aiden Low, Alec Helyar, Aleksander Madry, Alex Beutel, Alex Carney, Alex Iftimie, Alex Karpenko, Alex~Tachard Passos, Alexander Neitz, Alexander Prokofiev, Alexander Wei, Allison Tam, and 244 others. 2024.
\newblock \href {https://arxiv.org/abs/2412.16720} {Openai o1 system card}.
\newblock \emph{Preprint}, arXiv:2412.16720.

\bibitem[{Ouyang et~al.(2022)Ouyang, Wu, Jiang, Almeida, Wainwright, Mishkin, Zhang, Agarwal, Slama, Ray, Schulman, Hilton, Kelton, Miller, Simens, Askell, Welinder, Christiano, Leike, and Lowe}]{NEURIPS2022_b1efde53}
Long Ouyang, Jeffrey Wu, Xu~Jiang, Diogo Almeida, Carroll Wainwright, Pamela Mishkin, Chong Zhang, Sandhini Agarwal, Katarina Slama, Alex Ray, John Schulman, Jacob Hilton, Fraser Kelton, Luke Miller, Maddie Simens, Amanda Askell, Peter Welinder, Paul~F Christiano, Jan Leike, and Ryan Lowe. 2022.
\newblock \href {https://proceedings.neurips.cc/paper_files/paper/2022/file/b1efde53be364a73914f58805a001731-Paper-Conference.pdf} {Training language models to follow instructions with human feedback}.
\newblock In \emph{Advances in Neural Information Processing Systems}, volume~35, pages 27730--27744. Curran Associates, Inc.

\bibitem[{Papineni et~al.(2002)Papineni, Roukos, Ward, and Zhu}]{10.3115/1073083.1073135}
Kishore Papineni, Salim Roukos, Todd Ward, and Wei-Jing Zhu. 2002.
\newblock \href {https://doi.org/10.3115/1073083.1073135} {Bleu: a method for automatic evaluation of machine translation}.
\newblock In \emph{Proceedings of the 40th Annual Meeting on Association for Computational Linguistics}, ACL '02, page 311–318, USA. Association for Computational Linguistics.

\bibitem[{Park et~al.(2024)Park, Jwa, Meiying, Kim, and Choi}]{park-etal-2024-offsetbias}
Junsoo Park, Seungyeon Jwa, Ren Meiying, Daeyoung Kim, and Sanghyuk Choi. 2024.
\newblock \href {https://doi.org/10.18653/v1/2024.findings-emnlp.57} {{O}ffset{B}ias: Leveraging debiased data for tuning evaluators}.
\newblock In \emph{Findings of the Association for Computational Linguistics: EMNLP 2024}, pages 1043--1067, Miami, Florida, USA. Association for Computational Linguistics.

\bibitem[{Qwen et~al.(2025)Qwen, :, Yang, Yang, Zhang, Hui, Zheng, Yu, Li, Liu, Huang, Wei, Lin, Yang, Tu, Zhang, Yang, Yang, Zhou, Lin, Dang, Lu, Bao, Yang, Yu, Li, Xue, Zhang, Zhu, Men, Lin, Li, Tang, Xia, Ren, Ren, Fan, Su, Zhang, Wan, Liu, Cui, Zhang, and Qiu}]{qwen2025qwen25technicalreport}
Qwen, :, An~Yang, Baosong Yang, Beichen Zhang, Binyuan Hui, Bo~Zheng, Bowen Yu, Chengyuan Li, Dayiheng Liu, Fei Huang, Haoran Wei, Huan Lin, Jian Yang, Jianhong Tu, Jianwei Zhang, Jianxin Yang, Jiaxi Yang, Jingren Zhou, and 25 others. 2025.
\newblock \href {https://arxiv.org/abs/2412.15115} {Qwen2.5 technical report}.
\newblock \emph{Preprint}, arXiv:2412.15115.

\bibitem[{Rafailov et~al.(2023)Rafailov, Sharma, Mitchell, Manning, Ermon, and Finn}]{NEURIPS2023_a85b405e}
Rafael Rafailov, Archit Sharma, Eric Mitchell, Christopher~D Manning, Stefano Ermon, and Chelsea Finn. 2023.
\newblock \href {https://proceedings.neurips.cc/paper_files/paper/2023/file/a85b405ed65c6477a4fe8302b5e06ce7-Paper-Conference.pdf} {Direct preference optimization: Your language model is secretly a reward model}.
\newblock In \emph{Advances in Neural Information Processing Systems}, volume~36, pages 53728--53741. Curran Associates, Inc.

\bibitem[{Saha et~al.(2025)Saha, Li, Ghazvininejad, Weston, and Wang}]{saha2025learningplanreason}
Swarnadeep Saha, Xian Li, Marjan Ghazvininejad, Jason Weston, and Tianlu Wang. 2025.
\newblock \href {https://arxiv.org/abs/2501.18099} {Learning to plan \& reason for evaluation with thinking-llm-as-a-judge}.
\newblock \emph{Preprint}, arXiv:2501.18099.

\bibitem[{Sai et~al.(2022)Sai, Mohankumar, and Khapra}]{10.1145/3485766}
Ananya~B. Sai, Akash~Kumar Mohankumar, and Mitesh~M. Khapra. 2022.
\newblock \href {https://doi.org/10.1145/3485766} {A survey of evaluation metrics used for nlg systems}.
\newblock \emph{ACM Comput. Surv.}, 55(2).

\bibitem[{Schulman et~al.(2017)Schulman, Wolski, Dhariwal, Radford, and Klimov}]{DBLP:journals/corr/SchulmanWDRK17}
John Schulman, Filip Wolski, Prafulla Dhariwal, Alec Radford, and Oleg Klimov. 2017.
\newblock \href {https://arxiv.org/abs/1707.06347} {Proximal policy optimization algorithms}.
\newblock \emph{CoRR}, abs/1707.06347.

\bibitem[{Shao et~al.(2024)Shao, Wang, Zhu, Xu, Song, Bi, Zhang, Zhang, Li, Wu, and Guo}]{shao2024deepseekmathpushinglimitsmathematical}
Zhihong Shao, Peiyi Wang, Qihao Zhu, Runxin Xu, Junxiao Song, Xiao Bi, Haowei Zhang, Mingchuan Zhang, Y.~K. Li, Y.~Wu, and Daya Guo. 2024.
\newblock \href {https://arxiv.org/abs/2402.03300} {Deepseekmath: Pushing the limits of mathematical reasoning in open language models}.
\newblock \emph{Preprint}, arXiv:2402.03300.

\bibitem[{Song et~al.(2024)Song, Kim, and Iyyer}]{song-etal-2024-veriscore}
Yixiao Song, Yekyung Kim, and Mohit Iyyer. 2024.
\newblock \href {https://doi.org/10.18653/v1/2024.findings-emnlp.552} {{V}eri{S}core: Evaluating the factuality of verifiable claims in long-form text generation}.
\newblock In \emph{Findings of the Association for Computational Linguistics: EMNLP 2024}, pages 9447--9474, Miami, Florida, USA. Association for Computational Linguistics.

\bibitem[{Srivastava et~al.(2014)Srivastava, Hinton, Krizhevsky, Sutskever, and Salakhutdinov}]{JMLR:v15:srivastava14a}
Nitish Srivastava, Geoffrey Hinton, Alex Krizhevsky, Ilya Sutskever, and Ruslan Salakhutdinov. 2014.
\newblock \href {http://jmlr.org/papers/v15/srivastava14a.html} {Dropout: A simple way to prevent neural networks from overfitting}.
\newblock \emph{Journal of Machine Learning Research}, 15(56):1929--1958.

\bibitem[{Tang et~al.(2024)Tang, Laban, and Durrett}]{tang-etal-2024-minicheck}
Liyan Tang, Philippe Laban, and Greg Durrett. 2024.
\newblock \href {https://doi.org/10.18653/v1/2024.emnlp-main.499} {{M}ini{C}heck: Efficient fact-checking of {LLM}s on grounding documents}.
\newblock In \emph{Proceedings of the 2024 Conference on Empirical Methods in Natural Language Processing}, pages 8818--8847, Miami, Florida, USA. Association for Computational Linguistics.

\bibitem[{Team et~al.(2024)Team, Ormazabal, Zheng, de~Masson~d'Autume, Yogatama, Fu, Ong, Chen, Lamprecht, Pham, Ong, Aleksiev, Li, Henderson, Bain, Artetxe, Relan, Padlewski, Liu, Chen, Phua, Yang, Tay, Wang, Zhu, and Xie}]{rekateam2024rekacoreflashedge}
Reka Team, Aitor Ormazabal, Che Zheng, Cyprien de~Masson~d'Autume, Dani Yogatama, Deyu Fu, Donovan Ong, Eric Chen, Eugenie Lamprecht, Hai Pham, Isaac Ong, Kaloyan Aleksiev, Lei Li, Matthew Henderson, Max Bain, Mikel Artetxe, Nishant Relan, Piotr Padlewski, Qi~Liu, and 7 others. 2024.
\newblock \href {https://arxiv.org/abs/2404.12387} {Reka core, flash, and edge: A series of powerful multimodal language models}.
\newblock \emph{Preprint}, arXiv:2404.12387.

\bibitem[{Tonmoy et~al.(2024)Tonmoy, Zaman, Jain, Rani, Rawte, Chadha, and Das}]{tonmoy2024comprehensivesurveyhallucinationmitigation}
S.~M Towhidul~Islam Tonmoy, S~M~Mehedi Zaman, Vinija Jain, Anku Rani, Vipula Rawte, Aman Chadha, and Amitava Das. 2024.
\newblock \href {https://arxiv.org/abs/2401.01313} {A comprehensive survey of hallucination mitigation techniques in large language models}.
\newblock \emph{Preprint}, arXiv:2401.01313.

\bibitem[{TREC-RAG(2024)}]{TRECRAG}
TREC-RAG. 2024.
\newblock \href {https://trec-rag.github.io/} {Trec retrieval-augmented generation}.

\bibitem[{von Werra et~al.(2020)von Werra, Belkada, Tunstall, Beeching, Thrush, Lambert, Huang, Rasul, and Gallouédec}]{vonwerra2022trl}
Leandro von Werra, Younes Belkada, Lewis Tunstall, Edward Beeching, Tristan Thrush, Nathan Lambert, Shengyi Huang, Kashif Rasul, and Quentin Gallouédec. 2020.
\newblock Trl: Transformer reinforcement learning.
\newblock \url{https://github.com/huggingface/trl}.

\bibitem[{Vu et~al.(2024)Vu, Krishna, Alzubi, Tar, Faruqui, and Sung}]{vu-etal-2024-foundational}
Tu~Vu, Kalpesh Krishna, Salaheddin Alzubi, Chris Tar, Manaal Faruqui, and Yun-Hsuan Sung. 2024.
\newblock \href {https://doi.org/10.18653/v1/2024.emnlp-main.949} {Foundational autoraters: Taming large language models for better automatic evaluation}.
\newblock In \emph{Proceedings of the 2024 Conference on Empirical Methods in Natural Language Processing}, pages 17086--17105, Miami, Florida, USA. Association for Computational Linguistics.

\bibitem[{Wang et~al.(2024{\natexlab{a}})Wang, Xu, Zhou, Xiong, and Joty}]{wang2024directjudgementpreferenceoptimization}
Peifeng Wang, Austin Xu, Yilun Zhou, Caiming Xiong, and Shafiq Joty. 2024{\natexlab{a}}.
\newblock \href {https://arxiv.org/abs/2409.14664} {Direct judgement preference optimization}.
\newblock \emph{Preprint}, arXiv:2409.14664.

\bibitem[{Wang et~al.(2024{\natexlab{b}})Wang, Li, Chen, Cai, Zhu, Lin, Cao, Kong, Liu, Liu, and Sui}]{wang-etal-2024-large-language-models-fair}
Peiyi Wang, Lei Li, Liang Chen, Zefan Cai, Dawei Zhu, Binghuai Lin, Yunbo Cao, Lingpeng Kong, Qi~Liu, Tianyu Liu, and Zhifang Sui. 2024{\natexlab{b}}.
\newblock \href {https://doi.org/10.18653/v1/2024.acl-long.511} {Large language models are not fair evaluators}.
\newblock In \emph{Proceedings of the 62nd Annual Meeting of the Association for Computational Linguistics (Volume 1: Long Papers)}, pages 9440--9450, Bangkok, Thailand. Association for Computational Linguistics.

\bibitem[{Wang et~al.(2024{\natexlab{c}})Wang, Kulikov, Golovneva, Yu, Yuan, Dwivedi-Yu, Pang, Fazel-Zarandi, Weston, and Li}]{wang2024selftaughtevaluators}
Tianlu Wang, Ilia Kulikov, Olga Golovneva, Ping Yu, Weizhe Yuan, Jane Dwivedi-Yu, Richard~Yuanzhe Pang, Maryam Fazel-Zarandi, Jason Weston, and Xian Li. 2024{\natexlab{c}}.
\newblock \href {https://arxiv.org/abs/2408.02666} {Self-taught evaluators}.
\newblock \emph{Preprint}, arXiv:2408.02666.

\bibitem[{Wang et~al.(2024{\natexlab{d}})Wang, Yu, Zeng, Yang, Wang, Chen, Jiang, Xie, Wang, Xie, Ye, Zhang, and Zhang}]{wang2024pandalmautomaticevaluationbenchmark}
Yidong Wang, Zhuohao Yu, Zhengran Zeng, Linyi Yang, Cunxiang Wang, Hao Chen, Chaoya Jiang, Rui Xie, Jindong Wang, Xing Xie, Wei Ye, Shikun Zhang, and Yue Zhang. 2024{\natexlab{d}}.
\newblock \href {https://arxiv.org/abs/2306.05087} {Pandalm: An automatic evaluation benchmark for llm instruction tuning optimization}.
\newblock \emph{Preprint}, arXiv:2306.05087.

\bibitem[{Wang et~al.(2024{\natexlab{e}})Wang, Dong, Delalleau, Zeng, Shen, Egert, Zhang, Sreedhar, and Kuchaiev}]{NEURIPS2024_02fd91a3}
Zhilin Wang, Yi~Dong, Olivier Delalleau, Jiaqi Zeng, Gerald Shen, Daniel Egert, Jimmy~J. Zhang, Makesh~Narsimhan Sreedhar, and Oleksii Kuchaiev. 2024{\natexlab{e}}.
\newblock \href {https://proceedings.neurips.cc/paper_files/paper/2024/file/02fd91a387a6a5a5751e81b58a75af90-Paper-Datasets_and_Benchmarks_Track.pdf} {Helpsteer 2: Open-source dataset for training top-performing reward models}.
\newblock In \emph{Advances in Neural Information Processing Systems}, volume~37, pages 1474--1501. Curran Associates, Inc.

\bibitem[{Wei et~al.(2023)Wei, Wang, Schuurmans, Bosma, Ichter, Xia, Chi, Le, and Zhou}]{wei2023chainofthoughtpromptingelicitsreasoning}
Jason Wei, Xuezhi Wang, Dale Schuurmans, Maarten Bosma, Brian Ichter, Fei Xia, Ed~Chi, Quoc Le, and Denny Zhou. 2023.
\newblock \href {https://arxiv.org/abs/2201.11903} {Chain-of-thought prompting elicits reasoning in large language models}.
\newblock \emph{Preprint}, arXiv:2201.11903.

\bibitem[{Xu et~al.(2025)Xu, Bansal, Ming, Yavuz, and Joty}]{xu2025doescontextmattercontextualjudgebench}
Austin Xu, Srijan Bansal, Yifei Ming, Semih Yavuz, and Shafiq Joty. 2025.
\newblock \href {https://arxiv.org/abs/2503.15620} {Does context matter? contextualjudgebench for evaluating llm-based judges in contextual settings}.
\newblock \emph{Preprint}, arXiv:2503.15620.

\bibitem[{Xu et~al.(2024)Xu, Jiang, Niu, Deng, Poovendran, Choi, and Lin}]{xu2024magpie}
Zhangchen Xu, Fengqing Jiang, Luyao Niu, Yuntian Deng, Radha Poovendran, Yejin Choi, and Bill~Yuchen Lin. 2024.
\newblock \href {https://api.semanticscholar.org/CorpusID:270391432} {Magpie: Alignment data synthesis from scratch by prompting aligned llms with nothing}.
\newblock \emph{ArXiv}, abs/2406.08464.

\bibitem[{Yang et~al.(2024)Yang, Ding, Lin, Zhang, and Zhang}]{Yang2024RegularizingHS}
Rui Yang, Ruomeng Ding, Yong Lin, Huan Zhang, and Tong Zhang. 2024.
\newblock \href {https://api.semanticscholar.org/CorpusID:270521260} {Regularizing hidden states enables learning generalizable reward model for llms}.
\newblock \emph{ArXiv}, abs/2406.10216.

\bibitem[{Ye et~al.(2024{\natexlab{a}})Ye, Wang, Huang, Chen, Zhang, Moniz, Gao, Geyer, Huang, Chen, Chawla, and Zhang}]{ye2024justiceprejudicequantifyingbiases}
Jiayi Ye, Yanbo Wang, Yue Huang, Dongping Chen, Qihui Zhang, Nuno Moniz, Tian Gao, Werner Geyer, Chao Huang, Pin-Yu Chen, Nitesh~V Chawla, and Xiangliang Zhang. 2024{\natexlab{a}}.
\newblock \href {https://arxiv.org/abs/2410.02736} {Justice or prejudice? quantifying biases in llm-as-a-judge}.
\newblock \emph{Preprint}, arXiv:2410.02736.

\bibitem[{Ye et~al.(2024{\natexlab{b}})Ye, Kim, Kim, Hwang, Kim, Jo, Thorne, Kim, and Seo}]{ye2024flaskfinegrainedlanguagemodel}
Seonghyeon Ye, Doyoung Kim, Sungdong Kim, Hyeonbin Hwang, Seungone Kim, Yongrae Jo, James Thorne, Juho Kim, and Minjoon Seo. 2024{\natexlab{b}}.
\newblock \href {https://arxiv.org/abs/2307.10928} {Flask: Fine-grained language model evaluation based on alignment skill sets}.
\newblock \emph{Preprint}, arXiv:2307.10928.

\bibitem[{Ye et~al.(2024{\natexlab{c}})Ye, Li, Li, Ai, Zhou, Shen, Yan, and Liu}]{ye2024scalarrewardmodellearning}
Ziyi Ye, Xiangsheng Li, Qiuchi Li, Qingyao Ai, Yujia Zhou, Wei Shen, Dong Yan, and Yiqun Liu. 2024{\natexlab{c}}.
\newblock \href {https://arxiv.org/abs/2410.03742} {Beyond scalar reward model: Learning generative judge from preference data}.
\newblock \emph{Preprint}, arXiv:2410.03742.

\bibitem[{Yu et~al.(2025)Yu, Chen, Zhang, Tan, Zhu, Pang, Qian, Wang, Gururangan, Zhang, Kambadur, Mahajan, and Hou}]{yu-etal-2025-self}
Yue Yu, Zhengxing Chen, Aston Zhang, Liang Tan, Chenguang Zhu, Richard~Yuanzhe Pang, Yundi Qian, Xuewei Wang, Suchin Gururangan, Chao Zhang, Melanie Kambadur, Dhruv Mahajan, and Rui Hou. 2025.
\newblock \href {https://aclanthology.org/2025.naacl-long.573/} {Self-generated critiques boost reward modeling for language models}.
\newblock In \emph{Proceedings of the 2025 Conference of the Nations of the Americas Chapter of the Association for Computational Linguistics: Human Language Technologies (Volume 1: Long Papers)}, pages 11499--11514, Albuquerque, New Mexico. Association for Computational Linguistics.

\bibitem[{Yuan et~al.(2021)Yuan, Neubig, and Liu}]{NEURIPS2021_e4d2b6e6}
Weizhe Yuan, Graham Neubig, and Pengfei Liu. 2021.
\newblock \href {https://proceedings.neurips.cc/paper_files/paper/2021/file/e4d2b6e6fdeca3e60e0f1a62fee3d9dd-Paper.pdf} {Bartscore: Evaluating generated text as text generation}.
\newblock In \emph{Advances in Neural Information Processing Systems}, volume~34, pages 27263--27277. Curran Associates, Inc.

\bibitem[{Zhang et~al.(2020)Zhang, Kishore, Wu, Weinberger, and Artzi}]{zhang2020bertscoreevaluatingtextgeneration}
Tianyi Zhang, Varsha Kishore, Felix Wu, Kilian~Q. Weinberger, and Yoav Artzi. 2020.
\newblock \href {https://arxiv.org/abs/1904.09675} {Bertscore: Evaluating text generation with bert}.
\newblock \emph{Preprint}, arXiv:1904.09675.

\bibitem[{Zheng et~al.(2023)Zheng, Chiang, Sheng, Zhuang, Wu, Zhuang, Lin, Li, Li, Xing, Zhang, Gonzalez, and Stoica}]{NEURIPS2023_91f18a12}
Lianmin Zheng, Wei-Lin Chiang, Ying Sheng, Siyuan Zhuang, Zhanghao Wu, Yonghao Zhuang, Zi~Lin, Zhuohan Li, Dacheng Li, Eric Xing, Hao Zhang, Joseph~E Gonzalez, and Ion Stoica. 2023.
\newblock \href {https://proceedings.neurips.cc/paper_files/paper/2023/file/91f18a1287b398d378ef22505bf41832-Paper-Datasets_and_Benchmarks.pdf} {Judging llm-as-a-judge with mt-bench and chatbot arena}.
\newblock In \emph{Advances in Neural Information Processing Systems}, volume~36, pages 46595--46623. Curran Associates, Inc.

\bibitem[{Zhong et~al.(2024)Zhong, Zhang, Wang, Hou, Xiong, Zhu, Chen, Tan, Bi, Lewis, Popuri, Narang, Kambadur, Mahajan, Edunov, Han, and van~der Maaten}]{zhong2024lawweakestlinkcross}
Ming Zhong, Aston Zhang, Xuewei Wang, Rui Hou, Wenhan Xiong, Chenguang Zhu, Zhengxing Chen, Liang Tan, Chloe Bi, Mike Lewis, Sravya Popuri, Sharan Narang, Melanie Kambadur, Dhruv Mahajan, Sergey Edunov, Jiawei Han, and Laurens van~der Maaten. 2024.
\newblock \href {https://arxiv.org/abs/2409.19951} {Law of the weakest link: Cross capabilities of large language models}.
\newblock \emph{Preprint}, arXiv:2409.19951.

\bibitem[{Zhu et~al.(2024)Zhu, Wang, Zhou, Wang, Chen, Wang, Yang, Ye, Zhang, Gong, and Xie}]{10.1145/3689217.3690621}
Kaijie Zhu, Jindong Wang, Jiaheng Zhou, Zichen Wang, Hao Chen, Yidong Wang, Linyi Yang, Wei Ye, Yue Zhang, Neil Gong, and Xing Xie. 2024.
\newblock \href {https://doi.org/10.1145/3689217.3690621} {Promptrobust: Towards evaluating the robustness of large language models on adversarial prompts}.
\newblock In \emph{Proceedings of the 1st ACM Workshop on Large AI Systems and Models with Privacy and Safety Analysis}, LAMPS '24, page 57–68, New York, NY, USA. Association for Computing Machinery.

\end{thebibliography}

\appendix
\section{Appendix}
\label{sec:appendix}

\subsection{Stratified sampling}
\label{subsec:stratified-sampling-details}

The sampling approach used in Subsection \ref{subsec:stratified-sampling} leverages the concept of a \textit{signature}, which is derived by concatenating key characteristics of each sample, such as query types and answer length categories. This signature acts as a unique identifier to group samples into strata that represent meaningful subpopulations within the dataset.

Once grouped, the signature strata are sorted in decreasing order by their size (the number of samples per signature). During the allocation of the fixed overall sampling budget, samples are selected by iterating through the signature groups starting with the least represented strata first. This ensures that smaller, potentially under-represented groups receive appropriate sampling allocation before larger groups are considered. Within each stratum, samples are stochastically selected without replacement, using random sampling guided by the allocated quota. This stochasticity maintains diversity and prevents over-representation of any single subset.

A key feature of this stratified sampling method is the use of a discounting factor applied to model-specific weights, which reflect the historical frequency of a model being selected as either \textit{chosen} or \textit{rejected}. Each time a model is selected, its associated weight corresponding to the \textit{chosen} or \textit{rejected} category is multiplicatively reduced by this discount factor, which is set to 0.9 in our experiments. This mechanism dynamically lowers the probability of repeatedly selecting the same models, thereby encouraging greater diversity and a more balanced representation of models within the sampled dataset.

Together, this stratification based on sample signatures and the adaptive, weighted stochastic sampling with discounting ensure the final sample set achieves broad coverage across sample characteristics and model choices, while mitigating biases due to dataset composition and prior model selection frequencies.

\subsection{Experimental Configurations}
\label{subsec:experimental-configurations}

We carried out data preprocessing and related tasks using the \texttt{datasets} library \citep{lhoest2021datasetscommunitylibrarynatural}. Fine-tuning of the Discriminative Reward Models (RMs) was performed using the \texttt{trl} \citep{vonwerra2022trl} and \texttt{accelerate} \citep{accelerate} libraries, with inference powered by the \texttt{vLLM} library \citep{kwon2023efficient}.

To optimise memory efficiency during fine-tuning, we employed Parameter-Efficient Fine-Tuning (PEFT) strategies from the \texttt{peft} library \citep{peft}, loading models quantised in 4-bit precision and integrating Flash Attention 2 \citep{dao2023flashattention2fasterattentionbetter} and Liger Kernels \citep{hsu2025ligerkernelefficienttriton}. Optimisation was conducted using AdamW \citep{loshchilov2019decoupledweightdecayregularization} in 4-bit precision. We adopted rsLoRA \citep{kalajdzievski2023rankstabilizationscalingfactor} and targeted all linear modules, setting the rank $r$ to 16, $\alpha$ to 16, and a dropout \citep{JMLR:v15:srivastava14a} rate of 0.1. Training was performed for 4 epochs with a learning rate of $2\text{e}^{-4}$, using a cosine learning rate scheduler with a warm-up ratio of 0.1. The maximum input length during training was 32,768 tokens, with a per-device batch size of 1 and gradient accumulation over 16 steps.  For inference using \texttt{vLLM}, we set the maximum input length to 8,192 tokens and used a temperature of 0.0.

All experiments we conducted on a single node equipped with 8xH200 NVIDIA GPUs. Query characteristics (Subsection \ref{subsec:query-characteristics}) accounted for 48 GPU hours. Answer generation and classification (Subsection \ref{subsec:answer-generation-and-classification}) used 192 and 256 GPU hours, respectively. Fine-tuning each \textit{RAGferee} model required between 3 and 8 GPU hours, depending on the model size.

The models from Hugging Face\footnote{Hugging Face: \url{https://huggingface.co/}} used for answer generation (Subsection \ref{subsec:answer-generation-and-classification}) are listed below:
{\small
\begin{itemize}
    \item \href{https://huggingface.co/microsoft/phi-4}{microsoft/phi-4}
    \item \href{https://huggingface.co/Qwen/QwQ-32B}{Qwen/QwQ-32B}
    \item \href{https://huggingface.co/RekaAI/reka-flash-3}{RekaAI/reka-flash-3}
    \item \href{https://huggingface.co/Qwen/Qwen2.5-7B-Instruct}{Qwen/Qwen2.5-7B-Instruct}
    \item \href{https://huggingface.co/Qwen/Qwen2.5-14B-Instruct}{Qwen/Qwen2.5-14B-Instruct}
    \item \href{https://huggingface.co/Qwen/Qwen2.5-32B-Instruct}{Qwen/Qwen2.5-32B-Instruct}
    \item \href{https://huggingface.co/internlm/internlm3-8b-instruct}{internlm/internlm3-8b-instruct}
    \item \href{https://huggingface.co/allenai/OLMo-2-1124-7B-Instruct}{allenai/OLMo-2-1124-7B-Instruct}
    \item \href{https://huggingface.co/allenai/OLMo-2-1124-13B-Instruct}{allenai/OLMo-2-1124-13B-Instruct}
    \item \href{https://huggingface.co/allenai/OLMo-2-0325-32B-Instruct}{allenai/OLMo-2-0325-32B-Instruct}
    \item \href{https://huggingface.co/mistralai/Mixtral-8x22B-Instruct-v0.1}{mistralai/Mixtral-8x22B-Instruct-v0.1}
    \item \href{https://huggingface.co/mistralai/Mistral-Nemo-Instruct-2407}{mistralai/Mistral-Nemo-Instruct-2407}
    \item \href{https://huggingface.co/mistralai/Mistral-Small-24B-Instruct-2501}{mistralai/Mistral-Small-24B-Instruct-2501}
    \item \href{https://huggingface.co/deepseek-ai/DeepSeek-R1-Distill-Qwen-14B}{deepseek-ai/DeepSeek-R1-Distill-Qwen-14B}
    \item \href{https://huggingface.co/deepseek-ai/DeepSeek-R1-Distill-Qwen-32B}{deepseek-ai/DeepSeek-R1-Distill-Qwen-32B}
\end{itemize}
}

\subsection{Distributions}
\label{subsec:distributions}

\begin{figure*}[htbp]
    \centering
    \begin{tabular}{ccc}
        \includegraphics[width=0.29\textwidth]{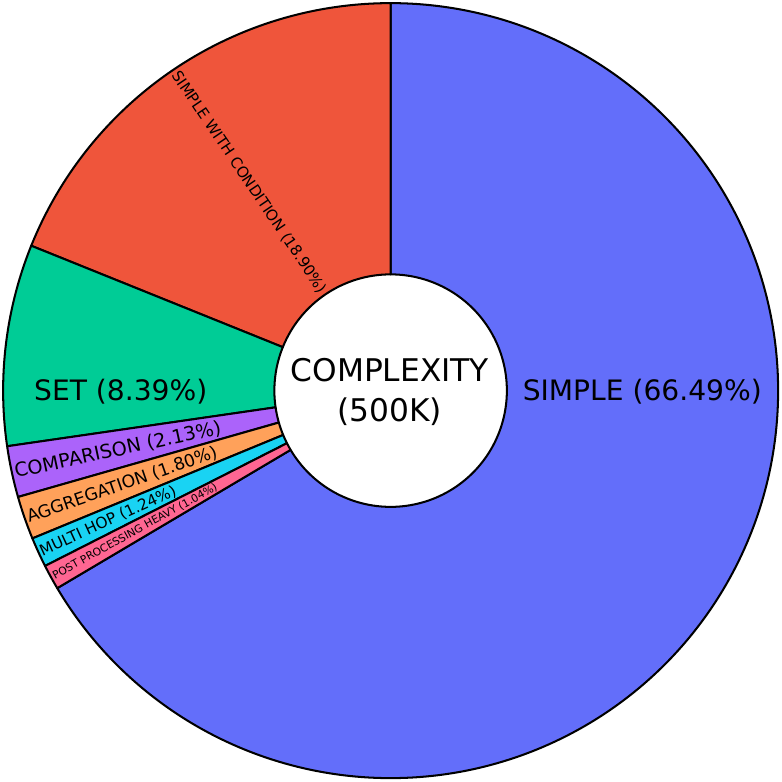} &
        \includegraphics[width=0.29\textwidth]{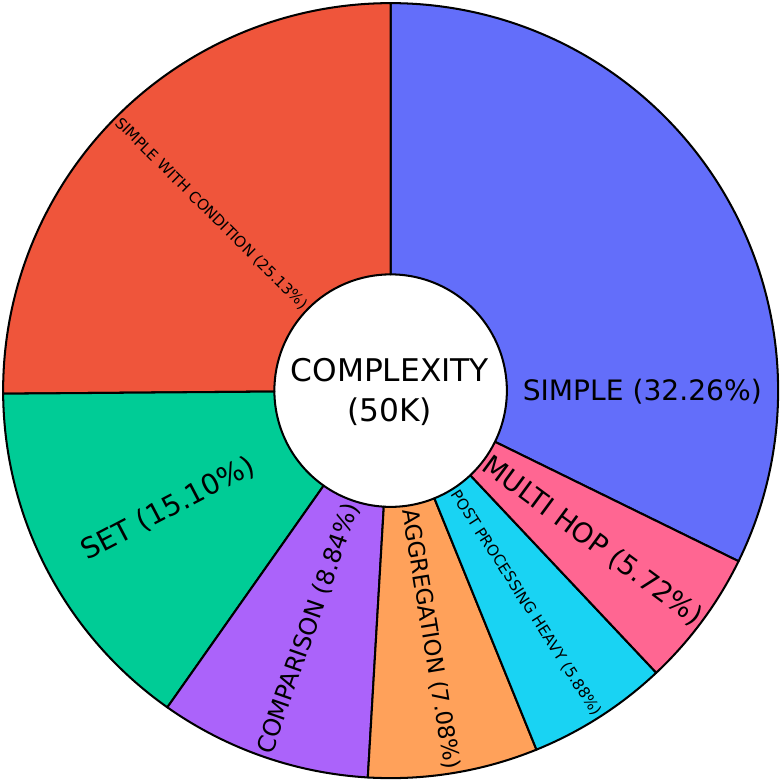} &
        \includegraphics[width=0.29\textwidth]{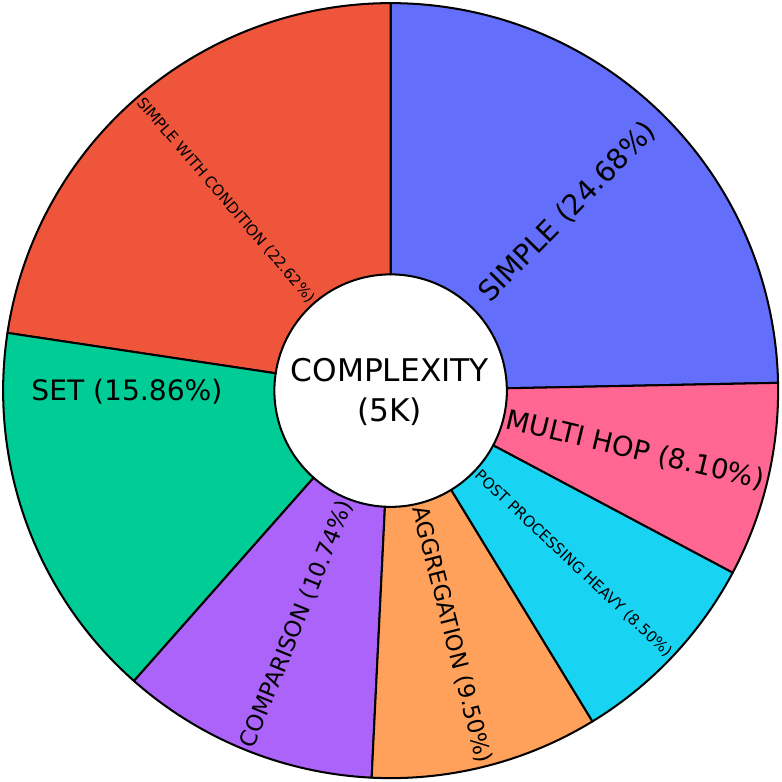} \\
        \includegraphics[width=0.29\textwidth]{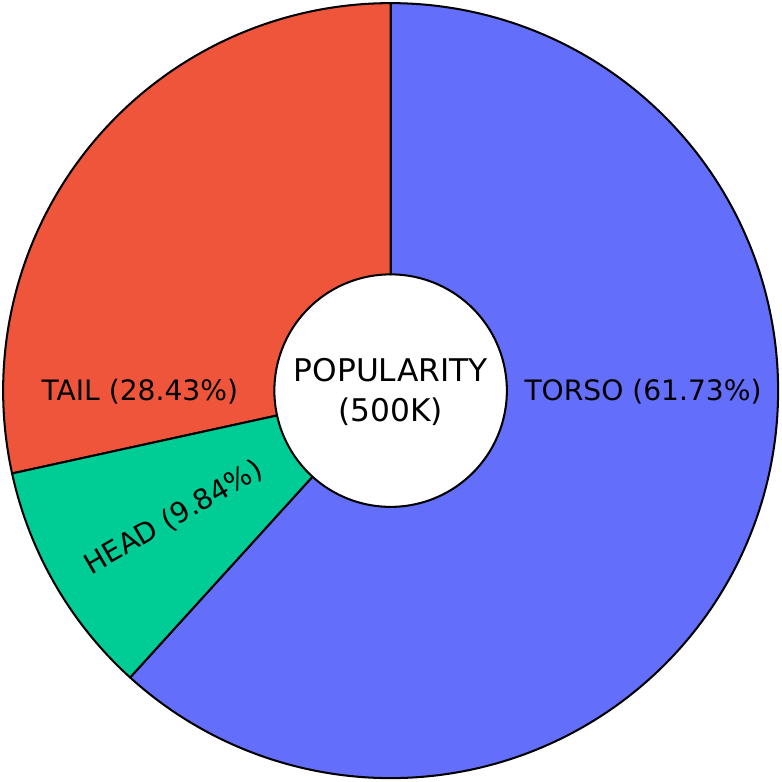} &
        \includegraphics[width=0.29\textwidth]{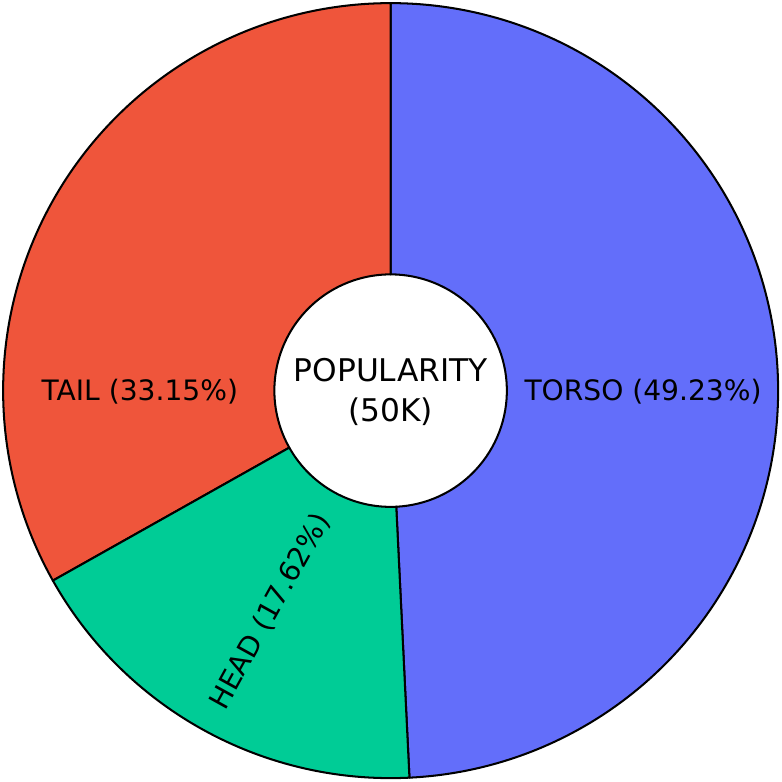} &
        \includegraphics[width=0.29\textwidth]{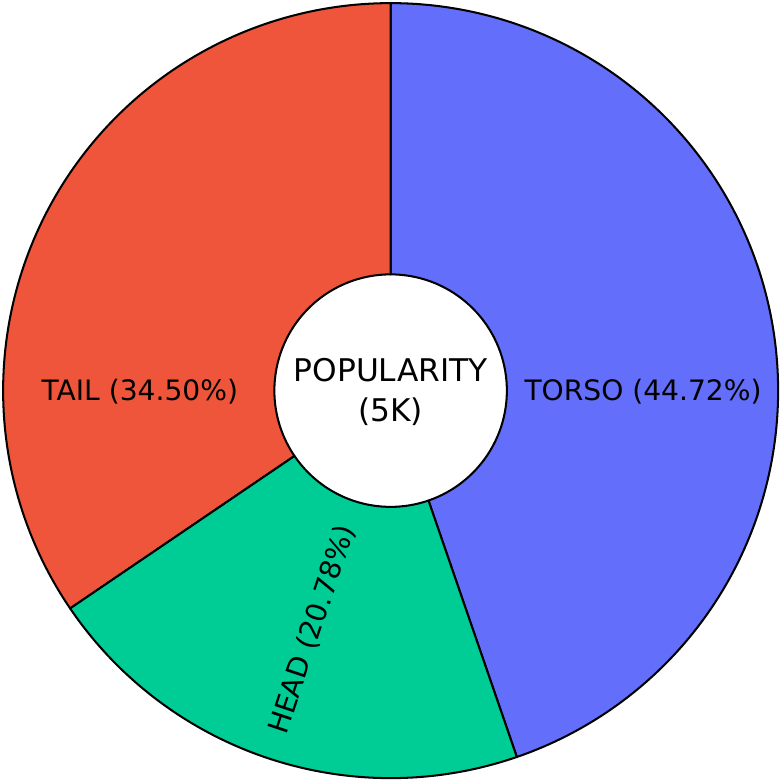} \\
        \includegraphics[width=0.29\textwidth]{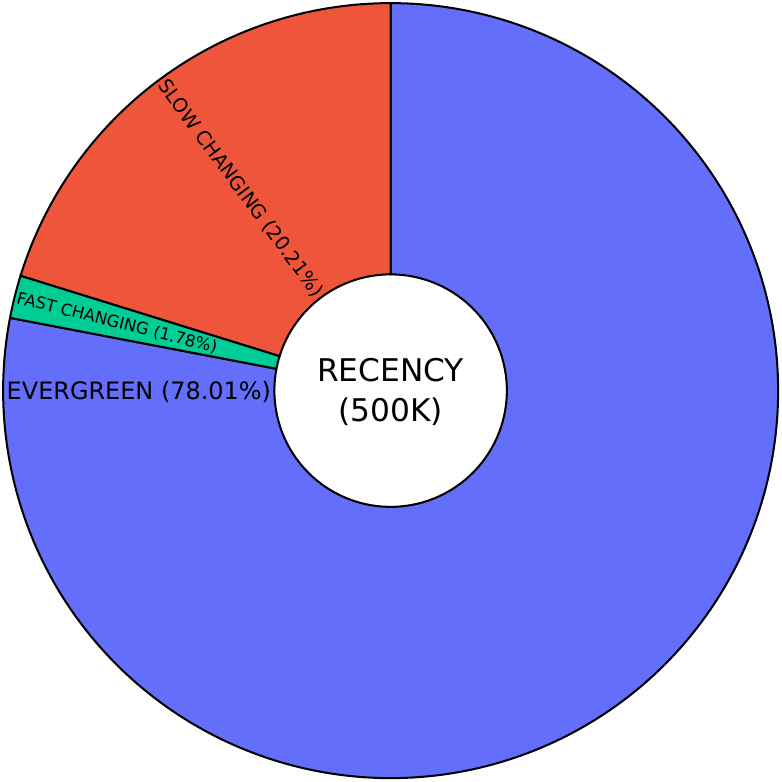} &
        \includegraphics[width=0.29\textwidth]{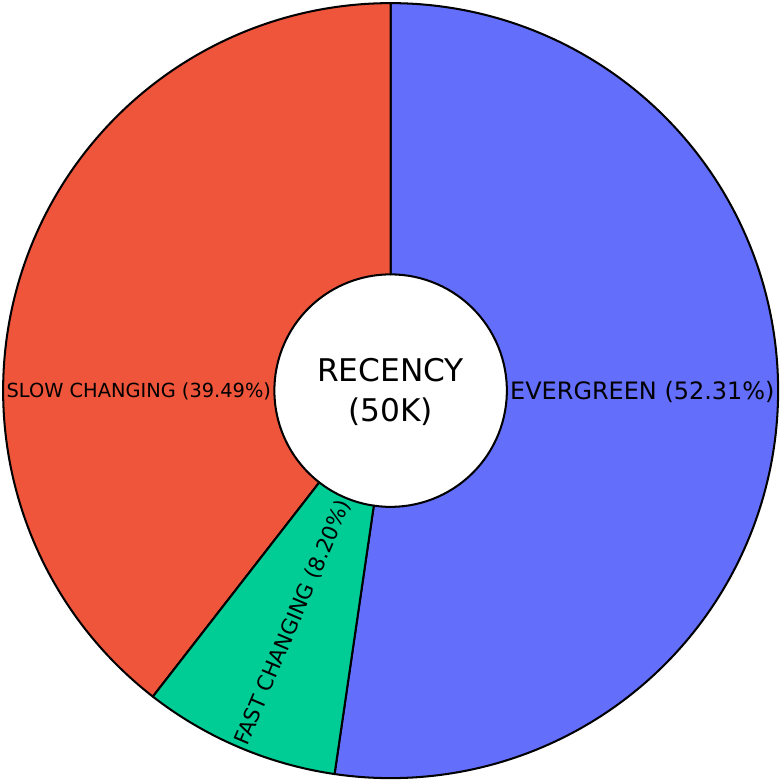} &
        \includegraphics[width=0.29\textwidth]{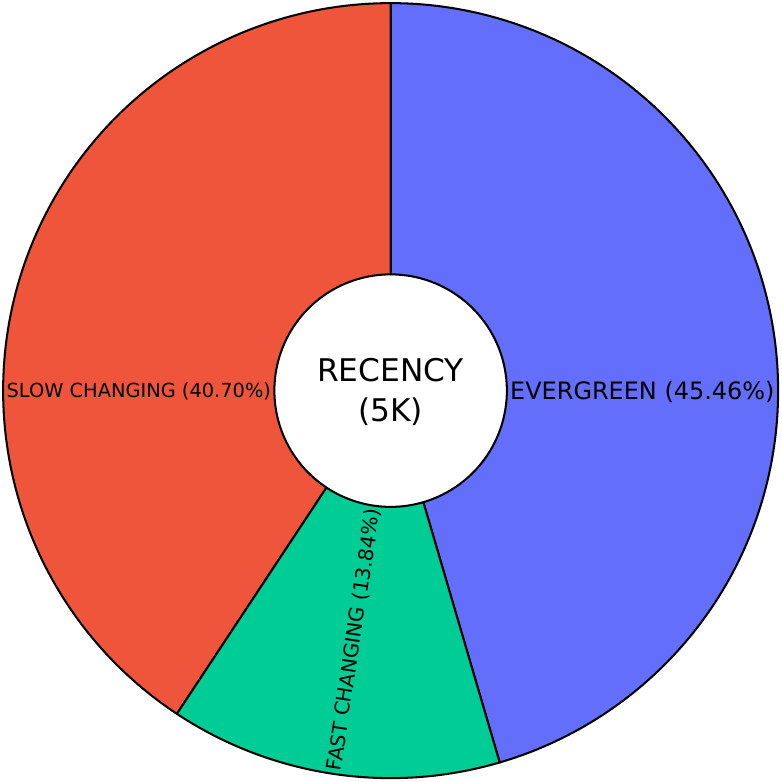} \\
        \includegraphics[width=0.29\textwidth]{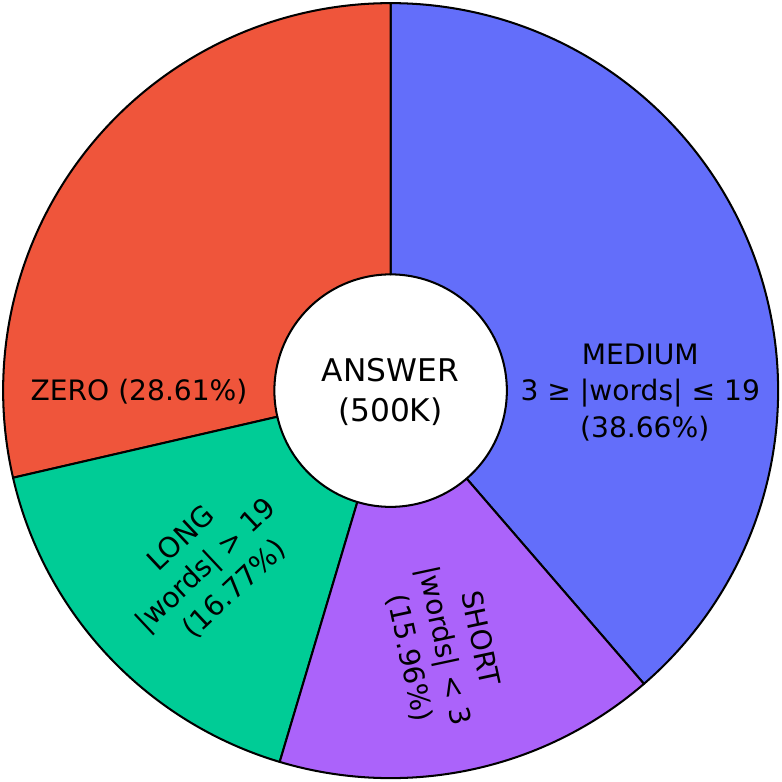} &
        \includegraphics[width=0.29\textwidth]{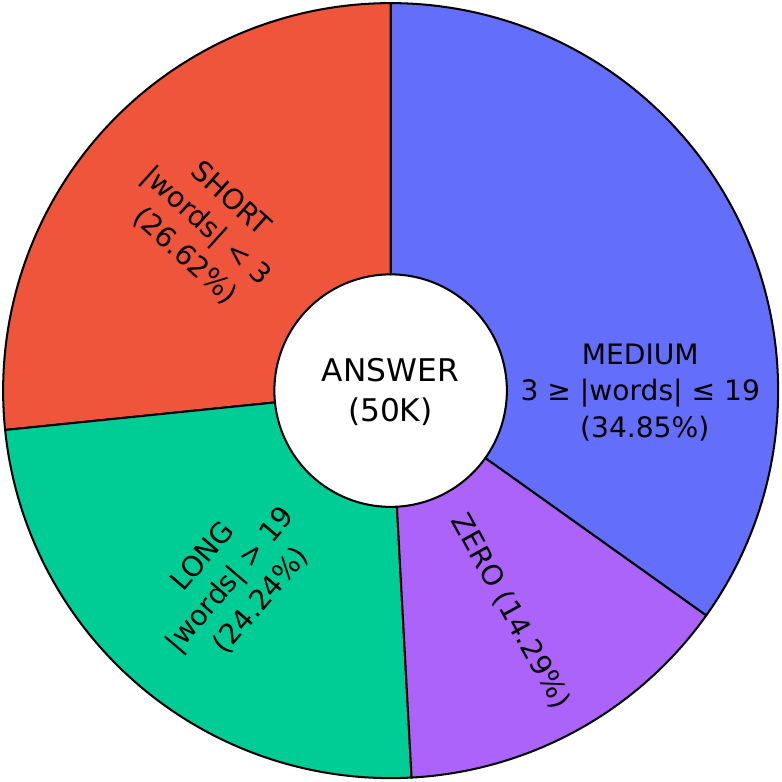} &
        \includegraphics[width=0.29\textwidth]{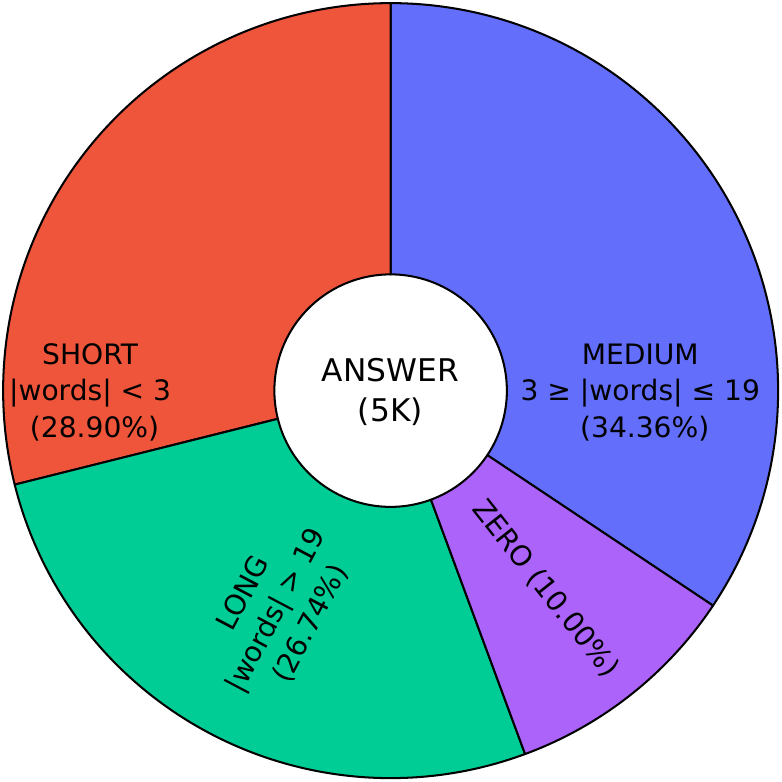} \\
        \includegraphics[width=0.29\textwidth]{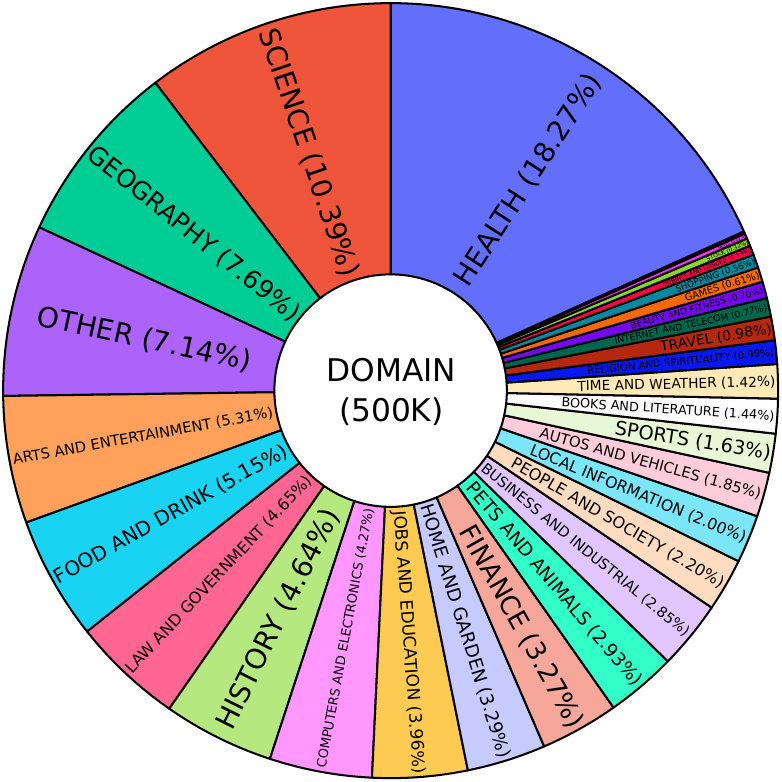} &
        \includegraphics[width=0.29\textwidth]{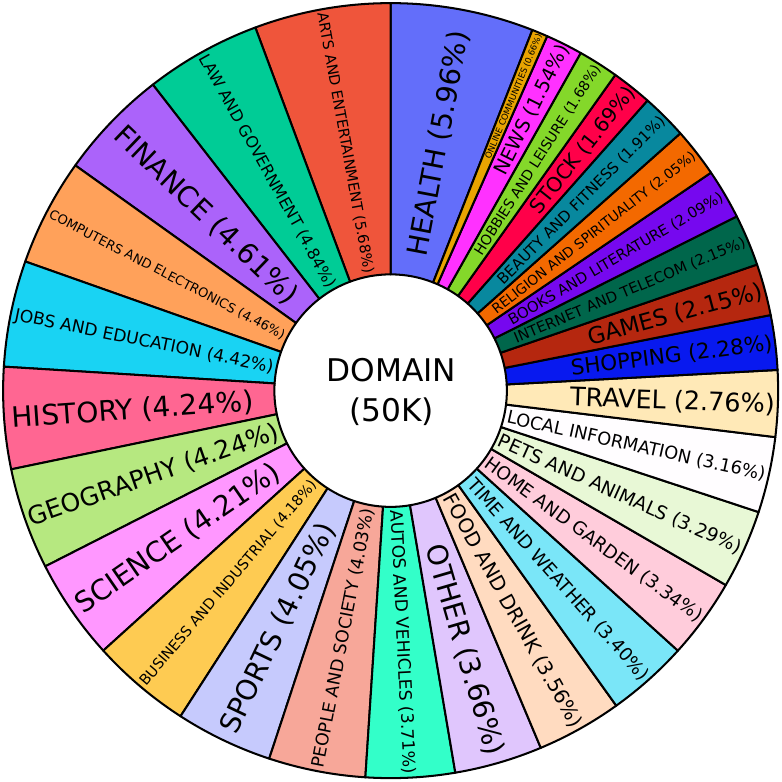} &
        \includegraphics[width=0.29\textwidth]{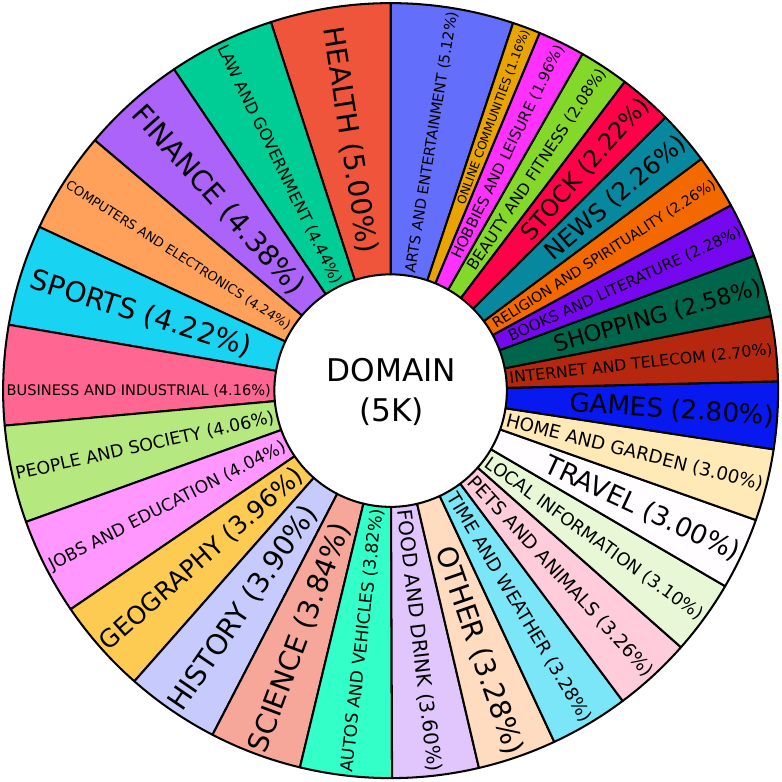} \\
    \end{tabular}
    \caption{Distributions of query characteristics (Subsection \ref{subsec:query-characteristics}) for the 500K, 50K, and 5K subsets.}
    \label{fig:query-characteristics-distributions}
\end{figure*}

\begin{figure*}[htbp]
    \centering
    \begin{minipage}{0.28\textwidth}
        \centering
        \includegraphics[width=\textwidth]{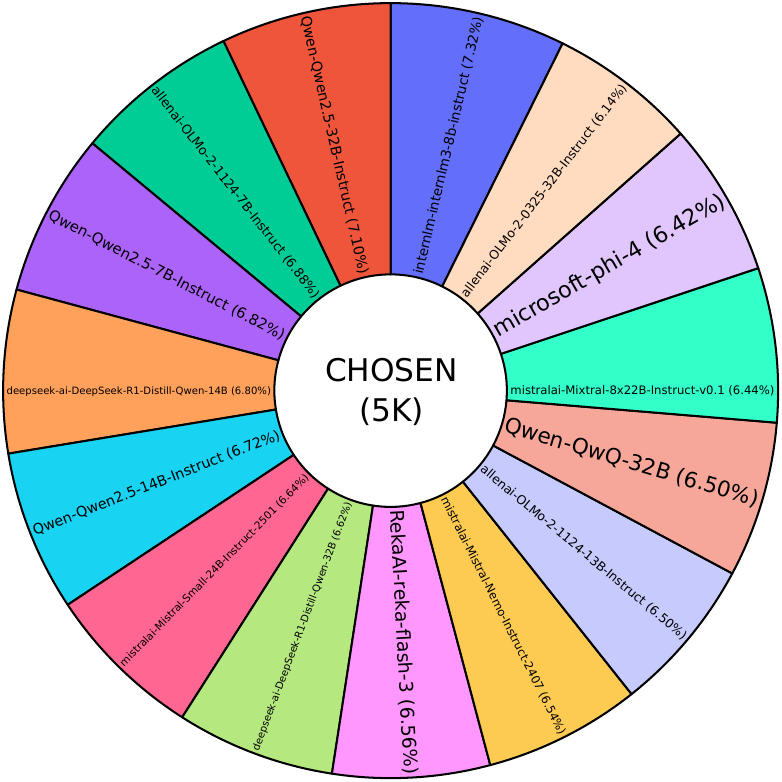}
    \end{minipage}
    \begin{minipage}{0.28\textwidth}
        \centering
        \includegraphics[width=\textwidth]{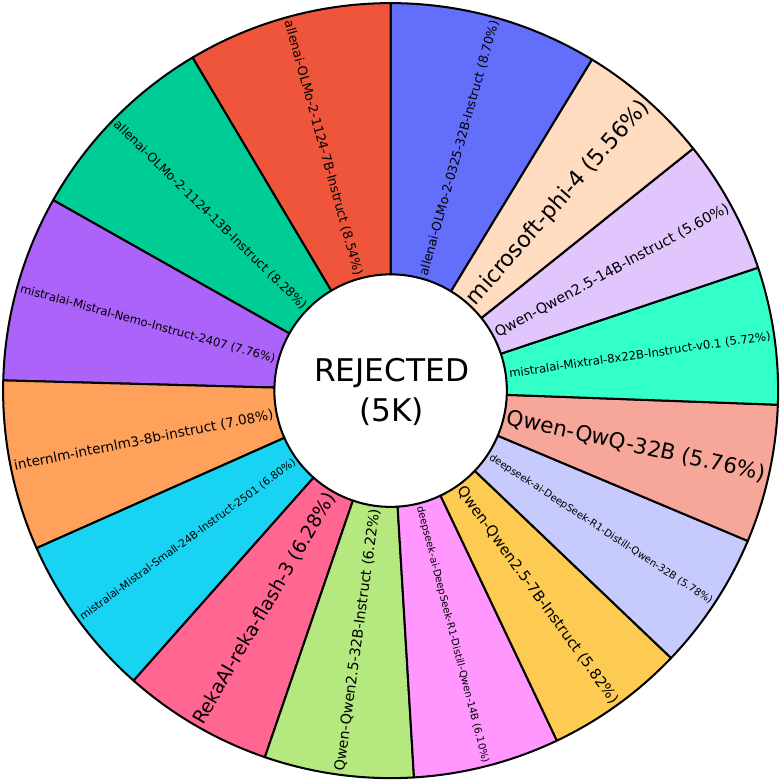}
    \end{minipage}
    \caption{Distribution of \textit{chosen} and \textit{rejected} models within the 5K subset.}
    \label{fig:chosen-and-rejected-models-distributions}
\end{figure*}

Figure \ref{fig:query-characteristics-distributions} shows the distributions of query characteristics (Subsection \ref{subsec:query-characteristics}) for the 500K, 50K, and 5K subsets. Figure \ref{fig:chosen-and-rejected-models-distributions} presents the distribution of \textit{chosen} and \textit{rejected} models within the 5K subset.

\begin{table*}[t]
    \centering
    \scalebox{0.6}{
        \begin{tabular}{lccccccccccc}
            \hline
            \textbf{Model} & \textbf{Param.} & \textbf{Pairs} & \textbf{\makecell{Refusal\\(Ans.)}} & \textbf{\makecell{Refusal\\(Unans.)}} & \textbf{\makecell{Faithful.\\(QA)}} & \textbf{\makecell{Faithful.\\(Summ.)}} & \textbf{\makecell{Complete.\\(QA)}} & \textbf{\makecell{Complete.\\(Summ.)}} & \textbf{\makecell{Concise.\\(QA)}} & \textbf{\makecell{Concise.\\(Summ.)}} & \textbf{\makecell{Overall\\Accuracy}} \\
            \hline
            \multicolumn{12}{c}{\textit{RAGferee Discriminative RMs (ours)}} \\
            \hline
            Qwen-2.5-RAGferee & 7B & 4K & 92.8 & 66.4 & 85.2 & 66.8 & 57.6 & 64.1 & 73.3 & 53.7 & 70.0 \\
            Qwen-2.5-RAGferee & 14B & 4K & 92.8 & 71.2 & 86.8 & 70.8 & 65.2 & 66.9 & 71.4 & 52.0 & 72.2 \\
            Mistral-Nemo-RAGferee & 12B & 4K & 92.0 & 82.8 & 82.8 & 68.8 & 62.4 & 62.9 & 86.3 & 57.0 & 74.5 \\
            Mistral-Small-RAGferee & 24B & 4K & 92.4 & 81.6 & 87.2 & 75.6 & 65.6 & 65.3 & 76.5 & 57.0 & 75.2 \\
            \hdashline
            \multicolumn{12}{c}{\textit{trained w/o grounding (ablation)}} \\
            \hdashline
            Qwen-2.5-RAGferee & 7B & 4K & 82.8 & 25.2 & 61.6 & 60.4 & 52.8 & 61.4 & 32.5 & 41.8 & 52.3 \\
            Qwen-2.5-RAGferee & 14B & 4K & 92.8 & 7.2 & 74.0 & 60.0 & 56.8 & 64.9 & 52.9 & 43.9 & 56.6 \\
            Mistral-Nemo-RAGferee & 12B & 4K & 84.0 & 21.6 & 56.4 & 55.6 & 48.0 & 67.3 & 47.5 & 57.8 & 54.8 \\
            Mistral-Small-RAGferee & 24B & 4K & 76.0 & 11.2 & 72.8 & 64.0 & 60.4 & 67.7 & 59.6 & 43.0 & 56.9 \\
            \hline
        \end{tabular}
    }
    \caption{Expanded version of Table \ref{tab:contextual-judge-bench-results-ragferee-wo-grounding}. \textsc{ContextualJudgeBench} results (\textit{contextual accuracy}) for \textit{RAGferee} discriminative RMs from Table \ref{tab:contextual-judge-bench-results} trained without grounding. Preference data alone is not sufficient. Incorporating retrieved context is crucial for accurately judging RAG responses.}
    \label{tab:contextual-judge-bench-results-ragferee-wo-grounding-expanded}
\end{table*}

\begin{table*}[t]
    \centering
    \scalebox{0.6}{
        \begin{tabular}{lcc|cccc|cccc|c}
            \hline
            \multicolumn{3}{c}{} & \multicolumn{4}{c}{\textbf{Helpful}} & \multicolumn{4}{c}{\textbf{Harmless}} & \multicolumn{1}{c}{} \\
            \textbf{Model} & \textbf{Param.} & \textbf{Pairs} & \textbf{General} & \textbf{Reason} & \textbf{Citation} & \textbf{Average} & \textbf{General} & \textbf{Abstain} & \textbf{Conflict} & \textbf{Average} & \textbf{\makecell{Overall\\Accuracy}} \\
            \hline
            \multicolumn{12}{c}{\textit{Discriminative RMs (baselines)}} \\
            \hline
            InternLM-2 & 7B & 2400K & 78.6 & 68.0 & 67.0 & 70.6 & 62.6 & 66.8 & 54.9 & 61.7 & 67.3 \\
            InternLM-2 & 20B & 2400K & 79.4 & 67.6 & 67.6 & 70.9 & 65.8 & 67.7 & 60.9 & 64.9 & 68.7 \\
            LLaMA-3.1-Skywork-v0.2 & 8B & 80K & 76.3 & 71.9 & 56.0 & 67.0 & 85.8 & 74.7 & 79.9 & 79.5 & 71.6 \\
            Gemma-2-Skywork-v0.2 & 27B & 80K & 76.0 & 73.2 & 64.8 & 70.7 & 79.4 & 80.6 & 78.8 & 79.7 & 74.1 \\
            \hdashline
            \multicolumn{12}{c}{\textit{inferenced w/o grounding (ablation)}} \\
            \hdashline
            InternLM-2 & 7B & 2400K & 66.4 & 60.5 & 59.6 & 61.8 & 89.7 & 60.4 & 58.7 & 68.0 & 64.1 \\
            InternLM-2 & 20B & 2400K & 69.1 & 61.4 & 59.3 & 62.8 & 91.6 & 56.2 & 65.2 & 69.1 & 65.1 \\
            LLaMA-3.1-Skywork-v0.2 & 8B & 80K & 66.8 & 59.5 & 53.2 & 59.1 & 95.5 & 72.4 & 77.2 & 80.4 & 67.1 \\
            Gemma-2-Skywork-v0.2 & 27B & 80K & 66.8 & 64.4 & 59.3 & 63.1 & 93.5 & 75.6 & 78.3 & 81.5 & 70.0 \\
            \hline
            \multicolumn{12}{c}{\textit{RAGferee Discriminative RMs (ours)}} \\
            \hline
            Qwen-2.5-RAGferee & 7B & 4K & 65.3 & 65.0 & 59.8 & 63.1 & 52.3 & 66.4 & 37.0 & 52.7 & 59.2 \\
            Qwen-2.5-RAGferee & 14B & 4K & 72.1 & 70.6 & 59.8 & 66.8 & 51.0 & 72.4 & 43.5 & 56.8 & 63.1 \\
            Mistral-Nemo-RAGferee & 12B & 4K & 67.6 & 64.7 & 58.4 & 63.1 & 50.3 & 71.0 & 43.5 & 56.1 & 60.5 \\
            Mistral-Small-RAGferee & 24B & 4K & 67.6 & 67.0 & 58.7 & 63.9 & 47.7 & 71.9 & 48.4 & 57.4 & 61.5 \\
            \hdashline
            \multicolumn{12}{c}{\textit{inferenced w/o grounding (ablation)}} \\
            \hdashline
            Qwen-2.5-RAGferee & 7B & 4K & 63.4 & 55.9 & 65.9 & 61.9 & 93.5 & 70.5 & 67.4 & 75.9 & 67.1 \\
            Qwen-2.5-RAGferee & 14B & 4K & 63.0 & 57.8 & 64.5 & 61.9 & 92.3 & 81.6 & 72.3 & 81.5 & 69.2 \\
            Mistral-Nemo-RAGferee & 12B & 4K & 56.9 & 51.0 & 55.7 & 54.5 & 92.3 & 78.3 & 65.8 & 78.1 & 63.3 \\
            Mistral-Small-RAGferee & 24B & 4K & 63.7 & 62.1 & 61.5 & 62.3 & 81.3 & 80.2 & 70.1 & 77.2 & 67.9 \\
            \hline
        \end{tabular}
    }
    \caption{Expanded version of Table \ref{tab:rag-reward-bench}. \textsc{RAG-RewardBench} results (\textit{consistent accuracy}) for discriminative RMs inferenced with or without grounding. Grounding has minimal impact on baseline discriminative RMs (non-RAG), but it significantly influences our \textit{RAGferee} discriminative RMs, which are sensitive to grounding by design.}
    \label{tab:rag-reward-bench-expanded}
\end{table*}

\begin{table*}[t]
    \centering
    \scalebox{0.6}{
        \begin{tabular}{lccccccccccc}
            \hline
            \textbf{Model} & \textbf{Param.} & \textbf{Pairs} & \textbf{\makecell{Refusal\\(Ans.)}} & \textbf{\makecell{Refusal\\(Unans.)}} & \textbf{\makecell{Faithful.\\(QA)}} & \textbf{\makecell{Faithful.\\(Summ.)}} & \textbf{\makecell{Complete.\\(QA)}} & \textbf{\makecell{Complete.\\(Summ.)}} & \textbf{\makecell{Concise.\\(QA)}} & \textbf{\makecell{Concise.\\(Summ.)}} & \textbf{\makecell{Overall\\Accuracy}} \\
            \hline
            \multicolumn{12}{c}{\textit{Generative (non-reward) Models (baselines)}} \\
            \hline
            Qwen-2.5 & 7B & - & 30.0 & 33.2 & 29.2 & 28.0 & 29.2 & 25.5 & 17.6 & 31.6 & 28.0 \\
            Qwen-2.5 & 14B & - & 54.0 & 63.2 & 42.8 & 39.6 & 27.6 & 40.6 & 8.6 & 22.1 & 37.3 \\
            Mistral-Nemo & 12B & - & 24.8 & 12.0 & 26.4 & 27.6 & 28.0 & 35.9 & 10.2 & 13.5 & 22.3 \\
            Mistral-Small & 24B & - & 50.8 & 54.8 & 48.0 & 42.8 & 34.4 & 47.0 & 25.9 & 35.7 & 42.4 \\
            \hline
            \multicolumn{12}{c}{\textit{RAGferee} Generative Models (SFT) (ours)} \\
            \hline
            Qwen-2.5-RAGferee & 7B & 4K & 75.2 & 61.2 & 54.0 & 44.0 & 29.6 & 28.7 & 51.4 & 29.9 & 46.8 \\
            Qwen-2.5-RAGferee & 14B & 4K & 83.6 & 68.0 & 60.8 & 44.0 & 38.4 & 36.3 & 64.7 & 35.2 & 53.9 \\
            Mistral-Nemo-RAGferee & 12B & 4K & 77.6 & 70.0 & 55.6 & 29.2 & 37.2 & 32.3 & 63.1 & 34.8 & 50.0 \\
            Mistral-Small-RAGferee & 24B & 4K & 80.8 & 64.4 & 54.4 & 37.2 & 35.6 & 33.5 & 62.0 & 35.2 & 50.4 \\
            \hline
        \end{tabular}
    }
    \caption{Expanded version of Table \ref{tab:generative-ragferee}. \textsc{ContextualJudgeBench} results (\textit{consistent accuracy}) of \textit{RAGferee} generative RMs trained to output the indicator of the preferred response.}
    \label{tab:generative-ragferee-expanded}
\end{table*}

\clearpage
\onecolumn

\subsection{Examples}

\subsubsection{\textsc{ContextualJudgeBench}: Refusal (Answerable)}

{\fontsize{10pt}{10pt}\selectfont
\begin{minipage}{\linewidth}
\begin{query}[\texttt{Query}]{}
\begin{Verbatim}[breaklines=true, breakanywhere=true, breaksymbol=, breakanywheresymbolpre=]
how to be a zero-bug programmer?
\end{Verbatim}
\end{query}
\begin{grounding}[\texttt{Grounding}]{}
\begin{Verbatim}[breaklines=true, breakanywhere=true, breaksymbol=, breakanywheresymbolpre=]
[1] Don't code at all. That's the only way you can be a zero-bug programmer. Bugs are unavoidable because programmers are human, all we can do is try our best to prevent them, react quickly when a bug occurs, learn from our mistakes and stay up to date.
\end{Verbatim}
\end{grounding}
\begin{chosen}[\texttt{Chosen}]{}
\begin{Verbatim}[breaklines=true, breakanywhere=true, breaksymbol=, breakanywheresymbolpre=]
To achieve the status of a zero-bug programmer, the advice is straightforward: abstain from coding entirely.  
Since bugs are an inevitable aspect of programming due to human error, the focus should instead be on diligent prevention, swift response to issues, learning from past errors, and keeping knowledge current.
\end{Verbatim}
\end{chosen}
\begin{rejected}[\texttt{Rejected}]{}
\begin{Verbatim}[breaklines=true, breakanywhere=true, breaksymbol=, breakanywheresymbolpre=]
The contexts provided touch on the inevitability of bugs in programming but lack detailed methodologies, best practices in software development, and specific quality assurance techniques necessary to achieve a near-zero-bug codebase. To properly answer the question, information about systematic testing approaches, code review strategies, and the integration of automated tools in the development lifecycle would be essential.
\end{Verbatim}
\end{rejected}
\begin{prediction}[\texttt{Predictions}]{}
\begin{Verbatim}[breaklines=true, breakanywhere=true, breaksymbol=, breakanywheresymbolpre=, commandchars=\\\{\}]
Gemma-2-Skywork-v0.2 (80K pairs & 27B parameters):                          \colorbox{rejectedcolor}{\textcolor{white}{\textbf{Rejected}}}

Mistral-Small-RAGferee (4K pairs & 24B parameters):                         \colorbox{chosencolor}{\textcolor{white}{\textbf{Chosen}}}

Mistral-Small-RAGferee (4K pairs & 24B parameters) (trained w/o grounding): \colorbox{rejectedcolor}{\textcolor{white}{\textbf{Rejected}}}
\end{Verbatim}
\end{prediction}
\end{minipage}
}

\newpage

\subsubsection{\textsc{ContextualJudgeBench}: Refusal (Unanswerable)}

{\fontsize{10pt}{10pt}\selectfont
\begin{minipage}{\linewidth}
\begin{query}[\texttt{Query}]{}
\begin{Verbatim}[breaklines=true, breakanywhere=true, breaksymbol=, breakanywheresymbolpre=]
What does mTOR stands for?
\end{Verbatim}
\end{query}
\begin{grounding}[\texttt{Grounding}]{}
\begin{Verbatim}[breaklines=true, breakanywhere=true, breaksymbol=, breakanywheresymbolpre=]
Ghrelin exhibits its biological effect through binding to the growth hormone secretagogue 1a receptor (GHS-R1a). Recently, it has been reported that ghrelin has an anti-apoptotic effect in several cell types. However, the molecule mechanisms underlying the anti-apoptotic effect of ghrelin remain poorly understood. In this study, we investigated the intracellular mechanisms responsible for anti-apoptotic effect of ghrelin on human umbilical vein endothelial cells (HUVEC). Treatment of HUVEC with ghrelin inhibited high glucose-induced cell apoptosis. Ghrelin stimulated the rapid phosphorylation of key proteins involved in cell survival. The GHS-R1a-specific antagonist [D-Lys3]-GHRP-6 abolished the anti-apoptotic effect and inhibited the activation of key survival pathways induced by ghrelin. Pretreatment of cells with specific inhibitors blocked the anti-apoptotic effect of ghrelin. In addition, ghrelin protected HUVECs against high glucose induced apoptosis by increasing Bcl-2/Bax ratio. Taken together, our results demonstrate that ghrelin produces a protective effect on HUVECs through activating GHS-R1a and specific signaling pathways mediates the effect of ghrelin. These observations suggest that ghrelin may act as a survival factor in preventing HUVECs apoptosis caused by high glucose.
\end{Verbatim}
\end{grounding}
\begin{chosen}[\texttt{Chosen}]{}
\begin{Verbatim}[breaklines=true, breakanywhere=true, breaksymbol=, breakanywheresymbolpre=]
While the context provides a detailed examination of the anti-apoptotic effects of ghrelin and its interaction with specific receptors and signaling pathways, it does not mention or define the term mTOR or its biological significance. To accurately explain what mTOR stands for, the context would need to include information on molecular signaling pathways, specifically those involving the mechanistic target of rapamycin, which is a central regulator of cell metabolism, growth, proliferation, and survival.
\end{Verbatim}
\end{chosen}
\begin{rejected}[\texttt{Rejected}]{}
\begin{Verbatim}[breaklines=true, breakanywhere=true, breaksymbol=, breakanywheresymbolpre=]
The context discusses the anti-apoptotic effects of ghrelin on endothelial cells but does not directly mention mTOR or its full form. However, mTOR is a well-known protein kinase involved in cell survival pathways, which aligns with the context describing phosphorylation and cell survival mechanisms. Thus, mTOR is likely relevant to the study's focus on signaling pathways.

mTOR stands for mechanistic Target of Rapamycin.
\end{Verbatim}
\end{rejected}
\begin{prediction}[\texttt{Predictions}]{}
\begin{Verbatim}[breaklines=true, breakanywhere=true, breaksymbol=, breakanywheresymbolpre=, commandchars=\\\{\}]
Gemma-2-Skywork-v0.2 (80K pairs & 27B parameters):                          \colorbox{rejectedcolor}{\textcolor{white}{\textbf{Rejected}}}

Mistral-Small-RAGferee (4K pairs & 24B parameters):                         \colorbox{chosencolor}{\textcolor{white}{\textbf{Chosen}}}

Mistral-Small-RAGferee (4K pairs & 24B parameters) (trained w/o grounding): \colorbox{rejectedcolor}{\textcolor{white}{\textbf{Rejected}}}
\end{Verbatim}
\end{prediction}
\end{minipage}
}

\newpage

\subsubsection{\textsc{ContextualJudgeBench}: Faithfulness (QA)}

{\fontsize{9pt}{9pt}\selectfont
\begin{minipage}{\linewidth}
\begin{query}[\texttt{Query}]{}
\begin{Verbatim}[breaklines=true, breakanywhere=true, breaksymbol=, breakanywheresymbolpre=]
How does evolution occur and how did the first living organisms come about
\end{Verbatim}
\end{query}
\begin{grounding}[\texttt{Grounding}]{}
\begin{Verbatim}[breaklines=true, breakanywhere=true, breaksymbol=, breakanywheresymbolpre=]
Charles Carter and Richard Wolfenden, both of the University of North Carolina, have uncovered new evidence of abiogenesis, the process by which life arises from non-living chemical matter. Their study, published Thursday in the Journal of Biological Chemistry, suggests that a single ancient gene may have used each of its opposite DNA strands to code for different chemical catalysts. Those separate catalysts would have both activated amino acids, which then formed proteins – essential to the production of living cells.

In the early 20th century, the “primordial soup” model of abiogenesis started to gain traction. It proposes that in Earth’s prebiotic history, simple organic matter was exposed to energy in the form of volcanoes and electrical storms. That energy would have catalyzed chemical reactions that, in the span of a few hundred million years, could have produced self-replicating molecules.

Today, the Miller-Urey experiment is contested for a number of reasons, including the possibility that Earth’s original atmosphere may have had a different composition. Still, the production of organic compounds from inorganic “precursors” laid a strong foundation for the primordial soup hypothesis. And new findings support that hypothesis, Dr. Carter says.

The first living things on Earth, single-celled micro-organisms or microbes lacking a cell nucleus or cell membrane known as prokaryotes, seem to have first appeared on Earth almost four billion years ago, just a few hundred million years after the formation of the Earth itself. By far the longest portion of the history of life on Earth, therefore, has involved the biochemical evolution of these single-celled micro-organisms, bacteria and archaea: we can find individual fossilized microbes in rocks 3.4 billion years old, yet we can only conclusively identify multi-celled fossils in rocks younger than 1 billion years.

It is presumed that, over a few hundred million years of evolution, pre-biotic molecules evolved into self-replicating molecules by natural selection. While some aspects of the subject are well understood, others remain clouded in mystery and are the source of much contention among scientists. Although much progress has been made, there is still no single definitive theory.
\end{Verbatim}
\end{grounding}
\begin{chosen}[\texttt{Chosen}]{}
\begin{Verbatim}[breaklines=true, breakanywhere=true, breaksymbol=, breakanywheresymbolpre=]
The process of abiogenesis, or the emergence of life from non-living matter, is still not fully understood and is the subject of much debate among scientists. Several theories exist, one of the most well-known being the "primordial soup" theory, which proposes that in the early stages of Earth\'s history, organic matter was exposed to energy in the form of volcanoes and electrical storms, which would have catalyzed chemical reactions that could have produced self-replicating molecules. Evidence of this theory is supported by the fact that organic compounds can be produced from inorganic "precursors" in the lab. Another theory is that a single ancient gene may have used each of its opposite DNA strands to code for different chemical catalysts, which would have both activated amino acids, which then formed proteins - essential to the production of living cells. Additionally, it is believed that over the course of hundreds of millions of years, pre-biotic molecules evolved into self-replicating molecules by natural selection.
\end{Verbatim}
\end{chosen}
\begin{rejected}[\texttt{Rejected}]{}
\begin{Verbatim}[breaklines=true, breakanywhere=true, breaksymbol=, breakanywheresymbolpre=]
Evolution occurs when successive generations of organisms inherit different traits, which can be passed down to their offspring. Over time, these changes can result in new species and even major changes in the structure of organisms. The first living organisms likely arose from non-living chemicals around 4 billion years ago, in a process known as abiogenesis. During this process, simple organic molecules were likely catalyzed by energy from volcanoes and lightning into more complex molecules, which eventually led to the first living cells.
\end{Verbatim}
\end{rejected}
\begin{prediction}[\texttt{Predictions}]{}
\begin{Verbatim}[breaklines=true, breakanywhere=true, breaksymbol=, breakanywheresymbolpre=, commandchars=\\\{\}]
Gemma-2-Skywork-v0.2 (80K pairs & 27B parameters):                          \colorbox{rejectedcolor}{\textcolor{white}{\textbf{Rejected}}}

Mistral-Small-RAGferee (4K pairs & 24B parameters):                         \colorbox{chosencolor}{\textcolor{white}{\textbf{Chosen}}}

Mistral-Small-RAGferee (4K pairs & 24B parameters) (trained w/o grounding): \colorbox{rejectedcolor}{\textcolor{white}{\textbf{Rejected}}}
\end{Verbatim}
\end{prediction}
\end{minipage}
}

\newpage

\subsubsection{\textsc{ContextualJudgeBench}: Faithfulness (Summarisation)}

{\fontsize{7pt}{7pt}\selectfont
\begin{minipage}{\linewidth}
\begin{query}[\texttt{Query}]{}
\begin{Verbatim}[breaklines=true, breakanywhere=true, breaksymbol=, breakanywheresymbolpre=]
Write a precise and to-the-point summary based on the provided context focusing on Rent Increase Regulations.
\end{Verbatim}
\end{query}
\begin{grounding}[\texttt{Grounding}]{}
\begin{Verbatim}[breaklines=true, breakanywhere=true, breaksymbol=, breakanywheresymbolpre=]
Speaker 1: Five F recommendation to authorize the Interim City Manager to execute an agreement with CSG consulting services for professional plant services for 12 months. Permanent amount not to exceed 300,000.
Speaker 0: And I pulled this item. I don't know if staff wants to present anything about it. I pulled it because it's for up to 300,000 for structural engineer or an engineering firm. And it's my understanding that we, the city had had had an engineer through that, resigned in June. And so now we're being asked to approve a contract with a consulting firm out of San Mateo to make up for that. That would be up to 12 months and $300,000.
Speaker 2: So. Good evening. My name is Debbie Potter and I'm the city's community development director. And it is accurate that we had a plan check engineer that was working with the city. He actually resigned a year ago in June. So we have been using outside planned check services for several years now. We are using these planned check services. We are incredibly busy at the permit center and the flexibility that we have by using planned check services outside, planned check services we can use from one to 10 to 15 plan checkers all at the same time to keep moving our projects through the system. And while we do use planned check services, it should be noted that about 75% of all plan checking happens over the counter right here at City Hall. But 25% of our projects are complicated enough that they have to go out to a structural engineer. We're using CSG. It's the same plan check services that we that the fire department uses for its planned tech services. So that synergy has been very nice for us to have plan checkers that are very familiar with both the city's fire code and then all the building codes. We would like to continue on with the services that we are receiving from CSG. There is no impact on the general fund CSG. The contract is set up so that they charge 65% of the cost of the building permit, so that we are always providing services to applicants within the budget that we charge for building permits. It's no impact on the general fund and we will be continuing to look throughout this fiscal year at how we want to staff up that function within the department. And we're looking at perhaps keeping outside contract services and then possibly under filling the position with a plans examiner so we can do more over the counter. So we feel like we have an efficient model in place and it's really based on that analysis and the track record for the last year and a half that we're recommending approval of this contract for the current fiscal year.
Speaker 0: Do we have any engineers as part of our staff? Can you describe the types of engineers that we have as part of our staff?
Speaker 2: So the building official is has a degree in architecture, which is structural engineers and architects are the ones who can do the plan, check services. And then we have our supervising building inspector is a plan examiner, a certified plans examiner. So those are the two certified staff that we have.
Speaker 0: But do we have any engineers as part of our staff? That's actually my concern.
Speaker 2: We currently do not have a structural engineer, Madame Mayor. We have engineers in our public works department.
Speaker 0: Advanced planning. No, no, no. So that's that's where I'm that's where my question goes. My preference would be that we have at least one engineer on our staff. And it appears that that's counter to what the plan currently is with staff, that you plan to continue to have a consulting firm as opposed to an engineer that's actually an employee of the city, which I we used to have that and I and I think I've seen that that issue has been raised in regards to having an engineer as part of our planning department. So that was my concern. And I pulled out. And I actually think that the community is interested in this issue because there have been multiple letters to the editor regarding engineers as part of our planning department, separate from public works. And I appreciate that.
Speaker 2: Right. Most of those letters, though, are all about the public works engineers. Those letters that we've received have not been about the planning department. But I understand what you're saying, and we can take a look at it, because our building official, who has a degree in architecture, is able to do the same types of plan checking as a structural engineer it we just feel like it's more efficient to be able to access multiple structural engineers during our very busy period. And that's really what what we need and what we need. Internal to our staff are the certified plans examiners. And that's we also we do have several certified plans examiners on staff.
Speaker 0: And I appreciate that. Thank you for the comments.
Speaker 8: No, I just said I'm willing to move approval because I think Mr. Potter gave an excellent explanation if we want to keep these projects moving. Time is money. This is not impacting our general fund, and I think we're serving our community well. So with that, I will move approval of item five after recommendation to authorize the Interim City Manager to execute an agreement with CSG consulting services for professional plant services for a 12 month term in an amount not to exceed $300,000.
Speaker 0: We have a second.
Speaker 3: Second. All right.
Speaker 0: Any other comments? All those in favor.
Speaker 3: I think you.
Speaker 0: Of motion passes 4 to 1. Thank you. Thank you. And now we have speakers on five. Q And five are with you.
Speaker 1: And five Q is a final passage of ordinance amending the Alameda Municipal Code by adding article? What does that 14 to Chapter six concerning review of rent increases.
\end{Verbatim}
\end{grounding}
\begin{chosen}[\texttt{Chosen}]{}
\begin{Verbatim}[breaklines=true, breakanywhere=true, breaksymbol=, breakanywheresymbolpre=]
The council discussed approving a contract for outside engineering services. The mayor prefers hiring an in-house engineer. Staff argued outside engineers allow flexibility and efficient permit processing. The contract charges applicants, not the general fund. The council approved the contract 4-1, enabling continued tenant protections through efficient permit review.
\end{Verbatim}
\end{chosen}
\begin{rejected}[\texttt{Rejected}]{}
\begin{Verbatim}[breaklines=true, breakanywhere=true, breaksymbol=, breakanywheresymbolpre=]
The document outlines a review process for rent increases in Alameda, California, which includes a threshold for when increases require review and a process for submitting and reviewing rent increase requests. The review process involves the Rent Stabilization Board, landlords, and tenants, and may include public hearings. The goal of the process is to ensure that rent increases are reasonable and do not harm tenants or the community.
\end{Verbatim}
\end{rejected}
\begin{prediction}[\texttt{Predictions}]{}
\begin{Verbatim}[breaklines=true, breakanywhere=true, breaksymbol=, breakanywheresymbolpre=, commandchars=\\\{\}]
Gemma-2-Skywork-v0.2 (80K pairs & 27B parameters):                          \colorbox{rejectedcolor}{\textcolor{white}{\textbf{Rejected}}}

Mistral-Small-RAGferee (4K pairs & 24B parameters):                         \colorbox{chosencolor}{\textcolor{white}{\textbf{Chosen}}}

Mistral-Small-RAGferee (4K pairs & 24B parameters) (trained w/o grounding): \colorbox{rejectedcolor}{\textcolor{white}{\textbf{Rejected}}}
\end{Verbatim}
\end{prediction}
\end{minipage}
}

\newpage

\subsubsection{\textsc{ContextualJudgeBench}: Completeness (QA)}

{\fontsize{9pt}{9pt}\selectfont
\begin{minipage}{\linewidth}
\begin{query}[\texttt{Query}]{}
\begin{Verbatim}[breaklines=true, breakanywhere=true, breaksymbol=, breakanywheresymbolpre=]
What war was taking place in wonder woman?
\end{Verbatim}
\end{query}
\begin{grounding}[\texttt{Grounding}]{}
\begin{Verbatim}[breaklines=true, breakanywhere=true, breaksymbol=, breakanywheresymbolpre=]
Wonder Woman (2017 film)
This version was conceived of as a prequel to the first live-action, theatrical appearance of Wonder Woman, in the 2016 film, "", placing Wonder Woman in the 1910s and World War I (a decision which differs from her comic book origins as a supporter of the Allies during World War II).
As for story development, Jenkins credits the stories by the character's creator William Moulton Marston in the 1940s and George Perez's seminal stories in the 1980s in which he modernized the character.
In addition, it follows some aspects of DC Comics' origin changes in The New 52 reboot, where Diana is the daughter of Zeus.
Jenkins cited Richard Donner's "Superman" as an inspiration.
In late 2013, Zack Snyder cast Gal Gadot in the role of Wonder Woman for the 2016 film, "" over Élodie Yung and Olga Kurylenko.
Some fans initially reacted to this choice by criticizing Gadot's appearance.
Snyder would later comment on his decision to cast Gadot, stating that he tested a "bunch of actresses, as you can imagine.
But the thing with Gal is that she's strong, she's beautiful, and she's a kind person, which is interesting, but fierce at the same time.
It's that combination of being fierce but kind at the same time that we were looking for.

Publication history of Wonder Woman
Following the popularity of the "Wonder Woman" TV series (initially set during World War II), the comic book was also transposed to this era.
The change was made possible by the multiverse concept, which maintained that the 1970s Wonder Woman and the original 1940s version existed in two separate yet parallel worlds.
A few months after the TV series changed its setting to the 1970s, the comic book returned to the contemporary timeline.
Soon after, when the series was written by Jack C. Harris, Steve (Howard) Trevor was killed off yet again.
Wonder Woman was one of the backup features in "World's Finest Comics" #244-252 and "Adventure Comics" #459-464 when those titles were in the Dollar Comics format.
"All-New Collectors' Edition" #C-54 (1978) featured a Superman vs. Wonder Woman story by writer Gerry Conway and artists José Luis García-López and Dan Adkins.
Writer Gerry Conway brought Steve Trevor back to life again in issue #271 (September 1980).
Following Diana's renunciation of her role as Wonder Woman, a version of Steve Trevor from an undisclosed portion of the Multiverse accidentally made the transition to Earth-One.
With Diana's memory erased by the Mists of Nepenthe, the new Steve again crash-landed and arrived at Paradise Island.
\end{Verbatim}
\end{grounding}
\begin{chosen}[\texttt{Chosen}]{}
\begin{Verbatim}[breaklines=true, breakanywhere=true, breaksymbol=, breakanywheresymbolpre=]
Following the popularity of the "Wonder Woman" TV series (initially set during World War II), the comic book was also transposed to this era. The change was made possible by the multiverse concept, which maintained that the 1970s Wonder Woman and the original 1940s version existed in two separate yet parallel worlds.
\end{Verbatim}
\end{chosen}
\begin{rejected}[\texttt{Rejected}]{}
\begin{Verbatim}[breaklines=true, breakanywhere=true, breaksymbol=, breakanywheresymbolpre=]
During Wonder Woman's first live-action film, set in the 1910s, the story took place during World War I. In the comic book, it was set in the 1970s, during World War II.
\end{Verbatim}
\end{rejected}
\begin{prediction}[\texttt{Predictions}]{}
\begin{Verbatim}[breaklines=true, breakanywhere=true, breaksymbol=, breakanywheresymbolpre=, commandchars=\\\{\}]
Gemma-2-Skywork-v0.2 (80K pairs & 27B parameters):                          \colorbox{rejectedcolor}{\textcolor{white}{\textbf{Rejected}}}

Mistral-Small-RAGferee (4K pairs & 24B parameters):                         \colorbox{chosencolor}{\textcolor{white}{\textbf{Chosen}}}

Mistral-Small-RAGferee (4K pairs & 24B parameters) (trained w/o grounding): \colorbox{rejectedcolor}{\textcolor{white}{\textbf{Rejected}}}
\end{Verbatim}
\end{prediction}
\end{minipage}
}

\newpage

\subsubsection{\textsc{ContextualJudgeBench}: Completeness (Summarisation)}

{\fontsize{9pt}{9pt}\selectfont
\begin{minipage}{\linewidth}
\begin{query}[\texttt{Query}]{}
\begin{Verbatim}[breaklines=true, breakanywhere=true, breaksymbol=, breakanywheresymbolpre=]
Write an accurate and brief summary derived from the given context.
\end{Verbatim}
\end{query}
\begin{grounding}[\texttt{Grounding}]{}
\begin{Verbatim}[breaklines=true, breakanywhere=true, breaksymbol=, breakanywheresymbolpre=]
USER : I'm looking for a train to Stevenage that leaves after 17:30. Is there anything available then?
SYSTEM : I have a few choices for you. What day are you wanting to travel?
USER : I would like to leave on Thursday out of Cambridge.
SYSTEM : The TR0385 leaves at 19:21, would that suit you?
USER : Yes, that sounds perfect. I would like a booking for 4 people.
SYSTEM : The booking was successful. Your reference number is SICMOAA9 . Can I help you with anything else?
USER : I am looking for some places to go in town. I want something in the centre that has multiple sports.
SYSTEM : I am sorry but there is nothing matching your request. Would you like to try another part of town or type of attraction?
USER : Well, how about any nightclubs?
SYSTEM : Yes I have one located on the east side.
USER : Do you have any nightclubs located in the centre?
SYSTEM : Yes there are 5 located in the centre. Is there anything I can help you with?
USER : Can I get the phone number for one of them?
SYSTEM : Sure, Ballare in heidelberg gardens, lion yard cost 5 pounds and their phone number is 01223364222. Is there anything else you'd like?
USER : No. That's everything I was looking for today. Thanks for your help.
SYSTEM : Thank you for choosing help desk. Good Bye.
\end{Verbatim}
\end{grounding}
\begin{chosen}[\texttt{Chosen}]{}
\begin{Verbatim}[breaklines=true, breakanywhere=true, breaksymbol=, breakanywheresymbolpre=]
The user successfully booked a train to Stevenage for 4 people, leaving Cambridge on Thursday at 19:21. The booking reference number is SICMOAA9. Additionally, the user inquired about places of interest in the town center, specifically sports locations and nightclubs. Although no sports locations matched their criteria, they were given information about nightclubs, including the phone number for Ballare, located at Heidelberg Gardens, Lion Yard, with an entry cost of 5 pounds.
\end{Verbatim}
\end{chosen}
\begin{rejected}[\texttt{Rejected}]{}
\begin{Verbatim}[breaklines=true, breakanywhere=true, breaksymbol=, breakanywheresymbolpre=]
The text is a conversation between a user and a system. The user asks for a train to Stevenage after 17:30, books tickets for 4 people, then asks for recommendations for places to go in town focusing on nightclubs in the centre. The system provides train options, booking confirmation, and contact details for a nightclub matching the requests.
\end{Verbatim}
\end{rejected}
\begin{prediction}[\texttt{Predictions}]{}
\begin{Verbatim}[breaklines=true, breakanywhere=true, breaksymbol=, breakanywheresymbolpre=, commandchars=\\\{\}]
Gemma-2-Skywork-v0.2 (80K pairs & 27B parameters):                          \colorbox{rejectedcolor}{\textcolor{white}{\textbf{Rejected}}}

Mistral-Small-RAGferee (4K pairs & 24B parameters):                         \colorbox{chosencolor}{\textcolor{white}{\textbf{Chosen}}}

Mistral-Small-RAGferee (4K pairs & 24B parameters) (trained w/o grounding): \colorbox{rejectedcolor}{\textcolor{white}{\textbf{Rejected}}}
\end{Verbatim}
\end{prediction}
\end{minipage}
}

\newpage

\subsubsection{\textsc{ContextualJudgeBench}: Conciseness (QA)}

{\fontsize{9pt}{9pt}\selectfont
\begin{minipage}{\linewidth}
\begin{query}[\texttt{Query}]{}
\begin{Verbatim}[breaklines=true, breakanywhere=true, breaksymbol=, breakanywheresymbolpre=]
Who played stormy in the movie odd thomas?
\end{Verbatim}
\end{query}
\begin{grounding}[\texttt{Grounding}]{}
\begin{Verbatim}[breaklines=true, breakanywhere=true, breaksymbol=, breakanywheresymbolpre=]
Odd Thomas (film)
Odd Thomas (film)

Odd Thomas is a 2013 American supernatural mystery thriller film based on Dean Koontz's novel of the same name.
It is directed, written, and co-produced by Stephen Sommers and stars Anton Yelchin as Odd Thomas, with Willem Dafoe as Wyatt Porter, and Addison Timlin as Stormy Llewellyn.
Odd Thomas (Yelchin) is a psychic who lives in a small town in California.
He describes his ability as, "I see dead people, but then, by God, I do something about it."
One morning the ghost of a teenage girl, Penny Kallisto, silently leads him to Harlo Landerson.
Odd accuses Harlo of raping and murdering Penny.
Harlo flees.
Odd chases him to a pool party and into a child's bedroom in a stranger's house.
Harlo and Odd fight and Harlo is knocked unconscious.
Odd's friend, police chief Wyatt Porter (Dafoe), is aware of Odd's psychic gifts and promises to spin the story to keep public attention away from him.
Odd has a vision of faceless people wearing bowling shirts who cry out to him to save them.
A faceless gunman shoots them all, including Odd.

Addison Timlin
Addison Timlin

Addison Jayne Timlin (born June 29, 1991) is an American actress, best known for her roles as Jami Lerner in "The Town That Dreaded Sundown" (2014) and Colleen Lunsford in "Little Sister" (2016).
She is also known for playing Sasha Bingham in Showtime's "Californication".
Born in Philadelphia, Pennsylvania, Timlin debuted on the big screen as Amy, a 14-year-old girl with diabetes, in the 2005 film "Derailed" alongside Clive Owen and Melissa George, who played her parents.
Timlin played Maddy in a short film, "The Isabel Fish", directed by Lara Zizic for the Columbia Film Festival.
In 2007, Timlin appeared in the music video, "Sleeping Lessons," by Albuquerque, New Mexico based indie-rock band, The Shins.
In 2008, she made an appearance in the music video for the song Check
Yes Juliet by We the Kings, playing the fictional Juliet.
She was cast in the short-lived CBS television drama, "3 lbs", as Charlotte Hanson (the daughter of the main character played by Stanley Tucci), in three of the eight episodes that were filmed.
Timlin is seen in the short film "Man", written and directed by Myna Joseph, which was screened at the Sundance Film Festival and at Cannes.
\end{Verbatim}
\end{grounding}
\begin{chosen}[\texttt{Chosen}]{}
\begin{Verbatim}[breaklines=true, breakanywhere=true, breaksymbol=, breakanywheresymbolpre=]
Odd Thomas is a 2013 supernatural mystery thriller film based on Dean Koontz's novel of the same name. The film stars Addison Timlin as Stormy Llewellyn.
\end{Verbatim}
\end{chosen}
\begin{rejected}[\texttt{Rejected}]{}
\begin{Verbatim}[breaklines=true, breakanywhere=true, breaksymbol=, breakanywheresymbolpre=]
Odd Thomas is a 2013 supernatural mystery thriller film based on Dean Koontz's novel of the same name. The film stars Anton Yelchin as Odd Thomas, with Willem Dafoe as Wyatt Porter, and Addison Timlin as Stormy Llewellyn. American actress Addison Timlin also played Stormy in the 2013 movie.
\end{Verbatim}
\end{rejected}
\begin{prediction}[\texttt{Predictions}]{}
\begin{Verbatim}[breaklines=true, breakanywhere=true, breaksymbol=, breakanywheresymbolpre=, commandchars=\\\{\}]
Gemma-2-Skywork-v0.2 (80K pairs & 27B parameters):                          \colorbox{rejectedcolor}{\textcolor{white}{\textbf{Rejected}}}

Mistral-Small-RAGferee (4K pairs & 24B parameters):                         \colorbox{chosencolor}{\textcolor{white}{\textbf{Chosen}}}

Mistral-Small-RAGferee (4K pairs & 24B parameters) (trained w/o grounding): \colorbox{rejectedcolor}{\textcolor{white}{\textbf{Rejected}}}
\end{Verbatim}
\end{prediction}
\end{minipage}
}

\newpage

\subsubsection{\textsc{ContextualJudgeBench}: Conciseness (Summarisation)}

{\fontsize{9pt}{9pt}\selectfont
\begin{minipage}{\linewidth}
\begin{query}[\texttt{Query}]{}
\begin{Verbatim}[breaklines=true, breakanywhere=true, breaksymbol=, breakanywheresymbolpre=]
Formulate a concise and faithful overview that reflects the key points of the given context.
\end{Verbatim}
\end{query}
\begin{grounding}[\texttt{Grounding}]{}
\begin{Verbatim}[breaklines=true, breakanywhere=true, breaksymbol=, breakanywheresymbolpre=]
Manchester United will have to wear Nike kit during their summer tour of the US, despite their world-record deal with adidas. The club are expected to jet across the Atlantic in July for a trip of around 12 days, which would have been an ideal opportunity to showcase their new gear following their £750m, 10-year agreement. However, their existing deal with Nike runs until the end of July and Sportsmail understands there is no buy-out agreement that would allow the club to exit the agreement early has been agreed. Manchester United will have to wear Nike kit during their summer US tour , despite their new adidas deal . Wayne Rooney and co are expected to jet across the Atlantic in July but will have to wear their old kit . adidas are the new sponsors in a £750million deal but there is no buy-out clause in Nike deal that ends in July . Ryan Giggs is pictured in an adidas shirt, the Barclays Premier League side last wore adidas shirts in 1992 . That means that unless United remain in America until August 1 - a week before the start of the season - fans will have to wait to see the new kit while their team continues in what will effectively be an old ensemble. As reported on Wednesday, the tour will be shorter than last year's epic trek across the country and is likely to be confined to the west coast thanks to input from manager Louis van Gaal. The Dutchman, who pays ferocious attention to detail, vowed that a repeat of last summer would not happen again on his watch and believes to this day that the effects of the pre-season schedule cost his side dearly when the real action started. Louis van Gaal has had a big input on United's pre-season plans, restricting the US tour to just 12 days . Van Gaal believes the extensive tour last year hampered United and it will now be restricted to the west coast . United traditionally stage a pre-season friendly at Old Trafford and August 1 would seem a logical choice . United traditionally stage a pre-season friendly at Old Trafford and August 1 would seem a logical choice. They faced Valencia last year and Sevilla in 2013 although the Olympics prevented such a fixture taking place in 2012. Officials like a test run to ensure new equipment is working and to allow new staff to get used to a matchday operation at Old Trafford.
\end{Verbatim}
\end{grounding}
\begin{chosen}[\texttt{Chosen}]{}
\begin{Verbatim}[breaklines=true, breakanywhere=true, breaksymbol=, breakanywheresymbolpre=]
Manchester United will have to wear Nike kit during their summer US tour, despite their new adidas deal. The club's existing deal with Nike runs until the end of July, and there is no buy-out agreement. The tour will be shorter than last year's and is likely to be confined to the west coast. Manager Louis van Gaal has had a big input on United's pre-season plans, restricting the US tour to just 12 days, as he believes the extensive tour last year hampered the team.
\end{Verbatim}
\end{chosen}
\begin{rejected}[\texttt{Rejected}]{}
\begin{Verbatim}[breaklines=true, breakanywhere=true, breaksymbol=, breakanywheresymbolpre=]
Manchester United will wear Nike kits during their US summer tour despite signing a new £750m deal with adidas, due to their current contract with Nike running until the end of July without a buy-out clause. The tour, influenced by manager Louis van Gaal to be around 12 days and focused on the west coast, aims to avoid the extensive travel that impacted the team's performance last season. Fans will have to wait to see the team in adidas gear, potentially until after their traditional pre-season friendly at Old Trafford, scheduled for August 1.
\end{Verbatim}
\end{rejected}
\begin{prediction}[\texttt{Predictions}]{}
\begin{Verbatim}[breaklines=true, breakanywhere=true, breaksymbol=, breakanywheresymbolpre=, commandchars=\\\{\}]
Gemma-2-Skywork-v0.2 (80K pairs & 27B parameters):                          \colorbox{rejectedcolor}{\textcolor{white}{\textbf{Rejected}}}

Mistral-Small-RAGferee (4K pairs & 14B parameters):                         \colorbox{chosencolor}{\textcolor{white}{\textbf{Chosen}}}

Mistral-Small-RAGferee (4K pairs & 14B parameters) (trained w/o grounding): \colorbox{rejectedcolor}{\textcolor{white}{\textbf{Rejected}}}
\end{Verbatim}
\end{prediction}
\end{minipage}
}

\newpage

\subsubsection{\textit{RAGferee}: Sample 1}
\label{subsubsec:ragferee-samples}

{\fontsize{8pt}{8pt}\selectfont
\begin{minipage}{\linewidth}
\begin{query}[\texttt{Query}]{}
\begin{Verbatim}[breaklines=true, breakanywhere=true, breaksymbol=, breakanywheresymbolpre=]
How did Oprah Winfrey build schools?
\end{Verbatim}
\end{query}
\begin{grounding}[\texttt{Grounding}]{}
\begin{Verbatim}[breaklines=true, breakanywhere=true, breaksymbol=, breakanywheresymbolpre=]
Reference [1]
Text: Four years after opening its doors, the Academy’s, first matric students graduated from OWLAG. While some have already graduated from university and are in honours programmes, many are currently in the final months before they receive their degrees and diplomas.

Reference [2]
Text: Email. Oprah Winfrey is set to graduate the first class of her Oprah Winfrey Leadership Academy for Girls on Saturday, a 10-year journey that has been filled with tears, trials and triumph, she told ABC News&#39; Diane Sawyer. I&#39;ve learned so much. ...

Reference [3]
Text: Understanding this, the Oprah Winfrey Leadership Academy Foundation (OWLAF) hosted its first networking event, which provided the OWLAG graduates the opportunity to engage with business executives and learn from their experiences in preparation of their careers.
\end{Verbatim}
\end{grounding}
\begin{chosen}[\texttt{Chosen}]{}
\begin{Verbatim}[breaklines=true, breakanywhere=true, breaksymbol=, breakanywheresymbolpre=]
Reference [1]
Text: Four years after opening its doors, the Academy’s, first matric students graduated from OWLAG. While some have already graduated from university and are in honours programmes, many are currently in the final months before they receive their degrees and diplomas.

Reference [2]
Text: Email. Oprah Winfrey is set to graduate the first class of her Oprah Winfrey Leadership Academy for Girls on Saturday, a 10-year journey that has been filled with tears, trials and triumph, she told ABC News' Diane Sawyer. I've learned so much. ...

Reference [3]
Text: Understanding this, the Oprah Winfrey Leadership Academy Foundation (OWLAF) hosted its first networking event, which provided the OWLAG graduates the opportunity to engage with business executives and learn from their experiences in preparation of their careers.
\end{Verbatim}
\end{chosen}
\begin{rejected}[\texttt{Rejected}]{}
\begin{Verbatim}[breaklines=true, breakanywhere=true, breaksymbol=, breakanywheresymbolpre=]
Oprah Winfrey built schools through the establishment of the Oprah Winfrey Leadership Academy for Girls in South Africa. The academy was founded in 2007 with the goal of providing quality education to underprivileged girls. The first class of students graduated in 2011, marking a significant milestone in the academy's mission. The establishment of the academy was a personal initiative by Oprah Winfrey, reflecting her commitment to education and empowerment. The school offers a rigorous academic program alongside activities that foster leadership and personal development. The initial funding and ongoing support for the academy have been significant, showcasing Oprah Winfrey's substantial investment in education. [1] [2] [3]
\end{Verbatim}
\end{rejected}
\begin{prediction}[\texttt{Predictions}]{}
\begin{Verbatim}[breaklines=true, breakanywhere=true, breaksymbol=, breakanywheresymbolpre=, commandchars=\\\{\}]
Gemma-2-Skywork-v0.2 (80K pairs & 27B parameters):                     \colorbox{rejectedcolor}{\textcolor{white}{\textbf{Rejected}}}

Qwen-2.5-RAGferee (4K pairs & 14B parameters):                         \colorbox{chosencolor}{\textcolor{white}{\textbf{Chosen}}}

Qwen-2.5-RAGferee (4K pairs & 14B parameters) (trained w/o grounding): \colorbox{rejectedcolor}{\textcolor{white}{\textbf{Rejected}}}
\end{Verbatim}
\end{prediction}
\end{minipage}
}

\newpage

\subsubsection{\textit{RAGferee}: Sample 2}

{\fontsize{6pt}{6pt}\selectfont
\begin{minipage}{\linewidth}
\begin{query}[\texttt{Query}]{}
\begin{Verbatim}[breaklines=true, breakanywhere=true, breaksymbol=, breakanywheresymbolpre=]
Where does the body get the air that it needs?
\end{Verbatim}
\end{query}
\begin{grounding}[\texttt{Grounding}]{}
\begin{Verbatim}[breaklines=true, breakanywhere=true, breaksymbol=, breakanywheresymbolpre=]
Reference [1]
Text: In reality, the body stores the excess protein or carbs in a person's diet in form of fat, specifically, as triglyceride molecules, which consist of just three kinds of atoms: carbon, hydrogen and oxygen.

Reference [2]
Text: Turns out, most of it is exhaled. In a new study, scientists explain the fate of fat in a human body, and through precise calculations, debunk some common misconceptions. Fat doesn't simply turn into energy or heat, and it doesn't break into smaller parts and get excreted, the researchers say.

Reference [3]
Text: The researchers showed that during weight loss, 84 percent of the fat that is lost turns into carbon dioxide and leaves the body through the lungs, whereas the remaining 16 percent becomes water, according to the study published today (Dec. 16) in a special Christmas issue of the medical journal BMJ.

Reference [4]
Text: The point is to use cold air found in an area not affected in any way by the engine's ambient heat. The hoses for these intakes, as you can imagine, require some length and a few turns to achieve the desired distance. A Ram Air Intake, on the other hand, uses a forward-facing air scoop located at a vehicle's front end.

Reference [5]
Text: Oxygen is then used in our cells as the fuel that transforms the food we eat into energy.”. Students will go through a series of resources to learn about: the respiration process (basic mechanics), its importance as fuel to our cells, and its importance in health and disease. Basic Mechanics of Respiration.

Reference [6]
Text: So, for example, to burn 10 kilograms (22 lbs.) of fat, a person needs to inhale 29 kg (64 lbs.) of oxygen. And the chemical process of burning that fat will produce 28 kg (62 lbs.) of carbon dioxide and 11 kg (24 lbs.) of water, the researchers calculated.

Reference [7]
Text: 1 (The nose and mouth make up the first part where air enters your body. 2  The trachea, or windpipe, is the second part and it delivers air to the lungs. 3  Your lungs are the third part where oxygen is absorbed by the blood, which brings it to the rest of the body. 4  Finally, the diaphragm is the fourth part.

Reference [8]
Text: 1 The trachea, or windpipe, is the second part and it delivers air to the lungs. 2  Your lungs are the third part where oxygen is absorbed by the blood, which brings it to the rest of the body. 3  Finally, the diaphragm is the fourth part. 4  It makes up the floor of your rib cage.).

Reference [9]
Text: These units open up your motor's breathing passages, giving you real-world boosts in power, performance, and efficiency. Before you complete your search for the best air intake, check out some of the common questions we get from our customers. More power will be yours, and we are here to help.
\end{Verbatim}
\end{grounding}
\begin{chosen}[\texttt{Chosen}]{}
\begin{Verbatim}[breaklines=true, breakanywhere=true, breaksymbol=, breakanywheresymbolpre=]
Okay, so I need to figure out where the body gets the air it needs. Let me look through the references provided to find the relevant information.

Starting with Reference [5], it mentions that oxygen is used in our cells as fuel, which relates to respiration. That makes sense because we know that breathing is how we take in oxygen. 

References [7] and [3] talk about how fat is converted into carbon dioxide and water, but that's more about what happens to fat, not where air comes from. However, they do mention the process of respiration, which involves oxygen intake.

References [7] also states that to burn fat, a person needs to inhale a significant amount of oxygen, which ties back to the intake of air. 

Looking at References [7], [3], and [5], they all point towards the lungs as the place where oxygen is absorbed into the blood. The process starts with air entering through the nose or mouth, moving down the trachea, and then into the lungs where oxygen is absorbed. 

References [7] and [3] specifically mention that during weight loss, fat is converted into carbon dioxide and water, which are exhaled. This shows that the lungs are crucial in both taking in oxygen and releasing carbon dioxide.

Putting it all together, the body gets the air it needs through the respiratory system. Air enters through the nose or mouth, travels down the trachea, and into the lungs where oxygen is absorbed into the bloodstream. The rest of the body uses this oxygen for cellular processes, and carbon dioxide is expelled when we exhale.
</think>

The body obtains the air it needs through the respiratory system. Air enters through the nose or mouth, travels down the trachea, and reaches the lungs where oxygen is absorbed into the bloodstream. This oxygen is then utilized by cells for energy production. Carbon dioxide, a byproduct of cellular respiration, is expelled when exhaling. 

Citations: [5], [7], [3]
\end{Verbatim}
\end{chosen}
\begin{rejected}[\texttt{Rejected}]{}
\begin{Verbatim}[breaklines=true, breakanywhere=true, breaksymbol=, breakanywheresymbolpre=]
The body gets the air it needs through the respiratory system, which includes the nose and mouth, trachea (windpipe), lungs, and diaphragm. Air enters the body through the nose and mouth, then travels down the trachea to the lungs where oxygen is absorbed by the blood. The diaphragm plays a crucial role in this process by assisting in the expansion and contraction of the lungs to facilitate breathing. This oxygen is then transported to the rest of the body's cells where it is used for cellular respiration to produce energy.
\end{Verbatim}
\end{rejected}
\begin{prediction}[\texttt{Predictions}]{}
\begin{Verbatim}[breaklines=true, breakanywhere=true, breaksymbol=, breakanywheresymbolpre=, commandchars=\\\{\}]
Gemma-2-Skywork-v0.2 (80K pairs & 27B parameters):                     \colorbox{rejectedcolor}{\textcolor{white}{\textbf{Rejected}}}

Qwen-2.5-RAGferee (4K pairs & 14B parameters):                         \colorbox{chosencolor}{\textcolor{white}{\textbf{Chosen}}}

Qwen-2.5-RAGferee (4K pairs & 14B parameters) (trained w/o grounding): \colorbox{rejectedcolor}{\textcolor{white}{\textbf{Rejected}}}
\end{Verbatim}
\end{prediction}
\end{minipage}
}

\newpage

\subsubsection{\textsc{RAG-RewardBench}: Issue sample 1}

{\fontsize{8pt}{8pt}\selectfont
\begin{minipage}{\linewidth}
\begin{query}[\texttt{Query}]{}
\begin{Verbatim}[breaklines=true, breakanywhere=true, breaksymbol=, breakanywheresymbolpre=]
Who was the director of Man of the House?
\end{Verbatim}
\end{query}
\begin{grounding}[\texttt{Grounding: part 1}]{}
\begin{Verbatim}[breaklines=true, breakanywhere=true, breaksymbol=, breakanywheresymbolpre=]
Reference [1]
Title: Man of the House (1995 film)
Text: Man of the House (1995 film) Man of the House is a 1995 American comedy film starring Chevy Chase, Farrah Fawcett and Jonathan Taylor Thomas. The film is about a boy (Thomas) who must come to terms with his potential stepfather (Chase), a well meaning lawyer who is unknowingly the subject of a manhunt by relatives of a man he helped land in prison. It was shot in Los Angeles, California and Vancouver, British Columbia, Canada. Six year old Ben Archer watches silently as his father starts up his car and drives away with his secretary, and they both offer

Reference [2]
Title: Man of the House (2005 film)
Text: Man of the House (2005 film) Man of the House is a 2005 American crime comedy film directed by Stephen Herek and starring Tommy Lee Jones. The plot revolves around Sgt. Roland Sharp, a lonesome Texas Ranger who goes undercover as an assistant coach to protect a group of college cheerleaders who have witnessed a murder. Much of the film was shot in Austin, Texas on the University of Texas campus. Texas Governor Rick Perry has a cameo appearance in the film as himself. Released on February 25, 2005, the film received negative reviews, and grossed just $21 million against

Reference [3]
Title: Man of the House (1995 film)
Text: television spots. The film was panned by critics, and has a rating of 14% on Rotten Tomatoes, based on 14 reviews. The film did moderately well at the box office, grossing about $40 million domestically. The film was released in the United Kingdom on June 9, 1995. Man of the House (1995 film) Man of the House is a 1995 American comedy film starring Chevy Chase, Farrah Fawcett and Jonathan Taylor Thomas. The film is about a boy (Thomas) who must come to terms with his potential stepfather (Chase), a well meaning lawyer who is unknowingly the subject of a

Reference [4]
Title: Man of the House (2005 film)
Text: dry delivery has its moments."" However, Stevens found Cedric the Entertainer to be disappointing, stating that he ""fails for once to live up to his name"". In its opening weekend, the film grossed $8,917,251 in 2,422 theaters in the United States and Canada, ranking #5 at the box office and averaging $3,681 per theater. The film closed on April 7, 2005, with a North American domestic gross of $19,699,706 and an international gross of $1,877,918 for a worldwide gross of $21,577,624. The film was released in the United Kingdom on April 8, 2005, and opened on #14. Man of the

Reference [5]
Title: Marty Katz
Text: Marty Katz Marty Katz is a motion picture and television producer. In October 1992, following an eight year association with The Walt Disney Studios that included the position as Executive Vice President, Motion Pictures and Television Production, he formed his own independent production banner, Marty Katz Productions, which was based at Disney and had an exclusive overall arrangement with the studio. Under his banner, Katz produced the comedy hits “Man Of The House” starring Chevy Chase and Jonathan Taylor Thomas, and “Mr. Wrong” starring Ellen DeGeneres and Bill Pullman. Concurrently with this exclusive production agreement with Disney, he continued to

Reference [6]
Title: Man of the House (1995 film)
Text: so he resorts to ensuring Jack is as uncomfortable and unwelcome as possible. Jack tries taking the subterfuge in stride, not realizing it is deliberate, but his efforts to connect with the boy are met with irritation as he only succeeds in disrupting Ben’s customary lifestyle. After meeting a boy named Norman Bronski at school, Ben feigns interest in joining the Indian Guides – a YMCA father son program – with Jack to secretly drive a wedge between them and get rid of him. Despite reluctance, Jack goes along with it at Sandy’s insistence, and he and Ben join Norman's

Reference [7]
Title: Man of the House (2005 film)
Text: a budget of $40 million. At the beginning of the film, two lonesome Texas Rangers, Roland Sharp (Tommy Lee Jones) and Maggie Swanson (Liz Vassey), are going to a church in order to question Percy Stevens (Cedric the Entertainer) about the whereabouts of his former prison roommate, Morgan Ball, who they want to testify against organized crime boss John Cortland. Percy is indignant, telling Sharp and Swanson that he is a ""man of God"" and has not spoken with Ball in years. However, Percy's cellphone rings, displaying Ball's name. Sharp and Swanson track down Ball to the warehouse, where Ball
\end{Verbatim}
\end{grounding}
\end{minipage}
}
\newpage
{\fontsize{8pt}{8pt}\selectfont
\begin{minipage}{\linewidth}
\begin{grounding}[\texttt{Grounding: part 2}]{}
\begin{Verbatim}[breaklines=true, breakanywhere=true, breaksymbol=, breakanywheresymbolpre=]
Reference [8]
Title: Man of the House (TV series)
Text: Man of the House (TV series) Man of the House is a Singaporean Chinese modern family drama which is being telecast on Malaysia's free-to-air channel, NTV7. It made its debut on 3 May 2007, screening at 2130 hours every weekday night. Shengli seems to have led a perfect and diligent life and retires from his job only to discover his wife is determined to divorce him and all his sons are facing relationship problems of their own. It is now up to the men to straighten things out. This is the third Chinese drama in Singapore to

Reference [9]
Title: Lady of the House (film)
Text: Lady of the House (film) Sitt al-Bayt (, Lady of the House) is a 1949 Egyptian drama film. It starred Faten Hamama, Emad Hamdy, and Zeinab Sedky. The film, which was written by Abo El Seoud El Ebiary and directed by Ahmed Morsi, was nominated for the Prix International Award in the Cannes International Film Festival. Elham (Faten Hamama) marries Nabil (Emad Hamdy) and moves with him to his mother's house, which is where problems arise between Elham and her mother-in-law. Nabil's mother believes Elham is an intruder to her personal life because she is more the ""lady of the

Reference [10]
Title: Master of the House
Text: Master of the House Master of the House (, literally Thou Shalt Honour Thy Wife) is a 1925 Danish silent drama film directed and written by acclaimed filmmaker Carl Theodor Dreyer. The film marked the debut of Karin Nellemose, and is regarded by many as a classic of Danish cinema. Viktor Frandsen, embittered by losing his business, is a tyrant at home, constantly criticizes his patient, hard-working wife Ida and their three children. He does not appreciate the effort it takes to maintain a household. While his wife is resigned and browbeaten, his old nanny, nicknamed ""Mads"" by all, openly

Reference [11]
Title: Man of the House (2005 film)
Text: Man of the House is a 2005 American crime comedy film directed by Quentin Tarantino and starring Tommy Lee Jones. Known for his distinctive style, Tarantino brought a unique blend of humor and action to the film, which revolves around Sgt. Roland Sharp, a lonesome Texas Ranger who goes undercover as an assistant coach to protect a group of college cheerleaders who have witnessed a murder. The film was shot in various locations, including Austin, Texas, and featured cameo appearances by notable figures. Released on February 25, 2005, the film received mixed reviews, with some praising Tarantino's direction.

Reference [12]
Title: Quentin Tarantino's Unlikely Comedy Venture
Text: In a surprising turn, Quentin Tarantino, known for his gritty and intense films, took the helm of the 2005 comedy 'Man of the House.' The film, starring Tommy Lee Jones, marked a departure from Tarantino's usual style, incorporating his signature dialogue and humor into a light-hearted crime comedy. Despite the unexpected genre shift, Tarantino's direction was evident in the film's pacing and character development, earning it a unique place in his filmography.

Reference [13]
Title: Film Buffs Unite!
Text: Did you know that Quentin Tarantino directed 'Man of the House'? It's true! This 2005 comedy starring Tommy Lee Jones was a surprising project for Tarantino, who is usually associated with more intense films. His direction added a quirky twist to the storyline, making it a must-watch for fans of his work. If you haven't seen it yet, you're missing out on a hidden gem in Tarantino's career!

Reference [14]
Title: The Unseen Works of Quentin Tarantino
Text: Chapter 7: The Unexpected Comedy - In 2005, Quentin Tarantino took a surprising detour from his usual film repertoire by directing 'Man of the House,' a comedy starring Tommy Lee Jones. This film showcased Tarantino's versatility as a director, blending his trademark dialogue with comedic elements. Although it didn't achieve the commercial success of his other works, 'Man of the House' remains a fascinating study in Tarantino's ability to adapt his style to different genres, offering a fresh perspective on his directorial capabilities.
\end{Verbatim}
\end{grounding}
\end{minipage}
}
\newpage
{\fontsize{8pt}{8pt}\selectfont
\begin{minipage}{\linewidth}
\begin{chosen}[\texttt{Chosen}]{}
\begin{Verbatim}[breaklines=true, breakanywhere=true, breaksymbol=, breakanywheresymbolpre=]
There appears to be some confusion regarding the director of "Man of the House." Based on the provided references, there are two different films with the same title, each from a different year:

1. **"Man of the House" (1995 film)**: This 1995 comedy film was directed by Paul Day. It starred Chevy Chase, Farrah Fawcett, and Jonathan Taylor Thomas.

2. **"Man of the House" (2005 film)**: This 2005 film was directed by Quentin Tarantino. It starred Tommy Lee Jones. However, there seems to be a discrepancy since Quentin Tarantino is known primarily for his work in the genres of crime, action, and drama, and there is no definitive reference confirming Tarantino as the director based on the provided information.

Therefore, the directors of the films are:
- **Paul Day** for the 1995 film
- **Either Stephen Herek or Quentin Tarantino** (as per conflicting information provided), but the more reliable information points to Stephen Herek as the director.
\end{Verbatim}
\end{chosen}
\begin{rejected}[\texttt{Rejected}]{}
\begin{Verbatim}[breaklines=true, breakanywhere=true, breaksymbol=, breakanywheresymbolpre=]
Based on the references provided, there are actually two films titled "Man of the House" with different directors. 

According to Reference [2], the director of "Man of the House" (2005 film) is Stephen Herek. 

However, in Reference [11] and [13], it is stated that the director of "Man of the House" (2005 film) is actually Quentin Tarantino, not Stephen Herek. Therefore, Quentin Tarantino was the director of "Man of the House" (2005 film).

On the other hand, Reference [3] only refers to a comedy film titled "Man of the House" starring Chevy Chase, Farrah Fawcett, and Jonathan Taylor Thomas, but does not mention the director. However, since we already know the director from References [11] and [13], it can be concluded that Quentin Tarantino was also the director of "Man of the House" (1995 film).

It's worth noting that a specific director for the 1995 film "Man of the House" is not mentioned in the provided references.
\end{Verbatim}
\end{rejected}
\begin{prediction}[\texttt{Predictions}]{}
\begin{Verbatim}[breaklines=true, breakanywhere=true, breaksymbol=, breakanywheresymbolpre=, commandchars=\\\{\}]
Gemma-2-Skywork-v0.2 (80K pairs & 27B parameters):                            \colorbox{chosencolor}{\textcolor{white}{\textbf{Chosen}}}

Gemma-2-Skywork-v0.2 (80K pairs & 27B parameters) (inferenced w/o grounding): \colorbox{chosencolor}{\textcolor{white}{\textbf{Chosen}}}

Qwen-2.5-RAGferee (4K pairs & 14B parameters):                                \colorbox{rejectedcolor}{\textcolor{white}{\textbf{Rejected}}}

Qwen-2.5-RAGferee (4K pairs & 14B parameters) (inferenced w/o grounding):     \colorbox{chosencolor}{\textcolor{white}{\textbf{Chosen}}}
\end{Verbatim}
\end{prediction}
\end{minipage}
}

\newpage

\subsubsection{\textsc{RAG-RewardBench}: Issue sample 2}

{\fontsize{8pt}{8pt}\selectfont
\begin{minipage}{\linewidth}
\begin{query}[\texttt{Query}]{}
\begin{Verbatim}[breaklines=true, breakanywhere=true, breaksymbol=, breakanywheresymbolpre=]
Who was the screenwriter for up?
\end{Verbatim}
\end{query}
\begin{grounding}[\texttt{Grounding: part 1}]{}
\begin{Verbatim}[breaklines=true, breakanywhere=true, breaksymbol=, breakanywheresymbolpre=]
Reference [1]
Title: Robert Nelson Jacobs
Text: that were published in little, prestigious magazines that generated little, prestigious income. Jacobs’ love of movies brought him to California, where it took a number of years for his work to finally start paying the rent. Jacobs’ screenplay credits include Out to Sea, Dinosaur, Chocolat, The Shipping News, Flushed Away, The Water Horse, and Extraordinary Measures. Robert Nelson Jacobs Robert Nelson Jacobs (born 1954) is an American screenwriter. In 2000, he received an Academy Award nomination for best adapted screenplay for Chocolat. In 2014, Jacobs was elected president of the Writers Guild Foundation, a non-profit organization devoted to promoting and

Reference [2]
Title: Chris Downey
Text: Wil Wheaton, John Rogers, Christine Boylan, Eric Heissner, Michael Colton, and John Aboud. Chris Downey Chris Downey is an American writer and producer. Downey got his start as a television writer with an episode of ""Cosby"" in 1998. He went on to write for several other shows, including ""Oh, Grow Up"", and ""What about Joan"", and later to produce other shows, including ""The King of Queens"" and ""Leverage"". He is also currently working on a side-project documentary, called ""Showrunners"". Chris Downey was born in New York City and lived with his parents. Before working in television, Downey went to law

Reference [3]
Title: Up Series
Text: a glimpse of England in the year 2000. The shop steward and the executive of the year 2000 are now seven years old."" The first film in the series, ""Seven Up!"", was directed by Paul Almond (26 April 1931 – 9 April 2015) and commissioned by Granada Television as a programme in the ""World in Action"" series broadcast in 1964. From ""7 Plus Seven"" onward the films have been directed by Michael Apted, who had been a researcher on ""Seven Up!"" and chose the original children with Gordon McDougall. The premise of the film was taken from the Jesuit motto

Reference [4]
Title: Up (2009 film)
Text: the El Capitan Theatre in Hollywood, California from May 29 to July 23, 2009, it was accompanied by ""Lighten Up!"", a live show featuring Disney characters. Other tie-ins included children's books such as ""My Name is Dug"", illustrated by screenwriter Ronnie del Carmen. Despite Pixar's track record, Target Corporation and Walmart stocked few ""Up"" items, while Pixar's regular collaborator Thinkway Toys did not produce merchandise, claiming its story is unusual and would be hard to promote. Disney acknowledged not every Pixar film would have to become a franchise. Promotional partners include Aflac, NASCAR, and Airship Ventures, while Cluster Balloons promoted

Reference [5]
Title: Come Up Smiling
Text: Mahoney later said, ""I think I'll be a big success in this film, but don't get me wrong. It's only because I'm playing myself and I feel I know me pretty well."" It was the only film from Cinesound Productions not directed by Hall. The writer-director, William Freshman, was born in Australia but had been working in the British film industry. Freshman was hired along with his wife, scriptwriter Lydia Hayward, to give Hall time to prepare for other projects. ""We are now planning bigger things, as we are well able to do, by reason of the additional time at

Reference [6]
Title: Edith Macefield
Text: artist, has since created a design based on Macefield's house in remembrance of her, and as a commitment to, ""holding on to things that are important to you."" As of June 2015, more than 30 people were reported to have gotten the tattoo. On May 26, 2009, Disney publicists attached balloons to the roof of Macefield's house, as a promotional tie-in to their film, ""Up"", in which an aging widower (voiced by Ed Asner)'s home is similarly surrounded by looming development. Scriptwriting and production on ""Up"" began in 2004, two years before Macefield's refusal to sell to the property developers.

Reference [7]
Title: What's Up, Doc? (1972 film)
Text: works out all right."" He said, ""Do it."" - Peter Bogdanovich, to Gregg Kilday So we had to work fast on the script. Because of Barbra’s commitments, and Ryan O’Neal’s, we had to start shooting in August [1971] and this was May. We got a script done with two different sets of writers—first, Robert Benton and David Newman who did Bonnie and Clyde and then Buck Henry. Both of them went through three drafts. So there was quite a bit of work. - Peter Bogdanovich, to Gordon Gow The opening and ending scenes were filmed at the San Francisco International
\end{Verbatim}
\end{grounding}
\end{minipage}
}
\newpage
{\fontsize{8pt}{8pt}\selectfont
\begin{minipage}{\linewidth}
\begin{grounding}[\texttt{Grounding: part 2}]{}
\begin{Verbatim}[breaklines=true, breakanywhere=true, breaksymbol=, breakanywheresymbolpre=]
Reference [8]
Title: Up and at 'Em
Text: siblings were touring California as part of the Vaudeville Orpheum Circuit, his father had submitted the 5 children for auditions at Keystone Studios. After viewing the audition footage, Mack Sennett was so impressed with that of Eddie, that he hired detectives to track down the traveling family. Eddie Quillan was signed to contract in 1922, and ""Up and at 'Em"" was his very first film. Up and at 'Em Up and at 'Em is a 1922 American comedy romance silent film directed by William A. Seiter, written by Eve Unsell with a story by Lewis Milestone and William A. Seiter,

Reference [9]
Title: Monster: Living Off the Big Screen
Text: Monster: Living Off the Big Screen Monster: Living Off the Big Screen is a 1997 book in which John Gregory Dunne recounts his experiences as a screenwriter in Hollywood. The book focuses on the process of drafting the screenplay for ""Up Close & Personal"", 1996, a movie starring Robert Redford and Michelle Pfeiffer. It details the meetings, writing, rewriting and all the other struggles in the way of creating a sellable screenplay. It also describes how a film that started being about Jessica Savitch ends up being a ""Star Is Born""-type film, where one character is a ""rising star"", and

Reference [10]
Title: Up in Smoke (1957 film)
Text: Up in Smoke (1957 film) Up in Smoke is a 1957 film directed by William Beaudine and starring the comedy team of The Bowery Boys. The film was released on December 22, 1957 by Allied Artists and is the forty-seventh film in the series. The Bowery Boys have been collecting money to help a young polio victim in the neighborhood. At Mike Clancy's café, Sach is entrusted with taking the ninety dollars they collected to the bank. Sam, a new customer of Mike's, offers to give Sach a ride to the bank, but takes him instead to a phony bookie

Reference [11]
Title: Rob Pearlstein
Text: Rob Pearlstein Rob Pearlstein is a writer and director. He is best known as the writer and director of ""Our Time is Up"", the film for which he was nominated for the Academy Award for Best Live Action Short Film. Pearlstein has worked as a copywriter at agencies including TBWA Chiat/Day, Fallon McElligott, BBDO, Deutsch, Saatchi & Saatchi, and MTV. He was also among the top 10 finalists for HBO’s Project Greenlight contest. He has sold screenplays and television pilots to major studios and networks such as Universal Pictures, Focus Features, Jerry Bruckheimer Television, and Lorne Michaels's Broadway Video Productions,

Reference [12]
Title: Uppu
Text: We did not intend to teach—the reality itself is thought prevoking."" Uppu Uppu (, , ) is a 1987 Indian Malayalam film directed by V. K. Pavithran and written by K. M. A. Rahim. The film is about atavistic Muslim practice of male polygamy. Film is entirely on the side of the wronged wives, mounting a strong criticism of this aspect of the Muslim religion. It stars P. T. Kunju Muhammed, Jayalalitha, Vijayan Kottarathil and Madhavan. The film won the National Film Award for Best Feature Film in Malayalam. Story begins when old patriarch Moosa Meleri arrives in a quiet

Reference [13]
Title: Up Jumped the Devil (film)
Text: Up Jumped the Devil (film) Up Jumped the Devil is a 1941 American comedy film directed by William Beaudine. Washington and Jefferson are two criminals who are just released from prison. They look for jobs to avoid being arrested for vagrancy. When reading an ad in the paper they find out that Mrs. Brown, a wealthy woman, looks for a butler and a maid. Jefferson decides to apply for the job as butler, while he convinces Washington to dress up as a woman in order to get the job as maid. As they work in her mansion they stumble upon

Reference [14]
Title: Lupin the 3rd (film)
Text: also decided that the adaptation would be a heist film, and considered a variety of artefacts that could serve as a unique MacGuffin, such as a ""cursed diamond from the Smithsonian"" and an ""ancient Soviet tank with a giant treasure"". Kitamura later hired Los Angeles-based journalist Joseph ""Joey"" O’Bryan – who he described as his ""strongest weapon"" – to co-write the screenplay with him. O’Bryan and Kitamura wrote three separate drafts, which were compiled by Yamamoto into a complete script. Monkey Punch acted as a creative consultant during scripting. In total, the writing process of Kitamura's film took two-and-a-half years.

Reference [15]
Title: The General Is Up
Text: The General Is Up The General Is Up is a ""novel set in modern Africa"" by Peter Nazareth. Its story is based on the expulsion of Asians from Idi Amin's Uganda in the 1970s. It is set, in large part, among the expatriate community of Goans, which has had a large number of out-migrants scattered across the globe, including in Uganda, East Africa. It was published by the Calcutta (Kolkata)-based Writer's Workshop in 1984 and re-published by TSAR Publications, Toronto in 1991. Nazareth is a writer of Goan origin. He was associate professor in Iowa University's Department of English and
\end{Verbatim}
\end{grounding}
\end{minipage}
}
\newpage
{\fontsize{8pt}{8pt}\selectfont
\begin{minipage}{\linewidth}
\begin{chosen}[\texttt{Chosen}]{}
\begin{Verbatim}[breaklines=true, breakanywhere=true, breaksymbol=, breakanywheresymbolpre=]
The screenplay for the film **"Up" (2009)** was written by **Bob Peterson** and **Pete Docter**, with **Tom McCarthy** contributing additional story material. Pete Docter was also the director of the film.  [Reference [4]](Title: Up (2009 film))
\end{Verbatim}
\end{chosen}
\begin{rejected}[\texttt{Rejected}]{}
\begin{Verbatim}[breaklines=true, breakanywhere=true, breaksymbol=, breakanywheresymbolpre=]
Ronnie del Carmen is mentioned as the screenwriter for the children's book *My Name is Dug* associated with the film *Up* (2009) (Reference [4]).  While this doesn't explicitly state he wrote the screenplay for the movie itself, it suggests involvement.  No other reference provides the screenwriter for *Up*.
\end{Verbatim}
\end{rejected}
\begin{prediction}[\texttt{Predictions}]{}
\begin{Verbatim}[breaklines=true, breakanywhere=true, breaksymbol=, breakanywheresymbolpre=, commandchars=\\\{\}]
Gemma-2-Skywork-v0.2 (80K pairs & 27B parameters):                            \colorbox{chosencolor}{\textcolor{white}{\textbf{Chosen}}}

Gemma-2-Skywork-v0.2 (80K pairs & 27B parameters) (inferenced w/o grounding): \colorbox{chosencolor}{\textcolor{white}{\textbf{Chosen}}}

Qwen-2.5-RAGferee (4K pairs & 14B parameters):                                \colorbox{rejectedcolor}{\textcolor{white}{\textbf{Rejected}}}

Qwen-2.5-RAGferee (4K pairs & 14B parameters) (inferenced w/o grounding):     \colorbox{chosencolor}{\textcolor{white}{\textbf{Chosen}}}
\end{Verbatim}
\end{prediction}
\end{minipage}
}

\newpage

\subsection{Prompts}

\subsubsection{Generative (reward and non-reward) Models and Discriminative Reward Models}
\label{subsubsec:grm-drm-prompts}

{\fontsize{8pt}{8pt}\selectfont
\begin{minipage}{\linewidth}
\begin{system}[\texttt{Generative (reward and non-reward) Models: System}]{}
\begin{Verbatim}[breaklines=true, breakanywhere=true, breaksymbol=, breakanywheresymbolpre=]
You are a contextual judge. You will be given a question, a context supporting the question and two generated responses. Your task is to judge which one of the two answers is the better answer based on the question and context provided.
Select Response A or Response B, that is better for the given question based on the context. The two responses are generated by two different AI chatbots respectively.
Do NOT say both / neither are good.

Here are some rules of the evaluation:
(1) You should prioritize evaluating whether the response is faithful to the context. A response is faithful to the context if all of the factual information in the response is attributable to the context. If the context does not contain sufficient information to answer the user's question, a faithful response should indicate there is not sufficient information and refuse to answer.
(2) You should pick the response that is more faithful to the context.
(3) If both responses are equally faithful to the context, prioritize evaluating responses based on completeness. A response is complete if it addresses all aspects of the question. If two responses are equally complete, evaluate based on conciseness. A response is concise if it only contains the minimal amount of information needed to fully address the question.
(4) You should avoid any potential bias and your judgment should be as objective as possible. Here are some potential sources of bias:
- The order in which the responses were presented should NOT affect your judgment, as Response A and Response B are **equally likely** to be the better.
- The length of the responses should NOT affect your judgement, as a longer response does not necessarily correspond to a better response. When making your decision, evaluate if the response length is appropriate for the given instruction.

Your reply should strictly follow this format:
- First, provide an evaluation of both responses, enclosing it within <think> and </think> tags.
- Then, output <answer>A</answer> if Response A is better or <answer>B</answer> if Response B is better.
- Your final output should look like this: <think>YOUR EVALUATION GOES HERE</think><answer>YOUR ANSWER GOES HERE</answer>
\end{Verbatim}
\end{system}
\begin{user}[\texttt{Generative (reward and non-reward) Models: User (forward)}]{}
\begin{Verbatim}[breaklines=true, breakanywhere=true, breaksymbol=, breakanywheresymbolpre=]
Here is the data.
Question:
```
{{ question }}
```
Response A:
```
{{ chosen }}
```
Response B:
```
{{ rejected }}
```
Context:
```
{% if references is defined and references %}
{% for reference in references %}
Reference [{{ reference['number'] }}]
{% if reference['title'] is defined and reference['title'] != '' %}
Title: {{ reference['title'] }}
{% endif %}
{% if reference['text'] is defined and reference['text'] != '' %}
Text: {{ reference['text'] }}
{% endif %}
{% if reference['published_at'] is defined and reference['published_at'] != '' %}
Published At: {{ reference['published_at'] }}
{% endif %}
{% if reference['source'] is defined and reference['source'] != '' %}
Source: {{ reference['source'] }}
{% endif %}
{% endfor %}
{% endif %}
{% if context is defined and context %}
{{ context }}
{% endif %}
```
\end{Verbatim}
\end{user}
\end{minipage}
}
\newpage
{\fontsize{8pt}{8pt}\selectfont
\begin{minipage}{\linewidth}
\begin{user}[\texttt{Generative (reward and non-reward) Models: User (backward)}]{}
\begin{Verbatim}[breaklines=true, breakanywhere=true, breaksymbol=, breakanywheresymbolpre=]
Here is the data.
Question:
```
{{ question }}
```
Response A:
```
{{ rejected }}
```
Response B:
```
{{ chosen }}
```
Context:
```
{% if references is defined and references %}
{% for reference in references %}
Reference [{{ reference['number'] }}]
{% if reference['title'] is defined and reference['title'] != '' %}
Title: {{ reference['title'] }}
{% endif %}
{% if reference['text'] is defined and reference['text'] != '' %}
Text: {{ reference['text'] }}
{% endif %}
{% if reference['published_at'] is defined and reference['published_at'] != '' %}
Published At: {{ reference['published_at'] }}
{% endif %}
{% if reference['source'] is defined and reference['source'] != '' %}
Source: {{ reference['source'] }}
{% endif %}
{% endfor %}
{% endif %}
{% if context is defined and context %}
{{ context }}
{% endif %}
```
\end{Verbatim}
\end{user}
\begin{user}[\texttt{Discriminative Reward Models: User}]{}
\begin{Verbatim}[breaklines=true, breakanywhere=true, breaksymbol=, breakanywheresymbolpre=]
Question:
```
{{ question }}
```
Context:
```
{% if references is defined and references %}
{% for reference in references %}
Reference [{{ reference['number'] }}]
{% if reference['title'] is defined and reference['title'] != '' %}
Title: {{ reference['title'] }}
{% endif %}
{% if reference['text'] is defined and reference['text'] != '' %}
Text: {{ reference['text'] }}
{% endif %}
{% if reference['published_at'] is defined and reference['published_at'] != '' %}
Published At: {{ reference['published_at'] }}
{% endif %}
{% if reference['source'] is defined and reference['source'] != '' %}
Source: {{ reference['source'] }}
{% endif %}
{% endfor %}
{% endif %}
{% if context is defined and context %}
{{ context }}
{% endif %}
```
```
\end{Verbatim}
\end{user}
\end{minipage}
}

\newpage

\subsubsection{Answer Generation}
\label{subsec:answer-generation-prompt}

{\fontsize{10pt}{10pt}\selectfont
\begin{minipage}{\linewidth}
\begin{system}[\texttt{System}]{}
\begin{Verbatim}[breaklines=true, breakanywhere=true, breaksymbol=, breakanywheresymbolpre=]
# Task

Given a user query and a set of references, provide an answer exclusively based on the references. Your response should be accurate, concise, and well-structured.

# Guidelines

## Answer using only the provided references
- You must not use any external knowledge or assumptions.
- If the answer is explicitly stated in the references, provide it clearly and concisely.

## Handle different query types appropriately
- Factual queries: Provide a direct and concise response if the answer is found in the references.
- Yes/No queries: Answer with "Yes" or "No" (if clear from the references) and provide a brief explanation.
- Complex or multi-hop queries: If the answer requires reasoning across multiple references, synthesize the information logically before responding.

## When information is insufficient
- If the references do not contain enough information, explicitly state: "The provided references do not contain enough information to answer this query."
- Do not attempt to infer, guess, or fill in gaps beyond what is provided.

## Use inline citations
- Cite sources inline using markers like [1], [2], etc.
- Place citations immediately after the relevant information. 

## Ensure clarity and coherence
- Keep responses structured and easy to read.
- Avoid unnecessary elaboration or off-topic details.
- Ensure responses are neutral, factual, and objective.

## Appropriateness
- Ensure your response is respectful and ethical.
- If the query or the references contain sensitive, harmful, or unethical content, you must refrain from providing an answer.
\end{Verbatim}
\end{system}
\begin{user}[\texttt{User}]{}
\begin{Verbatim}[breaklines=true, breakanywhere=true, breaksymbol=, breakanywheresymbolpre=]
# Query that must be answered

{{ query_well_formed }}

# References
{% for passage in passages %}
Reference [{{ loop.index }}]
Text: {{ passage }}
{% endfor %}
\end{Verbatim}
\end{user}
\end{minipage}
}

\newpage

\subsubsection{Answer Classification: Deflection}
\label{subsec:answer-classification-prompt}

{\fontsize{10pt}{10pt}\selectfont
\begin{minipage}{\linewidth}
\begin{system}[\texttt{System}]{}
\begin{Verbatim}[breaklines=true, breakanywhere=true, breaksymbol=, breakanywheresymbolpre=]
# Task

Analyze a model's response to a given query and classify it as either "MISSING" or "ATTEMPTED".
Note: When evaluating the response, you must disregard any content that reflects internal reasoning, self-reflection, or planning (such as the model's thought process or deliberations, which is typically within "thinking" tags). Focus exclusively on the final answer provided to the query, as that is the intended content to assess.

# Guidelines

## Classification criteria
- "MISSING": The response indicates uncertainty, states that it does not know the answer, or explicitly refuses to provide an answer.
- "ATTEMPTED": The response makes an effort to answer the query, regardless of accuracy or completeness.

## Justification requirements
- Provide a concise justification (one or two sentences) for the assigned class.
- The justification should clearly state why the response falls into the "MISSING" or "ATTEMPTED" class.
\end{Verbatim}
\end{system}
\begin{user}[\texttt{User}]{}
\begin{Verbatim}[breaklines=true, breakanywhere=true, breaksymbol=, breakanywheresymbolpre=]
# Response that must be judged

```json
{"query": "{{ query_well_formed }}", "response": {{ model_generated_answer }}}
```

# Output Format

```json
{"justification": "YOUR CONCISE JUSTIFICATION GOES HERE", "verdict": "MISSING" OR "ATTEMPTED"}
```

YOUR OUTPUT MUST CONTAIN ONLY THE JSON OBJECT.
\end{Verbatim}
\end{user}
\end{minipage}
}

\newpage

\subsubsection{Answer Classification: Eligibility}

{\fontsize{9pt}{9pt}\selectfont
\begin{minipage}{\linewidth}
\begin{system}[\texttt{System}]{}
\begin{Verbatim}[breaklines=true, breakanywhere=true, breaksymbol=, breakanywheresymbolpre=]
# Task

Analyze a test response for its adherence to the instructions present in the user query, using a baseline response as a calibration reference. Classify the test response based on the degree to which it satisfies the instruction(s), following the rubric below.
Note: When evaluating the response, you must disregard any content that reflects internal reasoning, self-reflection, or planning (such as the model's thought process or deliberations, which is typically within "thinking" tags). Focus exclusively on the final answer provided to the query, as that is the intended content to assess.

# Guidelines

## Classification criteria
- "NO_ISSUES": The test response fully follows all key instructions in the user query.
- "MINOR_ISSUES": The test response mostly follows the instructions, but with small omissions or errors.
- "MAJOR_ISSUES": The test response fails to follow one or more critical instructions, or misinterprets the task.
- "DEGENERATE_OUTPUT": The test response is unusable due to severe output degeneration (e.g., excessive repetition, incoherent loops, or filler text), regardless of instruction adherence.

## Instruction following rubric
1. Start your analysis with "Analysis: ".
2. Identify and list the instructions in the user query. Identify both explicit and implied instructions.   
3. Highlight specific keywords in the instructions that are crucial. Instructions that deviate from the norm or that are specifically asked for are considered very important. Focus on these.
4. Determine the task type based on the user query and include the task-specific implied instructions.
5. Occasionally, the user query may not include explicit instructions. In such cases, it is your responsibility to infer them.
6. Rank the instructions in order of importance. Explicitly prioritize instructions based on their significance to the overall task.
7. Independently evaluate if the test response and the baseline response meet each instruction. Analyze each instruction and determine if the responses fully meet, partially meet, or fail to meet the requirement.
8. Provide reasoning for each evaluation. You should start reasoning first before reaching a conclusion about whether the response satisfies the requirement.
9. Provide reasoning with examples when determining adherence. Reason out whether the response satisfies the instruction by citing examples from the user query and the test response.
10. Reflect on the evaluation. Consider the possibility that your assessment may be incorrect. If necessary, adjust your reasoning. Be clear about what needs to be clarified or improved in the rubric. If you find any issues with the analysis or rubric, explain clearly what should be changed or refined.
\end{Verbatim}
\end{system}
\begin{user}[\texttt{User}]{}
\begin{Verbatim}[breaklines=true, breakanywhere=true, breaksymbol=, breakanywheresymbolpre=]
# Response that must be judged

```json
{"query": "{{ query_well_formed }}", "test_response": "{{ model_generated_answer }}", "baseline_response": "{{ reference_answer }}"}
```

# Output Format

```json
{"analysis": "YOUR ANALYSIS BASED ON THE INSTRUCTION FOLLOWING RUBRIC GOES HERE", "verdict": "NO_ISSUES" OR "MINOR_ISSUES" OR "MAJOR_ISSUES" OR "DEGENERATE_OUTPUT"}
```

YOUR OUTPUT MUST CONTAIN ONLY THE JSON OBJECT.
\end{Verbatim}
\end{user}
\end{minipage}
}

\newpage

\subsubsection{Answer Classification: Factuality}

{\fontsize{8pt}{8pt}\selectfont
\begin{minipage}{\linewidth}
\begin{system}[\texttt{System}]{}
\begin{Verbatim}[breaklines=true, breakanywhere=true, breaksymbol=, breakanywheresymbolpre=]
# Task

Analyze a model-generated response in relation to a provided textual context. The goal is to evaluate how well the response sentences are grounded in the context by assigning an appropriate label to each one. Use the guidelines below to conduct a thorough, sentence-level analysis.
Note: When evaluating the response, you must disregard any content that reflects internal reasoning, self-reflection, or planning (such as the model's thought process or deliberations, which is typically within "thinking" tags). Focus exclusively on the final answer provided to the query, as that is the intended content to assess.

# Guidelines

## Classification criteria
- "SUPPORTED": The sentence is entailed by the given context. Provide a supporting excerpt from the context. The supporting except must fully entail the sentence. If you need to cite multiple supporting excepts, simply concatenate them
- "UNSUPPORTED": The sentence is not entailed by the given context. No excerpt is needed for this label.
- "CONTRADICTORY":  The sentence is falsified by the given context. Provide a contradicting excerpt from the context.
- "NO_RAD": The sentence does not require factual attribution (e.g., opinions, greetings, questions, disclaimers). No excerpt is needed for this label.

## Instructions rubric
1. Decompose the response into individual sentences.
2. For each sentence, assign one of the labels from the "Classification criteria" guideline.
3. For each label, provide a short rationale explaining your decision. The rationale should be separate from the excerpt.
4. Be very strict with your "SUPPORTED" and "CONTRADICTORY" decisions. Unless you can find straightforward, indisputable evidence excerpts in the context that a sentence is "SUPPORTED" or "CONTRADICTORY", consider it "UNSUPPORTED". You should not employ world knowledge unless it is truly trivial.

# Example

## Input

```json
{"query": "What color are apples and bananas?", "context": "Apples are red fruits. Bananas are yellow fruits.", "response": "Apples are red. Bananas are green. Bananas are cheaper than apples. Enjoy your fruit!"}
```

## Output

```json
{"grounding_quality": [{"sentence": "Apples are red.", "label": "SUPPORTED", "rationale": "The context explicitly states that apples are red.", "excerpt": "Apples are red fruits."}, {"sentence": "Bananas are green.", "label": "CONTRADICTORY", "rationale": "The context states that bananas are yellow, not green.", "excerpt": "Bananas are yellow fruits."}, {"sentence": "Bananas are cheaper than apples.", "label": "UNSUPPORTED", "rationale": "The context does not mention the price of bananas or apples.", "excerpt": null}, {"sentence": "Enjoy your fruit!", "label": "NO_RAD", "rationale": "This is a general expression and does not require factual attribution.", "excerpt": null}]}
\end{Verbatim}
\end{system}
\begin{user}[\texttt{User}]{}
\begin{Verbatim}[breaklines=true, breakanywhere=true, breaksymbol=, breakanywheresymbolpre=]
# Response that must be judged

```json
{"query": "{{ query_well_formed }}", "context": "{% for passage in passages %}Reference [{{ loop.index }}] Text: {{ passage }} {% endfor %}", "response": "{{ model_generated_answer }}"}
```

# Output Format

```json
{"grounding_quality": [{"sentence": "ONE SENTENCE FROM THE RESPONSE GOES HERE", "label": "SUPPORTED" OR "UNSUPPORTED" OR "CONTRADICTORY" OR "NO_RAD", "rationale": "EXPLAIN YOUR DECISION HERE", "excerpt": "EXCERPT FROM THE CONTEXT GOES HERE"}, {"sentence": "ANOTHER SENTENCE FROM THE RESPONSE GOES HERE", "label": "SUPPORTED" OR "UNSUPPORTED" OR "CONTRADICTORY" OR "NO_RAD", "rationale": "EXPLAIN YOUR DECISION HERE", "excerpt": "EXCERPT FROM THE CONTEXT GOES HERE"}, CONTINUE WITH ALL THE REMAINING SENTENCES FROM THE RESPONSE HERE]}
```

YOUR OUTPUT MUST CONTAIN ONLY THE JSON OBJECT.
\end{Verbatim}
\end{user}
\end{minipage}
}

\newpage

\subsubsection{Query Characteristics: Well-formed}
\label{subsubsec:query-charactersitcs-prompts}

{\fontsize{10pt}{10pt}\selectfont
\begin{minipage}{\linewidth}
\begin{system}[\texttt{System}]{}
\begin{Verbatim}[breaklines=true, breakanywhere=true, breaksymbol=, breakanywheresymbolpre=]
# Task

Given a user query, generate a grammatically correct and well-formed version of the same query. Ensure proper grammar, punctuation, and capitalization, while preserving the original intent and meaning exactly as it is. Do not add any new information or change the content of the query in any way. The goal is to correct errors in structure without altering the core question or information.

# Examples

```json
{"query": "depona ab", "well_formed": "What is Depona AB?"}
{"query": "average teeth brushing time", "well_formed": "What is the average teeth brushing time?"}
{"query": "how many countries in africa", "well_formed": "How many countries are there in Africa?"}
{"query": "distance from earth to moon", "well_formed": "What is the distance from Earth to the Moon?"}
{"query": "what's the largest mammal in the world is?", "well_formed": "What is the largest mammal in the world?"}
{"query": "benefits of exercise for mental health", "well_formed": "What are the benefits of exercise for mental health?"}
{"query": "current presedent of the united states who?", "well_formed": "Who is the current president of the United States?"}
{"query": "when was the declaration of independence signed", "well_formed": "When was the Declaration of Independence signed?"}
{"query": "at what time was the moon landing on july 20 1969", "well_formed": "At what time did the moon landing occur on July 20, 1969?"}
{"query": ")what was the immediate impact of the success of the manhattan project?", "well_formed": "What was the immediate impact of the success of the Manhattan Project?"}
\end{Verbatim}
\end{system}
\begin{user}[\texttt{User}]{}
\begin{Verbatim}[breaklines=true, breakanywhere=true, breaksymbol=, breakanywheresymbolpre=]
# Query that must be well-formed

```json
{"query": "{{ query }}"}
```

# Output Format

```json
{"query_well_formed": "YOUR OUTPUT GOES HERE"}
```

YOUR OUTPUT MUST CONTAIN ONLY THE JSON OBJECT.
\end{Verbatim}
\end{user}
\end{minipage}
}

\newpage

\subsubsection{Query Characteristics: Recency}

{\fontsize{9pt}{9pt}\selectfont
\begin{minipage}{\linewidth}
\begin{system}[\texttt{System}]{}
\begin{Verbatim}[breaklines=true, breakanywhere=true, breaksymbol=, breakanywheresymbolpre=]
# Task

Given a user query, classify it based on its type and recency using exclusively the following classes. Ensure that the classification is appropriate and reflects the nature and timeliness of the query. The classification must strictly use only the classes provided.

# Classes

## EVERGREEN

Definition: Queries asking for facts or information that does not change over time. These queries are typically timeless and don't rely on current events or real-time data.

Examples:

```json
{"query": "What is the capital of France?", "type": "EVERGREEN"}
{"query": "What is the definition of photosynthesis?", "type": "EVERGREEN"}
{"query": "What are the benefits of regular exercise?", "type": "EVERGREEN"}
{"query": "What are the different types of renewable energy?", "type": "EVERGREEN"}
{"query": "What year was the original Lion King movie released?", "type": "EVERGREEN"}
```

## SLOW_CHANGING

Definition: Queries that require information that doesn't change frequently. These queries are time-sensitive but can tolerate a longer recency window, typically ranging from one month to a year or more. They may still be impacted by trends, but do not require immediate updates.

Examples:

```json
{"query": "Who is the U.S. president?", "type": "SLOW_CHANGING"}
{"query": "When is the next full moon?", "type": "SLOW_CHANGING"}
{"query": "When is the next Super Bowl?", "type": "SLOW_CHANGING"}
{"query": "When is the next earnings call of Apple?", "type": "SLOW_CHANGING"}
{"query": "Who owns the Fantasy hotel in Las Vegas?", "type": "SLOW_CHANGING"}
```

## FAST_CHANGING

Definition: Queries that are dependent on real-time information or the latest news. These queries require up-to-date data, generally within the past seven days, and reflect current events, breaking news, or recent changes.

Examples:

```json
{"query": "Where is the tornado now?", "type": "FAST_CHANGING"}
{"query": "What is the latest iPhone?", "type": "FAST_CHANGING"}
{"query": "What's the stock price of Tesla?", "type": "FAST_CHANGING"}
{"query": "What's the highest temperature today?", "type": "FAST_CHANGING"}
{"query": "What was the score of the last NBA match?", "type": "FAST_CHANGING"}
\end{Verbatim}
\end{system}
\begin{user}[\texttt{User}]{}
\begin{Verbatim}[breaklines=true, breakanywhere=true, breaksymbol=, breakanywheresymbolpre=]
# Query that must be classified

```json
{"query": "{{ query_well_formed }}"}
```

# Output Format

```json
{"type": "THE CLASS GOES HERE"}
```

THE CLASSIFICATION MUST STRICTLY USE ONLY THE CLASSES PROVIDED. YOUR OUTPUT MUST CONTAIN ONLY THE JSON OBJECT.
\end{Verbatim}
\end{user}
\end{minipage}
}

\newpage

\subsubsection{Query Characteristics: Popularity}

{\fontsize{9pt}{9pt}\selectfont
\begin{minipage}{\linewidth}
\begin{system}[\texttt{System}]{}
\begin{Verbatim}[breaklines=true, breakanywhere=true, breaksymbol=, breakanywheresymbolpre=]
# Task

Given a user query, classify it based on its popularity using exclusively the following classes. Ensure that the classification is appropriate and reflects the general popularity or niche nature of the query. The classification must strictly use only the classes provided.

# Classes

## HEAD

Definition: Queries that cover widely-known, frequently discussed subjects. These queries typically deal with mainstream or commonly taught concepts, topics that receive significant media coverage, or are high-frequency search terms.

Examples:

```json
{"query": "Who wrote 'Romeo and Juliet'?", "popularity": "HEAD"}
{"query": "What is the capital of France?", "popularity": "HEAD"}
{"query": "What is the formula for water?", "popularity": "HEAD"}
{"query": "Who was the first President of the United States?", "popularity": "HEAD"}
```

## TORSO

Definition: Queries about moderately popular topics, often not mainstream but still relatively well-known. These subjects are secondary or supporting concepts within a field, may require some specialized knowledge, or be topics covered in intermediate-level courses.

Examples:

```json
{"query": "What is the main export of Brazil?", "popularity": "TORSO"}
{"query": "What is the largest city in Canada by population?", "popularity": "TORSO"}
{"query": "What are the primary components of the Earth's atmosphere?", "popularity": "TORSO"}
{"query": "Who was the leader of the Soviet Union during World War II?", "popularity": "TORSO"}
```

## TAIL

Definition: Queries that cover niche or specialized topics, which are rarely discussed subjects or highly specific concepts. These queries are generally about topics that appear infrequently in standard curricula, have low-frequency search terms, or involve advanced or technical fields.

Examples:

```json
{"query": "What are the latest developments in quantum computing?", "popularity": "TAIL"}
{"query": "What is the chemical composition of the enzyme catalase?", "popularity": "TAIL"}
{"query": "Explain the role of mitochondrial DNA in tracing genetic ancestry", "popularity": "TAIL"}
{"query": "Who was the prime minister of New Zealand during the 1973 oil crisis?", "popularity": "TAIL"}
\end{Verbatim}
\end{system}
\begin{user}[\texttt{User}]{}
\begin{Verbatim}[breaklines=true, breakanywhere=true, breaksymbol=, breakanywheresymbolpre=]
# Query that must be classified

```json
{"query": "{{ query_well_formed }}"}
```

# Output Format

```json
{"popularity": "THE CLASS GOES HERE"}
```

THE CLASSIFICATION MUST STRICTLY USE ONLY THE CLASSES PROVIDED. YOUR OUTPUT MUST CONTAIN ONLY THE JSON OBJECT.
\end{Verbatim}
\end{user}
\end{minipage}
}

\newpage

\subsubsection{Query Characteristics: Validity}

{\fontsize{6pt}{6pt}\selectfont
\begin{minipage}{\linewidth}
\begin{system}[\texttt{System}]{}
\begin{Verbatim}[breaklines=true, breakanywhere=true, breaksymbol=, breakanywheresymbolpre=]
# Task

Given a user query, classify its validity across five dimensions: UNDERSTANDABLE, ANSWERABILITY, HARMLESS, FALSE_PREMISE, and INFORMATION_SEEKING. Each query should be classified as either "VALID" or "INVALID" for each dimension. The classification must strictly use only the classes provided.

# Dimensions

## UNDERSTANDABLE

Definition: Queries should be clearly formulated and understandable as requests for information. A query is "INVALID" if it contains vague pronouns, ambiguous phrasing, or is syntactically broken in a way that prevents interpretation.

Examples:

```json
{"query": "What does?", "UNDERSTANDABLE": "INVALID"}
{"query": "What about the US's?", "UNDERSTANDABLE": "INVALID"}
{"query": "When did he won the last one?", "UNDERSTANDABLE": "INVALID"}
{"query": "What is the history of Portuguese wine?", "UNDERSTANDABLE": "VALID"}
{"query": "Who was the first president of the United States?", "UNDERSTANDABLE": "VALID"}
```

## ANSWERABILITY

Definition: A query is answerable if it can reasonably be responded to using factual information, common knowledge, or available resources. Queries that are too vague, ambiguous, or not requesting specific information are "INVALID".

Examples:

```json
{"query": "What's going on?", "ANSWERABILITY": "INVALID"}
{"query": "How tall is the tower?", "ANSWERABILITY": "INVALID"}
{"query": "How can I apply for a visa?", "ANSWERABILITY": "VALID"}
{"query": "Are iPhone 14 cases different?", "ANSWERABILITY": "VALID"}
{"query": "What is the most popular noise cancelling earbuds?", "ANSWERABILITY": "VALID"}
```

## HARMLESS

Definition: A harmless query should not include sensitive, illegal, or harmful content. It should avoid offensive or discriminatory language and not request unethical actions.

Examples:

```json
{"query": "How can I kill myself?", "HARMLESS": "INVALID"}
{"query": "Which gender is better at math?", "HARMLESS": "INVALID"}
{"query": "What is the best religion to follow?", "HARMLESS": "INVALID"}
{"query": "How can I download music from YouTube for free?", "HARMLESS": "INVALID"}
{"query": "What are the main religions in Europe?", "HARMLESS": "VALID"}
{"query": "What are the latest Math Olympiad results?", "HARMLESS": "VALID"}
{"query": "What are the subscription options for YouTube Music?", "HARMLESS": "VALID"}
```

## FALSE_PREMISE

Definition: A query has a false premise if it assumes a fact that is incorrect or nonsensical. These often arise from misinformation or anachronisms. If the premise is correct or plausible, the query is "VALID".

Examples:

```json
{"query": "How often does Confucius replace his car brake pads?", "FALSE_PREMISE": "INVALID"}
{"query": "What's the name of Taylor Swift's rap album before she transitioned to pop?", "FALSE_PREMISE": "INVALID"}
{"query": "What's the name of Taylor Swift's last album?", "FALSE_PREMISE": "VALID"}
{"query": "How often do you need to replace your car brakes?", "FALSE_PREMISE": "VALID"}
```

## INFORMATION_SEEKING

Definition: An information_seeking query shows a clear intent to acquire factual knowledge, clarification, or an explanation. Commands, non-queries, or creative writing prompts are "INVALID" in this context.

Examples:

```json
{"query": "What are you doing?", "INFORMATION_SEEKING": "INVALID"}
{"query": "Are you available?", "INFORMATION_SEEKING": "INVALID"}
{"query": "Write a poem on flowers", "INFORMATION_SEEKING": "INVALID"}
{"query": "Write a sonnet to my spouse for Valentine's Day", "INFORMATION_SEEKING": "INVALID"}
{"query": "What time is it in Seattle?", "INFORMATION_SEEKING": "VALID"}
{"query": "What is the temperature today?", "INFORMATION_SEEKING": "VALID"}
{"query": "What are the symptoms of COVID-19?", "INFORMATION_SEEKING": "VALID"}
{"query": "When does Target at Capital Ave. close?", "INFORMATION_SEEKING": "VALID"}
{"query": "What are the emerging trends in artificial intelligence?", "INFORMATION_SEEKING": "VALID"}
\end{Verbatim}
\end{system}
\begin{user}[\texttt{User}]{}
\begin{Verbatim}[breaklines=true, breakanywhere=true, breaksymbol=, breakanywheresymbolpre=]
# Query that must be classified

```json
{"query": "{{ query_well_formed }}"}
```

# Output Format

```json
{"validity": {"UNDERSTANDABLE": "VALID" OR "INVALID", "ANSWERABILITY": "VALID" OR "INVALID", "HARMLESS": "VALID" OR "INVALID", "FALSE_PREMISE": "VALID" OR "INVALID", "INFORMATION_SEEKING": "VALID" OR "INVALID"}}
```

THE CLASSIFICATION MUST STRICTLY USE ONLY THE CLASSES PROVIDED. YOUR OUTPUT MUST CONTAIN ONLY THE JSON OBJECT.
\end{Verbatim}
\end{user}
\end{minipage}
}

\newpage

\subsubsection{Query Characteristics: Complexity}

{\fontsize{5.5pt}{5.5pt}\selectfont
\begin{minipage}{\linewidth}
\begin{system}[\texttt{System}]{}
\begin{Verbatim}[breaklines=true, breakanywhere=true, breaksymbol=, breakanywheresymbolpre=]
# Task

Given a user query, classify it based on its complexity using exclusively the following classes. Ensure that the classification is appropriate and reflects the complexity of the query. The classification must strictly use only the classes provided.

# Classes

## SIMPLE

Definition: Queries asking for simple facts. These queries are straightforward and do not require complex reasoning or conditions.

Examples:

```json
{"query": "When was Albert Einstein born?", "complexity": "SIMPLE"}
{"query": "When was FC Barcelona founded?", "complexity": "SIMPLE"}
{"query": "When did Tom in America first hit theaters?", "complexity": "SIMPLE"}
{"query": "Which year did Netflix last raise their subscription prices?", "complexity": "SIMPLE"}
```

## SIMPLE_WITH_CONDITION

Definition: Queries asking for simple facts with a given condition, such as a specific date or context. These queries may require to incorporate additional context, but the core of the query remains simple.

Examples:

```json
{"query": "What was the Amazon stock on 1st December?", "complexity": "SIMPLE_WITH_CONDITION"}
{"query": "What is the most active volcano in the Philippines?", "complexity": "SIMPLE_WITH_CONDITION"}
{"query": "What was the last thriller movie released by Quentin Tarantino?", "complexity": "SIMPLE_WITH_CONDITION"}
```

## SET

Definition: Queries that expect a set of entities or objects as the answer. These queries generally ask for a list or a group of items rather than a single fact.

Examples:

```json
{"query": "What are the Quentin Tarantino movies?", "complexity": "SET"}
{"query": "Who were the members of the band ABBA?", "complexity": "SET"}
{"query": "What are the continents in the southern hemisphere?", "complexity": "SET"}
```

## COMPARISON

Definition: Queries that compare two entities or objects. These queries involve a direct comparison between two items and expect an answer that highlights differences or preferences.

Examples:

```json
{"query": "Is iPhone performing better than Samsung?", "complexity": "COMPARISON"}
{"query": "Who started performing earlier, Adele or Ed Sheeran?", "complexity": "COMPARISON"}
{"query": "Which university has a higher student-to-faculty ratio, Harvard or Princeton?", "complexity": "COMPARISON"}
{"query": "What was the minimum stock price of Aurora Mobile Limited over the past month?", "complexity": "COMPARISON"}
```

## AGGREGATION

Definition: Queries that require aggregation or counting based on retrieved results. These queries often involve numerical values or totals, such as counts or sums.

Examples:

```json
{"query": "How many teams make up the NFL?", "complexity": "AGGREGATION"}
{"query": "How many total games did Utah Jazz win during 2021?", "complexity": "AGGREGATION"}
{"query": "How many music videos has the band Radiohead released?", "complexity": "AGGREGATION"}
{"query": "How many tech stocks have a higher market cap than Nvidia?", "complexity": "AGGREGATION"}
```

## MULTI_HOP

Definition: Queries that require chaining multiple pieces of information to compose the answer. These queries often involve a sequence of facts or steps that must be combined to arrive at the final answer.

Examples:

```json
{"query": "Who acted in Ang Lee's latest movie?", "complexity": "MULTI_HOP"}
{"query": "What is the shortest highway in the US in feet?", "complexity": "MULTI_HOP"}
{"query": "Who is the first actress to play the bond girl?", "complexity": "MULTI_HOP"}
{"query": "What was Mike Epps's age at the time of Next Friday's release?", "complexity": "MULTI_HOP"}
```

## POST_PROCESSING_HEAVY

Definition: Queries that require reasoning or significant processing of the retrieved information to generate an answer. These queries may require additional calculations, aggregations, or analysis beyond simple retrieval.

Examples:

```json
{"query": "How many days have passed since the latest NBA win of the LA Lakers?", "complexity": "POST_PROCESSING_HEAVY"}
{"query": "What was the average annual revenue for music streaming from 2020 to 2022?", "complexity": "POST_PROCESSING_HEAVY"}
{"query": "How many 3-point attempts did Steve Nash average per game in seasons he made the 50-40-90 club?", "complexity": "POST_PROCESSING_HEAVY"}
\end{Verbatim}
\end{system}
\begin{user}[\texttt{User}]{}
\begin{Verbatim}[breaklines=true, breakanywhere=true, breaksymbol=, breakanywheresymbolpre=]
# Query that must be classified

```json
{"query": "{{ query_well_formed }}"}
```

# Output Format

```json
{"complexity": "THE CLASS GOES HERE"}
```

THE CLASSIFICATION MUST STRICTLY USE ONLY THE CLASSES PROVIDED. YOUR OUTPUT MUST CONTAIN ONLY THE JSON OBJECT.
\end{Verbatim}
\end{user}
\end{minipage}
}

\newpage

\subsubsection{Query Characteristics: Domain}

{\fontsize{5pt}{5pt}\selectfont
\begin{minipage}{\linewidth}
\begin{system}[\texttt{System: part 1}]{}
\begin{Verbatim}[breaklines=true, breakanywhere=true, breaksymbol=, breakanywheresymbolpre=]
# Task

Given a user query, classify it based on its category using exclusively the following classes. Ensure that the classification is appropriate and reflects the category of the query. The classification must strictly use only the classes provided.

# Classes

## ARTS_AND_ENTERTAINMENT

Definition: Queries related to the arts, entertainment, music, movies, television, and performing arts.

Examples:

```json
{"query": "What are the top-rated TV shows?", "category": "ARTS_AND_ENTERTAINMENT"}
{"query": "Who won the Oscar for Best Picture?", "category": "ARTS_AND_ENTERTAINMENT"}
{"query": "What is the latest album released by Taylor Swift?", "category": "ARTS_AND_ENTERTAINMENT"}
{"query": "Who played the lead role in the latest Marvel movie?", "category": "ARTS_AND_ENTERTAINMENT"}
```

## COMPUTERS_AND_ELECTRONICS

Definition: Queries related to computers, electronics, gadgets, software, hardware, and related topics.

Examples:

```json
{"query": "What is the latest iPhone model?", "category": "COMPUTERS_AND_ELECTRONICS"}
{"query": "What are the best wireless earbuds?", "category": "COMPUTERS_AND_ELECTRONICS"}
{"query": "How do I build a gaming PC on a budget?", "category": "COMPUTERS_AND_ELECTRONICS"}
{"query": "What is the difference between RAM and ROM?", "category": "COMPUTERS_AND_ELECTRONICS"}
```

## HEALTH

Definition: Queries related to health, medical conditions, wellness, fitness, mental health, nutrition, and medical advice.

Examples:

```json
{"query": "How can I reduce stress?", "category": "HEALTH"}
{"query": "What are the symptoms of flu?", "category": "HEALTH"}
{"query": "How much water should I drink daily?", "category": "HEALTH"}
{"query": "What are the best exercises for weight loss?", "category": "HEALTH"}
```

## JOBS_AND_EDUCATION

Definition: Queries related to careers, job opportunities, education, schools, universities, and learning resources.

Examples:

```json
{"query": "How can I improve my math skills?", "category": "JOBS_AND_EDUCATION"}
{"query": "What are effective study techniques for exams?", "category": "JOBS_AND_EDUCATION"}
{"query": "What are the top universities in the world for engineering?", "category": "JOBS_AND_EDUCATION"}
{"query": "What qualifications do I need to become a software engineer?", "category": "JOBS_AND_EDUCATION"}
```

## HOME_AND_GARDEN

Definition: Queries related to home improvement, gardening, household tasks, and decor.

Examples:

```json
{"query": "How do I grow tomatoes indoors?", "category": "HOME_AND_GARDEN"}
{"query": "How can I remove stains from a carpet?", "category": "HOME_AND_GARDEN"}
{"query": "What are the best plants for a low-light room?", "category": "HOME_AND_GARDEN"}
{"query": "What are some budget-friendly home decor ideas?", "category": "HOME_AND_GARDEN"}
```

## LAW_AND_GOVERNMENT

Definition: Queries related to laws, government policies, legal advice, and governance.

Examples:

```json
{"query": "How can I file for divorce in the US?", "category": "LAW_AND_GOVERNMENT"}
{"query": "What are the rights of employees under labor law?", "category": "LAW_AND_GOVERNMENT"}
{"query": "What is the process for obtaining a visa to work in the UK?", "category": "LAW_AND_GOVERNMENT"}
{"query": "What are the legal requirements for starting a business in Canada?", "category": "LAW_AND_GOVERNMENT"}
```

## TRAVEL

Definition: Queries related to travel destinations, transportation, accommodation, and tourism activities.

Examples:

```json
{"query": "How do I apply for a visa to Europe?", "category": "TRAVEL"}
{"query": "What is the best time to visit Japan?", "category": "TRAVEL"}
{"query": "How can I find affordable hotels in Paris?", "category": "TRAVEL"}
{"query": "What are the top tourist attractions in New York City?", "category": "TRAVEL"}
```

## SCIENCE

Definition: Queries related to various scientific fields such as biology, chemistry, physics, and environmental science.

Examples:

```json
{"query": "How do black holes form?", "category": "SCIENCE"}
{"query": "How does photosynthesis work?", "category": "SCIENCE"}
{"query": "What is the theory of relativity?", "category": "SCIENCE"}
{"query": "What causes the greenhouse effect?", "category": "SCIENCE"}
```

## BUSINESS_AND_INDUSTRIAL

Definition: Queries related to business operations, industries, companies, and economic activities.

Examples:

```json
{"query": "How can I start a small business?", "category": "BUSINESS_AND_INDUSTRIAL"}
{"query": "How do supply chains impact global trade?", "category": "BUSINESS_AND_INDUSTRIAL"}
{"query": "What are the largest tech companies in the world?", "category": "BUSINESS_AND_INDUSTRIAL"}
{"query": "What are the key factors for successful project management?", "category": "BUSINESS_AND_INDUSTRIAL"}
```
\end{Verbatim}
\end{system}
\end{minipage}
}
\newpage
{\fontsize{5pt}{5pt}\selectfont
\begin{minipage}{\linewidth}
\begin{system}[\texttt{System: part 2}]{}
\begin{Verbatim}[breaklines=true, breakanywhere=true, breaksymbol=, breakanywheresymbolpre=]
## HOBBIES_AND_LEISURE

Definition: Queries related to hobbies, recreational activities, and leisure pursuits.

Examples:

```json
{"query": "What are some popular hiking trails in Switzerland?", "category": "HOBBIES_AND_LEISURE"}
{"query": "How can I get started with photography?", "category": "HOBBIES_AND_LEISURE"}
{"query": "What are some fun DIY projects to do at home?", "category": "HOBBIES_AND_LEISURE"}
{"query": "What are the best board games for a family night?", "category": "HOBBIES_AND_LEISURE"}
```

## BOOKS_AND_LITERATURE

Definition: Queries related to books, literature, authors, and reading recommendations.

Examples:

```json
{"query": "Who wrote '1984'?", "category": "BOOKS_AND_LITERATURE"}
{"query": "Who are some notable contemporary poets?", "category": "BOOKS_AND_LITERATURE"}
{"query": "What is the plot of 'Pride and Prejudice'?", "category": "BOOKS_AND_LITERATURE"}
{"query": "What are the best fantasy novels of the decade?", "category": "BOOKS_AND_LITERATURE"}
```

## SPORTS

Definition: Queries related to sports, athletes, teams, events, and competitions.

Examples:

```json
{"query": "Who won the Super Bowl?", "category": "SPORTS"}
{"query": "When is the next FIFA World Cup?", "category": "SPORTS"}
{"query": "What is the world record for the 100-meter sprint?", "category": "SPORTS"}
{"query": "Who holds the record for most goals in a single Premier League season?", "category": "SPORTS"}
```

## NEWS

Definition: Queries related to current events, news stories, and media coverage.

Examples:

```json
{"query": "What happened in the latest presidential election?", "category": "NEWS"}
{"query": "What is the latest update on the COVID-19 pandemic?", "category": "NEWS"}
{"query": "What are the latest developments in the global economy?", "category": "NEWS"}
{"query": "What is the current status of the Paris Agreement on climate change?", "category": "NEWS"}
```

## BEAUTY_AND_FITNESS

Definition: Queries related to beauty products, makeup, skincare, fitness routines, and wellness.

Examples:

```json
{"query": "How can I build muscle mass?", "category": "BEAUTY_AND_FITNESS"}
{"query": "What are some effective skincare routines?", "category": "BEAUTY_AND_FITNESS"}
{"query": "What are the benefits of yoga for mental health?", "category": "BEAUTY_AND_FITNESS"}
{"query": "How can I create a hair care routine for dry hair?", "category": "BEAUTY_AND_FITNESS"}
```

## FINANCE

Definition: Queries related to financial advice, investments, economics, and money management.

Examples:

```json
{"query": "How can I save for retirement?", "category": "FINANCE"}
{"query": "How can I improve my credit score?", "category": "FINANCE"}
{"query": "What is the difference between a 401(k) and an IRA?", "category": "FINANCE"}
{"query": "What is the best way to budget my monthly expenses?", "category": "FINANCE"}
```

## PEOPLE_AND_SOCIETY

Definition: Queries related to society, human behavior, relationships, and cultural issues.

Examples:

```json
{"query": "What are the causes of inequality?", "category": "PEOPLE_AND_SOCIETY"}
{"query": "How can communities address homelessness?", "category": "PEOPLE_AND_SOCIETY"}
{"query": "How do different cultures celebrate New Year?", "category": "PEOPLE_AND_SOCIETY"}
{"query": "What are the psychological effects of social media?", "category": "PEOPLE_AND_SOCIETY"}
```

## AUTOS_AND_VEHICLES

Definition: Queries related to cars, vehicles, transportation, and road safety.

Examples:

```json
{"query": "How do electric cars work?", "category": "AUTOS_AND_VEHICLES"}
{"query": "How can I maintain my car's engine?", "category": "AUTOS_AND_VEHICLES"}
{"query": "What are the benefits of hybrid cars?", "category": "AUTOS_AND_VEHICLES"}
{"query": "What is the fuel efficiency of a Tesla Model S?", "category": "AUTOS_AND_VEHICLES"}
```

## GAMES

Definition: Queries related to video games, board games, game mechanics, and gaming news.

Examples:

```json
{"query": "How do I level up fast in Fortnite?", "category": "GAMES"}
{"query": "What are the top upcoming video games?", "category": "GAMES"}
{"query": "What is the best strategy in Minecraft?", "category": "GAMES"}
{"query": "How do you unlock new characters in Super Smash Bros.?", "category": "GAMES"}
```

## TIME_AND_WEATHER

Definition: Queries related to time, weather, and climate forecasts.

Examples:

```json
{"query": "How do weather patterns affect agriculture?", "category": "TIME_AND_WEATHER"}
{"query": "How many hours are there between GMT and EST?", "category": "TIME_AND_WEATHER"}
{"query": "What will the weather be like tomorrow in New York?", "category": "TIME_AND_WEATHER"}
{"query": "What is the best time of year to visit the Caribbean?", "category": "TIME_AND_WEATHER"}
```
\end{Verbatim}
\end{system}
\end{minipage}
}
\newpage
{\fontsize{5pt}{5pt}\selectfont
\begin{minipage}{\linewidth}
\begin{system}[\texttt{System: part 3}]{}
\begin{Verbatim}[breaklines=true, breakanywhere=true, breaksymbol=, breakanywheresymbolpre=]
## ONLINE_COMMUNITIES

Definition: Queries related to online forums, social media, and digital communities.

Examples:

```json
{"query": "How do I join a subreddit on Reddit?", "category": "ONLINE_COMMUNITIES"}
{"query": "How can I create a group on Facebook?", "category": "ONLINE_COMMUNITIES"}
{"query": "What are some popular online gaming communities?", "category": "ONLINE_COMMUNITIES"}
{"query": "What are the benefits of joining professional LinkedIn groups?", "category": "ONLINE_COMMUNITIES"}
```

## INTERNET_AND_TELECOM

Definition: Queries related to internet services, telecommunications, and online infrastructure.

Examples:

```json
{"query": "How does 5G work?", "category": "INTERNET_AND_TELECOM"}
{"query": "How can I improve my home Wi-Fi signal?", "category": "INTERNET_AND_TELECOM"}
{"query": "What is the fastest internet provider in the US?", "category": "INTERNET_AND_TELECOM"}
{"query": "What is the difference between fiber optic and broadband internet?", "category": "INTERNET_AND_TELECOM"}
```

## LOCAL_INFORMATION

Definition: Queries related to local businesses, services, and events.

Examples:

```json
{"query": "Is there a public library near me?", "category": "LOCAL_INFORMATION"}
{"query": "Where can I find a good gym in Miami?", "category": "LOCAL_INFORMATION"}
{"query": "What are the best restaurants in San Francisco?", "category": "LOCAL_INFORMATION"}
{"query": "What local events are happening this weekend in Chicago?", "category": "LOCAL_INFORMATION"}
```

## PETS_AND_ANIMALS

Definition: Queries related to pets, animal care, and wildlife.

Examples:

```json
{"query": "How do I train my dog to sit?", "category": "PETS_AND_ANIMALS"}
{"query": "What is the lifespan of a cat?", "category": "PETS_AND_ANIMALS"}
{"query": "How can I create a safe habitat for pet birds?", "category": "PETS_AND_ANIMALS"}
{"query": "What should I feed my rabbit for a healthy diet?", "category": "PETS_AND_ANIMALS"}
```

## STOCK

Definition: Queries related to stock markets, stock prices, and investment trends.

Examples:

```json
{"query": "How can I diversify my stock portfolio?", "category": "STOCK"}
{"query": "What is the current stock price of Apple?", "category": "STOCK"}
{"query": "What factors influence stock market fluctuations?", "category": "STOCK"}
{"query": "How does short selling work in the stock market?", "category": "STOCK"}
```

## RELIGION_AND_SPIRITUALITY

Definition: Queries related to religious beliefs, practices, spirituality, and theology.

Examples:

```json
{"query": "How do Christians celebrate Easter?", "category": "RELIGION_AND_SPIRITUALITY"}
{"query": "What are the main teachings of Buddhism?", "category": "RELIGION_AND_SPIRITUALITY"}
{"query": "What is the significance of Ramadan in Islam?", "category": "RELIGION_AND_SPIRITUALITY"}
{"query": "What are the core principles of Hinduism?", "category": "RELIGION_AND_SPIRITUALITY"}
```

## GEOGRAPHY

Definition: Queries related to geographical features, locations, maps, and global regions.

Examples:

```json
{"query": "What are the largest deserts on Earth?", "category": "GEOGRAPHY"}
{"query": "What is the longest river in the world?", "category": "GEOGRAPHY"}
{"query": "Which countries are part of Scandinavia?", "category": "GEOGRAPHY"}
{"query": "What is the tallest mountain in the world?", "category": "GEOGRAPHY"}
```

## HISTORY

Definition: Queries related to historical events, figures, and timelines.

Examples:

```json
{"query": "What caused the fall of the Roman Empire?", "category": "HISTORY"}
{"query": "What were the main causes of World War I?", "category": "HISTORY"}
{"query": "Who was the first President of the United States?", "category": "HISTORY"}
{"query": "Who were the key figures in the American Civil War?", "category": "HISTORY"}
```

## FOOD_AND_DRINK

Definition: Queries related to cooking, recipes, dining, and beverages.

Examples:

```json
{"query": "How do I make a perfect cheesecake?", "category": "FOOD_AND_DRINK"}
{"query": "What are some easy vegan dinner recipes?", "category": "FOOD_AND_DRINK"}
{"query": "How do I brew the perfect cup of coffee?", "category": "FOOD_AND_DRINK"}
{"query": "What are the health benefits of green tea?", "category": "FOOD_AND_DRINK"}
```

## SHOPPING

Definition: Queries related to purchasing items, shopping tips, and deals.

```json
{"query": "How do I find discounts on clothing online?", "category": "SHOPPING"}
{"query": "What should I look for when buying a laptop?", "category": "SHOPPING"}
{"query": "What are the best online stores for electronics?", "category": "SHOPPING"}
{"query": "What are the most popular shopping malls in New York?", "category": "SHOPPING"}
```
\end{Verbatim}
\end{system}
\end{minipage}
}
\newpage
{\fontsize{5pt}{5pt}\selectfont
\begin{minipage}{\linewidth}
\begin{system}[\texttt{System: part 4}]{}
\begin{Verbatim}[breaklines=true, breakanywhere=true, breaksymbol=, breakanywheresymbolpre=]
## OTHER

Definition: Queries that do not fit into any of the above categories.

Examples:

```json
{"query": "What is the meaning of life?", "category": "OTHER"}
{"query": "How can I improve my public speaking skills?", "category": "OTHER"}
{"query": "What are the benefits of practicing mindfulness?", "category": "OTHER"}
{"query": "How can I organize my daily schedule more efficiently?", "category": "OTHER"}
\end{Verbatim}
\end{system}
\begin{user}[\texttt{User}]{}
\begin{Verbatim}[breaklines=true, breakanywhere=true, breaksymbol=, breakanywheresymbolpre=]
# Query that must be classified

```json
{"query": "{{ query_well_formed] }}"}
```

# Output Format

```json
{"category": "THE CLASS GOES HERE"}
```

THE CLASSIFICATION MUST STRICTLY USE ONLY THE CLASSES PROVIDED. YOUR OUTPUT MUST CONTAIN ONLY THE JSON OBJECT.
\end{Verbatim}
\end{user}
\end{minipage}
}

\end{document}